\documentclass[twoside,11pt]{article}

\usepackage{blindtext}

\usepackage[preprint]{jmlr2e}

\newcommand*{\st}{s.t.\@\xspace}
\newcommand*{\eg}{e.g.\@\xspace}
\newcommand*{\ie}{i.e.\@\xspace}

\newcommand*{\cf}{cf.\@\xspace}
\newcommand*{\p}{p.\@\xspace}
\newcommand*{\wrt}{wrt.\@\xspace}

\newcommand{\naturals}{\mathbb{N}}
\newcommand{\reals}{\mathbb{R}}

\newcommand{\forecaster}{v}
\newcommand{\nature}{y}

\newcommand{\sceptic}{G}

\newcommand{\forecastingset}{\mathcal{P}}
\newcommand{\belief}{b}
\newcommand{\ball}{\mathbb{B}}
\newcommand{\dirac}{\delta}
\newcommand{\indexsetgamblers}{I}

\newcommand{\restrictionset}{R}

\newcommand{\gamble}{g}

\newcommand{\UnscaledRegretGambles}{\mathcal{L}}
\newcommand{\CalibrationGambles}{V}
\newcommand{\ScaledRegretGambles}{\mathcal{S}}

\newcommand{\capital}{K}

\newcommand{\ca}{\operatorname{ca}}
\newcommand{\ident}{\nu}
\newcommand{\scoring}{\ell} 
\newcommand{\loss}{\ell} 

\newcommand{\localupperexp}{\overline{\mathbb{E}}}

\newcommand{\Y}{\mathcal{Y}}
\newcommand{\Q}{\mathcal{Q}}

\newcommand{\offer}{\mathcal{G}}
\newcommand{\credal}{\mathcal{P}}
\newcommand{\CY}{C(\Y)}
\newcommand{\pqtopology}{\sigma(\CY, \ca(\Y))}
\newcommand{\qptopology}{\sigma(\ca(\Y), \CY)}

\newcommand{\cl}{\operatorname{cl}}
\newcommand{\cvxcl}{\overline{\operatorname{co}}}
\newcommand{\spr}{\operatorname{spr}}
\newcommand{\aggregation}{\mathbf{A}}
\newcommand{\propspace}{\mathcal{V}}

\newcommand{\argmin}{\operatornamewithlimits{argmin}}

\usepackage[dvipsnames]{xcolor}
\hypersetup{
colorlinks=true,
urlcolor=blue,
linkcolor=blue,
citecolor=OliveGreen
}

\usepackage{lastpage}
\jmlrheading{23}{2024}{1-\pageref{LastPage}}{?/??; Revised ?/??}{?/??}{21-0000}{Rabanus Derr and Robert C. Williamson}

\ShortHeadings{Evaluation Metrics As Averaged Outcomes of Fair Gambles}{Derr and Williamson}
\firstpageno{1}

\begin{document}

\title{Evaluation Metrics As Averaged Outcomes of Fair Gambles}
\author{\name Rabanus Derr \email rabanus.derr@uni-tuebingen.de \\
       \addr University of T\"{u}bingen and\\
       T\"{u}bingen AI Center\\
       T\"{u}bingen, 72076, Germany
       \AND
       \name Robert C. Williamson \email bob.williamson@uni-tuebingen.de \\
       \addr University of T\"{u}bingen and\\
       T\"{u}bingen AI Center\\
       T\"{u}bingen, 72076, Germany}

\editor{??}

\maketitle

\begin{abstract}
In the current practices of machine learning, the evaluation of forecasts has become a cornerstone of scientific progress.
A multitude of evaluation metrics have been suggested and used to qualify ``good'' forecasts.
What do those metrics share?
How are they related?
In this work, we use a protocol borrowed from game-theoretic probability to show that a large part of evaluation metrics are averaged outcomes of fair gambles.
Intuitively, a fair gambler is one which a forecaster would expect to fail. Hence, the gambler's ability to gain disproves the quality of the forecast.
Standard evaluation metrics are then variants of choices of such fair gambles.
In particular, this choice is structured along two dimensions, one of which separates calibration-type and regret-type metrics. In particular, this framework sheds light on the relationship of calibration and regret showing a theoretical equivalence in their ability to evaluate when being scaled appropriately, but the incomparability of obtained scores.
\end{abstract}
\begin{keywords}
    Evaluation, Calibration, Regret, Game-Theoretic Probability and Statistics
\end{keywords}



\section{Introduction}
\label{introduction}

In the current practices of machine learning, the evaluation of forecasts has become a cornerstone of scientific progress \citep{blum2015ladder, hardt2025emerging}. For instance, OpenAI reports loss scores and calibration scores for GPT-4 on different tasks \citep{openai2024gpt-4}. The inventors of the GraphCast weather forecasting model provide several metrics, among them root mean squared error \citep{lam2023learning}. The introduction of AlphaFold was accompanied by a series of evaluation metrics, \eg, accuracy or absolute difference, used to discuss the quality of the protein folding forecasts \citep{senior2020improved}. On the more formal side machine learning scholars have come up with dozens of evaluation metrics: different types of (proper) loss functions \citep{williamson2022geometry}, a multitude of calibration notions \citep{derr2025three, gopalan2025calibration}, and numerous types of regret \citep{cesa2006prediction}. \textbf{What do all these evaluation metrics share? In which regard do the evaluation metrics differ? How can they be viewed from a unified perspective?}

Among the used metrics, two main types have crystallized, one which we call \emph{regret-type} and one which we call \emph{calibration type}. Regret-type evaluation metrics compare a loss incurred by a predictor with the loss of some comparison predictor. Regret-type evaluation metrics require a loss function and a comparison baseline, which is either explicitly or implicitly given.\footnote{For instance, loss scores have an implicit baseline which is the predictor which knows the truth and hence often, depending on the definition of the loss function, incurs $0$ loss.} Regret-type evaluation metrics include external regret \citep[page 80]{cesa2006prediction}, swap regret \citep[page 91]{cesa2006prediction}, and loss scores (\cf Table~\ref{tab:belief + restriction gives evaulation metric}).

Calibration-type evaluation metrics (traditionally) compare average outcomes to average forecasts on certain instances \citep{dawid1985calibration, dawid2017individual, holtgen2023richness}. Those metrics have been generalized beyond the average, as \eg, mean and variance calibration \citep{jung2021moment}, quantile calibration \citep{jung2023batch} or property calibration \citep{gneiting2023regression, noarov2023scope, derr2025three}.
This raises the questions: \textbf{What is the relationship between the regret-type and calibration-type evaluation metrics? What is the structure of the regret-type and the calibration-type evaluation metrics?} (\cf Section~\ref{Related Work: Linking Calibration and Loss})

Towards answering the posed questions we commit to a general framework which we call an \emph{evaluation protocol}.
This protocol is based upon the game-theoretic probability setup developed in \citep{shafer2005probability, shafer2019game}.
Roughly summarized, the evaluation protocol consists of a nature, a forecaster and a set of gamblers.
For every forecast, (fair)\footnote{A gamble is fair if the forecaster literally expects the gamble to not be favorable for the gambler.} gambles by the gamblers are played and evaluated on the actual outcome revealed by nature.
For each gambler, the average realized value of the gambles is summarized as its capital. The capital is the obtained score.
It shows how much the forecasts and outcomes match, the bigger the capital the worse they match. Examples for scores are the expected calibration error, false positive rate, false negative rate, accuracy, mean squared error, multicalibration, bias in the large, external regret, swap regret, Brier score, and the log loss.
With our machinery we will argue that all those (and many more) \textbf{scores obtained by evaluation metrics are averaged realized values of (fair) gambles}.

\paragraph{Warm-Up}
\label{para:warm-up}
Let's shortly illustrate the evaluation protocol in a little example. Let $(\nature_t)_{t \in T}$ be a tuple of binary outcomes, \ie, $\nature_t \in \{ 0,1\}$ for all $t \in T$. Furthermore, let $(\forecaster_t)_{t \in T}$ be a tuple of probabilistic forecasts, $\forecaster_t \in  [0,1]$ for all $t \in T$. For every instance $t \in T$, a forecast $\forecaster_t$ is made, and let us suppose there is a gambler, who chooses to play the gamble $\sceptic_t \colon \nature \mapsto (\nature - \forecaster_t)^2 - (\nature - \nature_t)^2$.
Such a gamble is \emph{fair}, that is,
\begin{align*}
    \mathbb{E}_{\forecaster_t}[\sceptic_t] = (1-\forecaster_t) \sceptic_t(0) + \forecaster_t \sceptic_t(1) \le 0.
\end{align*}
In other words, the forecaster expects the gambler to not win. Hence, if indeed the gambler achieves a large capital, it disproves the abilities of the forecaster. Crucially, the gambler, when playing $\sceptic_t$ makes use of the actual outcome $\nature_t$ (see the definition). This (hypothetical) information is \emph{not} accessible to the forecaster in this fictive play.

Now, when we actually evaluate the gambles on the binary outcomes, and average, we obtain,
\begin{align}
\label{eq:Brier score}
    \frac{1}{n} \sum_{t \in T} \sceptic_t(\nature_t) = \frac{1}{n} \sum_{t \in T} (\nature_t - \forecaster_t)^2  - (\nature_t - \nature_t)^2 =  \frac{1}{n} \sum_{t \in T} (\nature_t - \forecaster_t)^2,
\end{align}
which is the \emph{Brier score} an evaluation metric introduced in meteorology \citep{brier1950verification} and used to evaluate class probability estimates.

It is the case that besides the Brier score, all kinds of calibration, \eg, expected calibration error, and regret-type evaluation metrics can be embedded in a comparable fashion. In particular, all those metrics respect that the gambler only plays \emph{fair} gambles. However, to generally reveal the underlying gambling structure it is necessary to move beyond binary outcomes and simple probabilistic forecasts.



\subsection{Paper Outline}
First, we summarize our contribution and clarify relevant terms and the scope of our work in the current section.
Then, we introduce the formal setup (Section~\ref{formal setup}). We require some machinery from convex analysis in infinite dimensional spaces. The complexity of the mathematical tools is needed to accurately reflect a wide range of evaluation metrics used in settings from classification to regression.
We continue by constructing the evaluation protocol based on game protocols used in game-theoretic probability.
We characterize the set of evaluation metrics which are representable in the evaluation protocol.
Furthermore, we adapt a ``sanity'' condition for the gamblers which we call \emph{fairness} (Definition~\ref{def:fair gambler and availability}) from \citep{shafer2019game}.\footnote{In \citep{shafer2019game}, the authors use the term ``available'' gambles, but don't formally define it.} Essentially, a gambler is fair if it only plays \emph{available} gambles, \ie, gambles which the forecaster assumes, in expectation, to lead to a non-positive outcome. We show that this criterion is equivalent to guaranteeing that whenever the forecaster is ``clairvoyant'', \ie, exactly forecasts the true outcome, the gambler will not increase its capital above $0$ (Proposition~\ref{proposition:sanity check}).

The general evaluation protocol allows forecasts to correspond to sets of probability distributions, \cf imprecise probabilities \citep{augustin2014introduction}.
We demarcate a subset of such forecasts using the concepts of elicitablity and identifiability from statistics, \eg \citep{gneiting2011making, steinwart2014elicitation}.
Equipped with these tools, we present our contributions which can be separated into two main parts (Section~\ref{sec:Characterization of Available Gambles} and Section~\ref{The Fair, Restricted and Rational Gambler}).
Finally, we contextualize our findings in the literature on evaluation of forecasts (Section~\ref{sec:related work}).

\begin{figure}[ht]
    \centering
    \def\svgwidth{0.95\columnwidth}
    {\footnotesize
    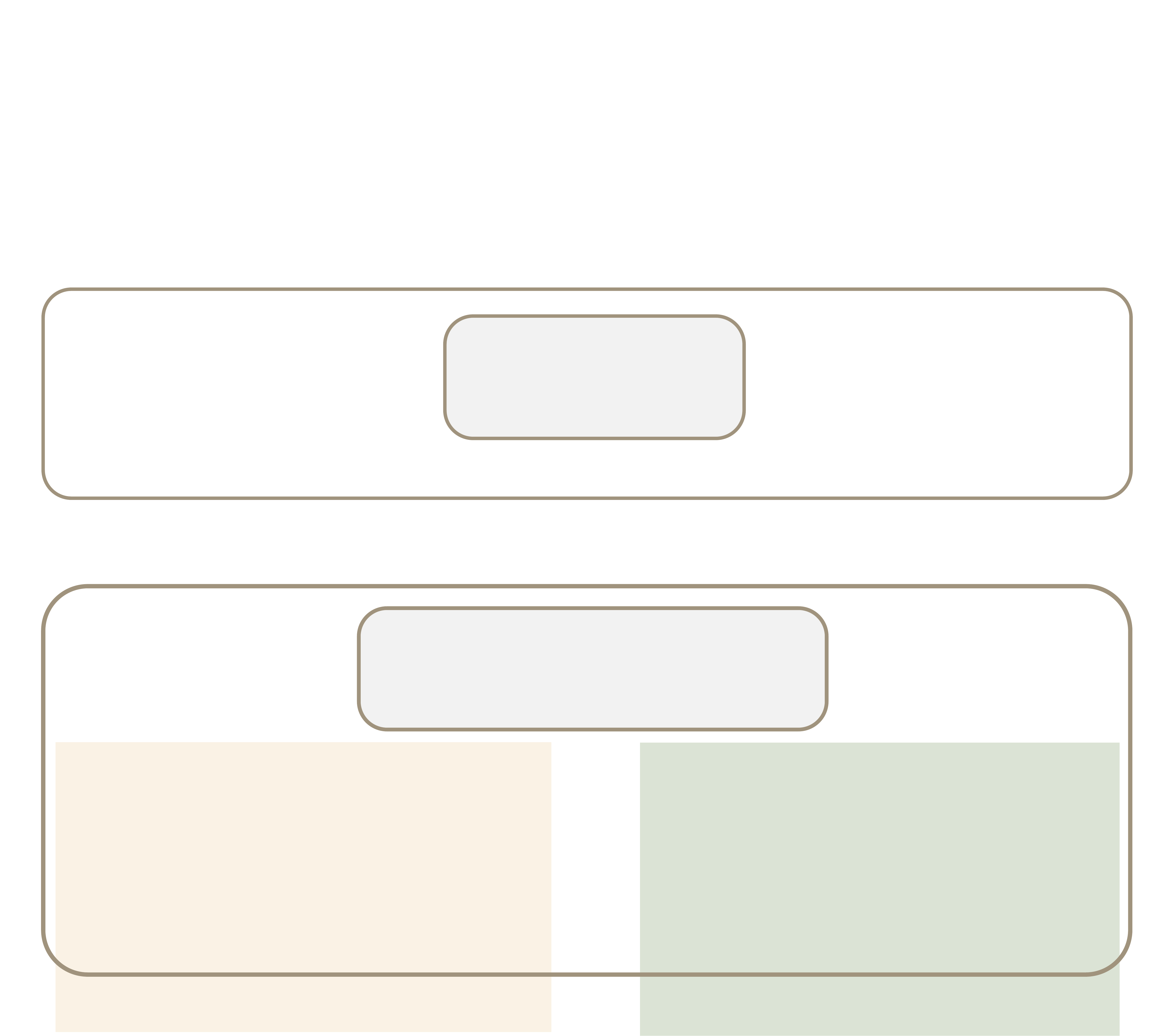
    }
    \caption{\emph{Graphical summary of the results in Section~\ref{the availability criterion}.} The mathematical notations are explained in Table~\ref{tab:notation}. The central element is the polar duality between closed, convex sets of probability distributions, \ie, credal sets, and convex cones of gambles, \ie, available gambles (highlighted in gray). We show that the set of available gambles, \ie, unfavorable gambles for a gambler, with respect to a forecasting set induced by an identifiable (respectively elicitable) property is equal to the set of gambles bounded above by calibration gambles (respectively scaled regret gambles). For example, the forecast of the set of probability distributions which share the mean $M \in \reals$, are tested by the set of gambles which are upper bounded by $\alpha(\nature - M)$ for some $\alpha \in \reals$, respectively upper bounded by $\beta (\nature - M)^2 - \beta (\nature - C)^2$ for some $\beta \ge 0$ and $C \in \reals$. Roughly speaking, calibration and scaled regret are the most stringent forms of testing a forecast.}
    \label{fig: this paper as a drawing}
\end{figure}

\paragraph{Contributions -- Section~\ref{sec:Characterization of Available Gambles}}
First, we develop a duality which allows us to fully characterize the set of available gambles. The central results of this part are summarized in Figure~\ref{fig: this paper as a drawing}.
\begin{enumerate}
    \item We show that \emph{(unscaled) regret gambles} and \emph{calibration gambles} (\cf Section~\ref{Regret Gambles and Calibration Gambles are Available}), which are central to the recovery results in Section~\ref{The Fair, Restricted and Rational Gambler}, are both available (Proposition~\ref{prop:regret gambles are available} and Proposition~\ref{prop:calibration gambles are available}).
    Furthermore, we fully characterize the set of available gambles in case the forecasts are aligned to elicitable (Theorem~\ref{thm:Characterization of Available Gambles in Protocol 2 with Elicitable Property}) or identifiable Properties (Theorem~\ref{thm:Characterization of Available Gambles in Protocol 2 with Identifiable Property}). In particular, \emph{scaled regret gambles} (respectively \emph{calibration gambles}) approximately dominate all available gambles. These statements are based upon a general polar duality between closed, convex sets of probability distributions, \ie, credal sets, and convex cones of gambles, \ie, available gambles (Theorem~\ref{thm:representation: credal sets - offers - MAIN}).\footnote{This duality is known for different technical setups, \eg, \citep{follmer2011stochastic, benavoli2017polarity}, however, not for the $\CY$-$\ca(\Y)$-pairing, \cf Section~\ref{sec:Proving the Characterizations}.} 
    \item The characterizations lead to the insight that scaled regret gambles are, in principle, equally powerful as calibration gambles. Both those types dominate the same set of available gambles (Corollary~\ref{corollary:Duality of Calibration and Regret}).
\end{enumerate}

\paragraph{Contributions -- Section~\ref{The Fair, Restricted and Rational Gambler}}
Second, we further constrain the gamblers to recover existing evaluation metrics and reveal a two-dimensional hierarchy of evaluation metrics.
\begin{enumerate}
    \item For the recovery of existing evaluation metrics, we introduce two further conditions on the gamblers, \emph{restrictiveness} and \emph{rationality}. A restricted gambler is bounded to gambles in a certain set. Intuitively, there is a capital constraint set on the gambler. For rationality, we equip the gambler with a belief about the outcome, comparable to an alternative hypothesis of what the gambler actually believes is true, and demand that the gambler optimizes the choice of gamble with respect to the expected value under this belief. By tuning restriction and rationality, we obtain the evaluation metrics summarized in Table~\ref{tab:belief + restriction gives evaulation metric}. For instance,
    the capital of a set of gamblers with access to all gambles with a bounded norm and a (true) belief about the average outcome on the subsets in which a certain value has been predicted is aggregated to calibration scores (Proposition~\ref{prop:recovery of calibration score}). The two dimensional structure, restriction and belief, leads to the conclusion that standard calibration scores and standard loss scores are incommensurable in the sense that they differ in both the dimensions.
    \item 
    We provide a hierarchy along the dimension of restrictions via set-containment (Proposition~\ref{prop:Bounding Capital by Hierarchy on Restrictions}) and along the dimension of beliefs via refinement (Proposition~\ref{prop:Bounding Capital by Refinment on Belief}) which generalize known inequalities between calibration and swap regret (Corollary~\ref{corollary:Recovery of Theorem 12}), between regret notions (Corollary~\ref{corollary:Recovery of Known Order of Regret}) and between calibration notions (Corollary~\ref{corollary:Recovery of Known Order of calibration scores}).
\end{enumerate}
\begin{table}[ht]
    \centering
\setlength{\extrarowheight}{15pt}
    \begin{tabular}{l|ll}
                                           \multicolumn{1}{c}{}           & \multicolumn{2}{c}{\textbf{Restriction on gambler}}                                \\
                        \textbf{Belief of gambler}                           & \multicolumn{1}{c}{ Regret Gambles $\UnscaledRegretGambles_{\gamma}$} & \multicolumn{1}{c}{Norm Ball $\ball_\alpha$}                    \\ \hhline{-|--}
                        Average outcome     & \makecell[l]{External Regret\\\tiny{(Proposition~\ref{prop:Recovery of (Generalized) External Regret})}}                  & \makecell[l]{Bias in the large\\ \tiny{(Proposition~\ref{prop:recovery of bias in the large})} }           \\
                        \makecell[l]{Average outcome\\on prediction-groups} & \makecell[l]{Swap Regret\\\tiny{(Proposition~\ref{prop:recovery of swap regret})}}                      & \makecell[l]{Calibration Score\\ \tiny{(Proposition~\ref{prop:recovery of calibration score})}}             \\
                        \makecell[l]{Average outcome\\on subgroups} & \makecell[l]{Group-Wise Swap Regret\\\tiny{(Proposition~\ref{prop:Recovery of Group-Wise Swap Regret})}}                      & \makecell[l]{Multicalibration Score\\ \tiny{(Proposition~\ref{prop:Recovery of (Generalized) Multicalibration Score})}}             \\
                        \makecell[l]{Exact knowledge of\\ true outcome}        & \makecell[l]{Zeroed Loss Score\\\tiny{(Proposition~\ref{prop:recovery of loss scores})}}                  & \makecell[l]{Individual Calibration Score\\\tiny{(Proposition~\ref{prop:recovery of individual calibration})}} 
\end{tabular}
    \caption{
    Comparison of belief and restrictions of gambler and the resulting evaluation metrics. The corresponding propositions where the recovery is shown are referenced below the metric. We neglect details about the type of aggregation, the outcome set $\Y$ and the forecasted property $\Gamma$. A \emph{prediction-group} is a set of instances which share the predicted value $\gamma$. A \emph{subgroup} is a set of instances which share the predicted value $\gamma$ and a certain attribute $S$.
    }
    \label{tab:belief + restriction gives evaulation metric}
\end{table}

\subsection{Clarification of Terms and Scope}
\label{sec:clarification of terms and scope}

\paragraph{Evaluation as Certificate}
The question of how one should evaluate forecasts has always been part of debates since the beginning of machine learning \citep{degroot1983comparison, schervish1989general, gneiting2011making, williamson2022geometry}. However, large parts of machine learning literature concentrate on evaluation of forecasts as ``modeling an objective'', \eg, \citep{bartlett2006convexity, jung2021moment, gupta2022faster, deng2023happy}. We focus on evaluation of forecasts as ``providing a certificate of quality'', a perspective shared by the classical literature on the evaluation of meteorological forecasts \citep{brier1950verification, murphy1967verification}. In particular, the evaluation of forecasts is critical for the value of a forecast. Forecasts on their own are of limited value. When they come with an accompanying guide how to evaluate them, they become useful. The evaluation of forecasts \emph{make} the forecasts, \cf \citep{dawid2017individual, holtgen2024practical, perdomo2025defense}.

\paragraph{Clarification of Terms}
To be clear, the test for consistency of forecasts with observations is what we call \emph{evaluation}.\footnote{This is what has been called ``empirical evaluation'' by \citet{murphy1967verification}. The same authors name another type of evaluation, which they refer to as ``operational'', concerned with the value of the forecast to the user.} The actual mapping from forecasts and outcomes to some real number is what we call an \emph{evaluation metric} and the resulting number the \emph{score}.

\paragraph{No Algorithmic Solution}
In this work, we suggest an abstract evaluation protocol to map the landscape of \emph{evaluation metrics}. We do not propose any algorithmic solution to give forecasts optimizing an evaluation metric.\footnote{Nevertheless, we mention the work by \citet{abernethy2011blackwell} and the summary by \citet{perdomo2025defense} which highlight the versatility of online learning algorithms via Blackwell approachability (respectively defensive forecasting) for a large classes of evaluation metrics treated in this work. In particular, \citep{perdomo2025defense} makes a related point to ours, in emphasizing the \emph{defining} role of the evaluation for the forecasts.} Even though the evaluation-protocol is motivated by game-theoretic probability and statistics \citep{ramdas2023game}, we do not give claims about equilibria or convergence.

\paragraph{Helpful Untruth}
The evaluation protocol is not meant to be \emph{real}, in the sense that the protocol is actually carried out. Instead, the evaluation protocol is a fictive play, a ``helpful untruth'' \citep{vaihinger1911philosophie}. By proceeding ``as if'' there were gamblers who play to test forecasts, we \emph{re}cover existing evaluation metrics and \emph{dis}cover their structural relationships.\footnote{In this regard, we follow the arguments by \citet{vaihinger1911philosophie} and \citet{appiah2017if} that deliberately wrong assumptions can help to deepen our understanding of the world.}. For instance, we will equip the gamblers with knowledge about the actual outcomes, knowledge which the forecaster might not have. Clearly, this cannot be true in an actual evaluation scenarios, but it helps to understand what different evaluation metrics do.

\section{Formal Setup}
\label{formal setup}
Let $\Y$ be a compact and metrizable topological space. We call $\Y$ the \emph{outcome set}. Let $\Sigma(\Y)$ be the Borel-$\sigma$-algebra on $\Y$. The set of signed measures with bounded variation on $\Sigma(\Y)$ is denoted as $\ca(\Y)$. The set of continuous real-valued functions on $\Y$ is denoted $\CY$. Note that every such function is measurable. Furthermore, every such function is integrable with respect to a measure $\phi \in \ca(\Y)$, which defines a bilinear mapping,
\begin{align*}
    \langle \cdot, \cdot \rangle \colon \ca(\Y) \times \CY \to \reals ; \quad \langle \phi, f \rangle \coloneqq \int fd\phi.
\end{align*}
The space $\CY$ (respectively $\ca(\Y)$) can be normed by the supremum norm $\| f \|_\infty \mapsto \sup_{y \in \Y} f(y)$ (respectively total variation). The space $\ca(\Y)$ is isomorphic to the norm dual of $\CY$, \ie, every linear continuous mapping $\CY \to \reals$ can be written as $f \mapsto \langle \phi, f \rangle$ \citep[Theorem 14.15]{aliprantis2006infinite}. Hence, the two spaces $\CY$ and $\ca(\Y)$ together with the bilinear mapping form a dual pairing \citep[Definition 5.90]{aliprantis2006infinite}.

The normed space $\CY$ (respectively $\ca(\Y)$) can be re-topologized in $\pqtopology$ topology, \ie, the topology which makes all evaluation functionals continuous. An evaluation function is defined as $\phi^* \colon \CY \rightarrow \reals$, $f \mapsto \langle \phi, f \rangle$, for some $\phi \in \ca(\Y)$. Analogously, the space $\ca(\Y)$ allows for the $\qptopology$ topology, \ie, the topology which makes all evaluation functionals, $f^* \colon \ca(\Y) \rightarrow \reals$, $\phi \mapsto \langle \phi, f \rangle$ for $f \in \CY$, continuous. The topologies $\pqtopology$ and $\qptopology$ are weak topologies in the sense of \citep[\p 758]{schechter1997handbook}.

The norm-topology and the weak-topology relate to each other as summarized in the following lemma. Set containment of topologies can be understood as fine-graining or strengthening a topology. If topology $\mathcal{T}_1$ is contained in another topology $\mathcal{T}_2$, then all open sets of $\mathcal{T}_1$ are open sets in $\mathcal{T}_2$ \citep[\p 24]{aliprantis2006infinite}.
\begin{lemma}[A Hierarchy of Topologies on $\CY$]
\label{lemma:A Hierarchy of Topologies on Lp}
    Let $\CY$ be a Banach space defined by the supremum norm $\| \cdot \|_\infty$ and $\ca(\Y)$ the paired space. We have the following containment,\footnote{We denote the topology induced by the supremum norm as $\sigma(\| \cdot \|_\infty)$.}
    \begin{align*}
        \pqtopology \subseteq \sigma(\| \cdot \|_\infty).
    \end{align*}
    If $S \subseteq \CY$ is convex and closed with respect to $\sigma(\| \cdot \|_\infty)$, then it is closed with respect to $\pqtopology$.
\end{lemma}
\begin{proof}
    The set containment is a special case of \citep[28.13.b]{schechter1997handbook}. The second statement follows from \citep[Theorem 28.14.a]{schechter1997handbook}, since $\CY$ with the norm-topology is a locally convex topological vector space \citep[\p 688]{schechter1997handbook}.
\end{proof}

We refer to the set of all non-positive gambles
\begin{align*}
    \CY_{\le 0} \coloneqq \{ f \in \CY \colon f \le 0\} = \{ f \in \CY \colon \sup f \le 0\},
\end{align*}
\ie, that is the negative orthant. For notational convenience we introduce the following shortcut when $A$ is a subset of $\CY$ or $\ca(\Y)$,
\begin{align*}
    \reals_{\ge 0} A \coloneqq \{ r a\colon r \in \reals_{\ge 0}, a \in A\},
\end{align*}
which is called the \emph{cone} generated by $A$. It is closed under multiplication with a positive scalar. If $A$ is convex, \ie, closed under convex combinations, then $\reals_{\ge 0} A$ is a \emph{convex cone} (Lemma~\ref{lemma:conic hull}).

The subset of probability measures, \ie, $\phi \in \ca(\Y)$ such that $\phi(A) \ge 0$ for all $A \in \Sigma(\Y)$ and $\mu(\Y) = 1$, on $\Sigma(\Y)$ is denoted as $\Delta(\Y) \subseteq \ca(\Y)$. If $\phi \in \Delta(\Y)$, we define the expectation operator,
\begin{align*}
    \mathbb{E}_\phi[f] \coloneqq\langle \phi, f \rangle.
\end{align*}
For elements in $\ca(\Y)$ we use, by default, greek letters, $\phi, \psi \in \ca(\Y)$. A special case of a probability distribution is the Dirac-distribution on an element $y \in \Y$ denoted as $\dirac(y) \in \Delta(\Y)$, which gives $\mathbb{E}_{\dirac(y)}[f] = f(y)$.
For elements in $\CY$, termed ``gambles'', we use latin letters $f,g \in \CY$.

The setup we chose, \ie, the pairing of $\CY$ and $\ca(\Y)$ spaces, is not the most general framework in which our statements hold. We conjecture that a more general pairing of locally convex topological vector spaces would lead to the same results, given meaningful definitions for properties of distributions exist. The reason for this is that the crucial Bipolar Theorem~\ref{prop:properties of polar set - bipolar theorem} holds in those more general cases. However, since those general spaces might include the often unfamiliar finitely additive probability measures we decided to simplify statements by restricting ourselves to the $\CY$ and $\ca(\Y)$ pairing. On the other hand, we need to work with abstract Banach spaces, since we do \emph{not} want to restrict our statements to hold for finite $\Y$. This way our setup encompass regression tasks. If $\Y$ would be finite the topologies of the Banach spaces would collapse to the Euclidean topology and the convexity argument could be carried out in finite dimensional Euclidean spaces. This step mainly simplifies the topological parts of the proofs, \ie, closure of sets, but not the duality statements.

We have summarized the most important definitions used in this paper in Table~\ref{tab:notation}. Not all of the denoted concepts have already been introduced, but will be introduced subsequently.
\newpage
\begin{table}[t]
\centering
    \begin{tabular}{r|l}
        \makecell[cr]{Set of probability\\ distributions on $\Y$} & $\Delta(\Y) \coloneqq \{ \phi \in \ca(\Y) \colon \phi \ge 0, \phi(\Y) = 1\}$\\[0.4 cm]
        \makecell[cr]{Set of continuous \\real-valued functions on $\Y$} & $\CY$\\[0.4 cm]
        Forecasting set & $\forecastingset \subseteq \Delta(\Y)$\\[0.2 cm]
        Domain of property & $\Q \subseteq \Delta(\Y)$, $\qptopology$-closed and convex\\[0.2 cm]
        Property of distribution & $\Gamma \colon \Q \rightarrow 2^\propspace$\\[0.3 cm]
        Level set of property & $\Gamma^{-1}(\gamma) \coloneqq \{ \phi \in \Q \colon \Gamma(\phi) = \gamma\}$, Definition~\ref{def:level set}\\[0.3 cm]
        \makecell[cr]{Consistent scoring\\ function for property $\Gamma$} & \makecell[cl]{$\loss \colon \Y \times \propspace \rightarrow \reals$ \st $\Gamma(\phi) = \argmin_{c\in \propspace}\mathbb{E}_\phi [\loss_c]$,\\ short $\loss_\gamma \colon y \mapsto \loss(\gamma, y)$, Definition~\ref{def:scoring function}}\\[0.4 cm]
        \makecell[cr]{Identification function\\ for property $\Gamma$} & \makecell[cl]{$\ident \colon \Y \times \propspace \rightarrow \reals$ \st $\mathbb{E}_\phi [\ident_\gamma] = 0 \Leftrightarrow \Gamma(\phi) = \gamma$,\\ short $\ident_\gamma \colon y \mapsto \ident(\gamma, y)$, Definition~\ref{def:identification function}}\\[0.4 cm]

        Credal set & closed and convex $\forecastingset \subseteq \Delta(\Y)$, Definition~\ref{def:credal set}\\[0.2 cm]
        Forecaster aligned to property $\Gamma$ & $\forecaster$, for every $t \in T \coloneqq \{ 1,\ldots, n\}$, $\forecaster_t \in \propspace$, Definition~\ref{def:evaluation protocol}\\[0.2 cm]
        Set of gamblers with index set $\indexsetgamblers$& $\sceptic^\indexsetgamblers \coloneqq \{\sceptic^i \colon i \in \indexsetgamblers\}$, Definition~\ref{def:evaluation protocol}\\[0.2 cm]
        $i$th gambler & $\sceptic^i$ for $i \in \indexsetgamblers$\\[0.2 cm]
        Gamble at instance $t$ of $i$th gambler & $\sceptic^i_t \in \CY$ \\[0.2cm]

        Generic gamble & $g \in \CY$\\[0.2cm]
        Nature & $\nature$, for every $t \in T \coloneqq \{ 1,\ldots, n\}$, $\nature_t \in \Y$, Definition~\ref{def:evaluation protocol}\\[0.2 cm]
        Capital of $i$th gambler $\sceptic^i$ & $\capital(\sceptic^i)$, Definition~\ref{def:evaluation protocol}\\[0.2 cm]

        Available gamble & $\gamble \in \CY$ such that $\sup_{\phi \in \Gamma^{-1}(\forecaster)} \mathbb{E}_\phi[\gamble] \le 0$, Definition~\ref{def:fair gambler and availability}\\[0.2 cm]
        Set of available gambles & $\offer_{\Gamma^{-1}(\forecaster)}$, Definition~\ref{def:fair gambler and availability} and $\offer_\forecastingset$, Theorem~\ref{thm:Available Gambles Form Offer}\\[0.2 cm]
        Offer & \makecell[cl]{$\offer \subseteq \CY$ additive, positive homogeneous\\coherent, default available and closed, Definition~\ref{def:offer}}\\[0.4 cm]

        (Unscaled) regret gambles & $\UnscaledRegretGambles_\gamma \coloneqq \{ \loss_\gamma - \loss_c \colon c \in \propspace \}$, Equation~\eqref{def:UnscaledRegretGambles}\\[0.2 cm]
        Scaled regret gambles & $\ScaledRegretGambles_\gamma \coloneqq \{ \beta(\loss_\gamma - \loss_c ) \colon \beta \in \reals_{\ge 0}, c \in \propspace \}$, Equation~\eqref{def:ScaledRegretGambles}\\[0.2 cm]
        Calibration gambles & $\CalibrationGambles_\gamma \coloneqq \{ \alpha \ident_\gamma \colon \alpha \in \reals \}$, Equation~\eqref{def:CalibrationGambles}\\[0.2 cm]
        
        \makecell[cr]{Superprediction set for\\ scoring function $\loss$} & $\spr(\scoring) \coloneqq \{ \gamble \in \CY \colon \exists c \in \propspace, \scoring_c \le \gamble \}$, Definition~\ref{def:superprediction set}
    \end{tabular}
    \caption{Summary of important definitions.}
    \label{tab:notation}
\end{table}


\section{The Evaluation Protocol}
\label{sec:The Protocol}
\begin{figure}
    \centering
    \def\svgwidth{0.85\columnwidth}
    {\footnotesize
    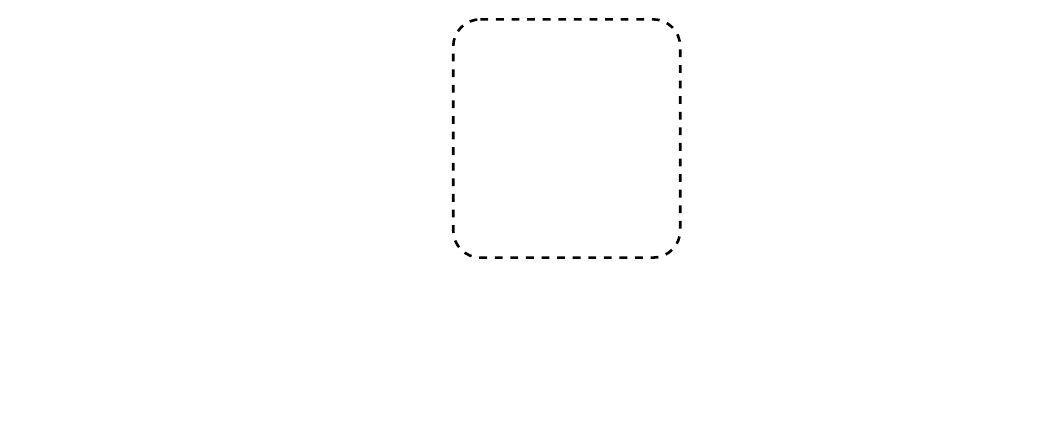
    }
    \caption{Evaluation Protocol following Definition~\ref{def:evaluation protocol} -- The Game is played in (potentially concurrent) rounds, for every $t\in T$. In a single round, first Forecaster provides a forecast aligned to a property $\Gamma$, then the $i$th gambler reacts with a gamble, and finally Nature reveals an outcome. The realized values of the gambles are aggregated along the axis of all played rounds.}
    \label{fig:the game}
\end{figure}

The \emph{evaluation protocol} is at the core of this study. We first give an intuitive introduction of it, which we formalize in Definition~\ref{def:evaluation protocol}. The evaluation protocol is illustrated in Figure~\ref{fig:the game}. Then, we shortly discuss the definition of Forecaster, before we characterize which evaluation metrics can be represented by the evaluation protocol. It turns out that further restrictions need to be set on the gamblers to demarcate reasonable representable evaluation metrics. This will lead us to the characterization of available gambles in the next section.

The evaluation protocol is conducted in (potentially concurrent) rounds, indexed by the round parameter $T \coloneqq \{1,\ldots, n \}, n \in \naturals$. It mainly revolves around three entities: Nature, Forecaster and a set of gamblers.\footnote{We henceforth use a first capital letter when referring to a gambler, a forecaster or a nature, when using the word as name.} Roughly summarized, the protocol does ``as if'' Forecaster forecasts the next outcome of Nature, Gambler gambles against this forecast, and lastly Nature reveals an outcome. The gamble is evaluated on the outcome. Finally, the gambles' values are aggregated to Gambler's capital. Semantically, Forecaster's goal is to keep Gambler's capital low, while Gambler's job is to cast doubt on the forecast and increase its own gain. We generally assume that positive outcome values of gambles are ``good'' and negative are ``bad''. Hence, we take on Gambler's perspective.
\begin{definition}[Evaluation Protocol]
\label{def:evaluation protocol}
We call $(\Y, \Gamma, T, \forecaster, \sceptic^I, \nature)$ \emph{evaluation protocol}, where
\begin{enumerate}[(a)]
    \item $\Y$ is an outcome space.
    \item $\Gamma \colon \Q \rightarrow 2^\propspace$ is a property (Section~\ref{the forecasts}) and some $\qptopology$-closed, convex set of distributions $\Q \subseteq \Delta(\Y)$ and $\propspace$ some set.
    \item $T \coloneqq \{1,\ldots, n \}, n \in \naturals$ is a set of instances.
    \item $\forecaster \colon T \rightarrow \propspace$ is a forecaster.
    \item $\nature \colon T \rightarrow \Y$ is a nature.
    \item $\sceptic^\indexsetgamblers \coloneqq \{ \sceptic^i \colon i \in \indexsetgamblers\}$ for some finite index set $\indexsetgamblers$, where $\sceptic^i \colon T \rightarrow \CY$.
\end{enumerate}
For every $t \in T$,
    \begin{enumerate}[1)]
        \item Forecaster forecasts property $\Gamma$, \ie, $\forecaster_t \in \propspace$.
        \item informed by the forecast, for every $i \in \indexsetgamblers$, the gambler $\sceptic^i$ plays the gamble $\sceptic^i_t \in \CY$.
        \item Nature reveals an outcome $\nature_t \in \Y$.
    \end{enumerate}
    The \emph{capital}\footnote{For the sake of readability, we do not make explicit the dependencies of the exact type of the capital $\capital(\sceptic^i)$, which clearly depends on more than only the gambler.
} of the $i$th gambler is the average realized value of all played gambles, \ie, for all $i \in \indexsetgamblers$,
    \begin{align*}
        \capital(\sceptic^i) \coloneqq \frac{1}{|T|} \sum_{t \in T} \sceptic^i_t(\nature_t).
    \end{align*}
\end{definition}
Note that we explicitly make a distinction between generic gambles $g \in \CY$, and gambles played by the $i$th gambler $\sceptic^i$ at instance $t$, which is denoted $\sceptic^i_t \in \CY$. When there is no index $i$, then the gambler is denoted $\sceptic$, and its gamble at instance $t$, $\sceptic_t$.

\subsection{Forecaster Forecasts a Property Value}
\label{the forecasts}
What do we mean by ``Forecaster forecasts property $\Gamma$''?
We were motivated to use the construction via $\Gamma$ because there are plenty of types of forecasts in machine learning.
Forecasts in machine learning include full distribution estimates over classes (class probability estimation), means (regression), quantiles (quantile regression), and many more. All those forecasts share that they are properties of a distribution on the outcome set.

Let $\Q$ always denote a $\qptopology$-closed, convex set of distributions $\Q \subseteq \Delta(\Y)$ The set $\Q$ is the domain of a \emph{property} $\Gamma \colon \Q \rightarrow 2^\propspace$\footnote{We denote the power set of a set $A$ as $2^A$.} which is a map to some subset of all possible property values $\propspace$, \eg, the mean, the median, the mode. We follow the exposition of \citet[\S 2.2]{gneiting2011making} allowing the property to map to \emph{sets} of property values.\footnote{Thereby, we allow for an elicitable property with non-strict consistent scoring functions (\cf Definition~\ref{def:scoring function}). For example, both possible elements in $\{ 0,1\}$ are the median of a uniform Bernoulli distribution on this set.} The property value set $\propspace$ is deliberately general, as it could be equal to the real line $\reals$, \eg, for means (\cf \citep{gneiting2011making}), to a finite-dimensional vector space $\reals^d$ (\cf \citep{frongillo2015vector}), a discrete set $\{ 1,\ldots, n\}$ (\cf \citep{lambert_elicitation_2019}) or the set of all distributions $\Delta(\Y)$ (\cf \citep{gneiting2007strictly, williamson2022geometry}).

We say that a forecaster \emph{forecasts a property $\Gamma$} or \emph{is aligned to property $\Gamma$}, if the forecast $\forecaster_t \in \propspace$ provides an estimate of the property $\Gamma$ for every $t \in T$ \citep{gneiting2011making}.\footnote{Within the assumption that the forecaster exactly provides its own estimate of the property, we hide the fact that the forecaster can freely choose the predicted value among the set of possible property values. In principle, the forecaster could be bound to a certain hypothesis class which disallows certain predictions.} What matters to us is that a forecasted property value corresponds to a set of probability distributions which possess the announced property value. This is formalized using the pre-image of a property value. 
\begin{definition}[Level Set of Property]
\label{def:level set}
    Let $\gamma \in \propspace$. The \emph{level set} of a property $\Gamma \colon \Q \to 2^{\propspace}$ is defined as
    \begin{align*}
        \Gamma^{-1}(\gamma) \coloneqq \Gamma^{-1}(\{ \gamma\}) = \{ \phi \in \Q \colon  \gamma \in \Gamma(\phi) \}.
    \end{align*}
    We assume level sets to be non-empty in the following, \ie, $\Gamma(\Q) = \propspace$.
\end{definition}
Hence, for every forecast $\gamma \in \propspace$ there is a corresponding, implicit, set of probability distributions $\Gamma^{-1}(\gamma)\subseteq \Delta(\Y)$. We call such a set \emph{forecasting set}. By the generality of the definition of a property, every non-empty set $\forecastingset \subseteq \Delta(\Y)$ is a forecasting set for some property $\Gamma$. If the property $\Gamma$ is not of interest, we neglect to mention the property and directly refer to the forecasting set $\forecastingset$.

Concluding, a forecaster forecasts a property $\Gamma$ if the forecasts provide an estimate of the property $\Gamma$. This estimate is in the set $\propspace$. Equivalently, one can understand a forecaster as forecasting a set of probability distributions in $\Delta(\Y)$ which share a property value. This abstraction step from a single probability distribution to a set of probability distributions will be shown to be necessary to fully capture the variety of evaluation metrics and linking calibration and regret.

\subsection{The Protocol is Based on a Game-Theoretic Probability Protocol}
\label{relation to existing forecasting protocols}
Our evaluation protocol is closely related to Protocol 6.11 and 6.12 in \citep{shafer2019game}. However, in contrast to theirs:
\begin{description}
    \item[Modifications] Their forecasts are closed, convex sets of probability distributions. Our forecasts are aligned to properties. They aggregate the capital via addition. We aggregate the capital via average\footnote{Actually, this was mainly a choice of convenience as it makes some of the recovery results (Section~\ref{recovery of existing evaluation notions}) more streamlined.}. Furthermore, \citet{shafer2019game} leave the gambler with a restricted set of options (called ``offer''). So far, we did not incorporate this constraint. However, we will introduce this additional constraint in Section~\ref{the availability criterion}.
    \item[Extensions] They consider a single gambler, while we treat several gamblers at once. Where we make no assumptions on the exchange of information between Forecaster, the gamblers and Nature, they assume that the involved agents have access to the past moves of all agents.
    \item[Purpose] We are focusing on different conditions put on a gambler $\sceptic$ in order to recover and understand existing evaluation metrics. They aim for a recovery of measure-theoretic probability statements via game-theoretic tools.
\end{description}

\subsection{More Than Single-Instance-Based Evaluation Metrics can be Represented by the Protocol}
\label{representability of evaluation metrics}
Having introduced the evaluation protocol, a first simple question is which evaluation metrics are representable via the evaluation protocol?

We understand a general \emph{evaluation metric} as a map, for some finite $T \coloneqq \{ 1,\ldots, n\}$, from outcomes and corresponding forecasts to some real number, $\Y^T \times \propspace^T \to \reals$.  We call an evaluation metric \emph{representable in an evaluation protocol} if there exists $(\Y, \Gamma, T, \forecaster, \sceptic^\indexsetgamblers, \nature)$ following Definition~\ref{def:evaluation protocol}, such that some aggregation $\aggregation \colon \reals^{|\indexsetgamblers|} \to \reals$ of the capitals of the gamblers is equal to the evaluation metric. In other words, the representable evaluation metric is ``as if'' there were gamblers who aggregate their capital.\footnote{We do not delve deeper in the discussion for reasonable aggregation functionals, see \citep{frohlich2022risk, holtgen2023richness, cabrera2024lossaggregation} for the discussion of aggregations for evaluation metrics.} The set of representable evaluation metrics is readily characterized. Essentially, the argument is the (un)currying of the the gamblers' functions.
\begin{proposition}[Characterization of Representability]
    \label{prop:Recoverability of Evaluation Metric}
    Let $T \coloneqq \{ 1, \ldots, n\}$, $\Y$ be an outcome set and $\propspace$ a set of property values.
    An evaluation metric $m \colon \Y^T \times \propspace^T \to \reals$ is \emph{representable in an evaluation protocol}, if and only if, there exists a set of functions $f^i_t \colon \Y \times \propspace \to \reals$ indexed by $t \in T$ and $i \in I$ for some finite $I$ and an aggregation $\aggregation \colon \reals^{|I|} \to \reals$, such that for all $\nature \in \Y^T$ and $\forecaster \in \propspace^T$,
    \begin{align*}
        m((\nature_t)_{t \in T}, (\forecaster_t)_{t \in T}) = \aggregation\left[\left(\frac{1}{|T|} \sum_{t \in T}f^i_t(\nature_t,\forecaster_t)\right)_{i \in I}\right].
    \end{align*}
\end{proposition}
The proof is deferred to Appendix~\ref{appendix:representability of evaluation metrics}. In particular, every \emph{single-instance-based evaluation metric} is representable (Appendix~\ref{appendix:representability of evaluation metrics}, Proposition~\ref{prop:single-instance based evaluation metrics are representable of Evaluation Metric}). A \emph{single-instance-based evaluation metric} (\cf \citep{sandroni2003reproducible,fortnow2009complexity}) is a metric which is an aggregation of comparisons between forecast and outcome for each instance. 

Despite the generality of evaluation metrics which are representable by the protocol, there exist evaluation metrics which do not fit into this form. In \citep{sandroni2003reproducible} the authors consider general evaluation metrics, what they call ``tests'', which have access to all forecasts and outcomes at once and can compare and analyze predictions and outcomes across rounds. Furthermore, the gamblers only gamble against the actual forecasts for the outcomes and do not have the power to ask for forecasts for unrealized outcomes.\footnote{In this regard, the forecast evaluation is \emph{prequential} \citep{dawid1999prequential}, respectively has no oracle-access \citep{dwork2021outcome}.} In oracle-access outcome indistinguishability (respectively code-access outcome indistinguishability) the ``distinguisher'', which are (computationally implementable) test functions, have access to predictions made for unobserved instances (respectively access to the description of the circuit which produced the predictions) \citep{dwork2021outcome}. 
The evaluation protocol is a powerful, but not all-encompassing framework for evaluation metrics.


\subsection{Fairness of the Gambler is a Minimal Criterion for Reasonable Evaluation Metrics}
\label{the availability criterion}
It is easy to see that the capital $\capital$, and hence the aggregated capitals, is an essentially meaningless quantity so far. For instance, a gambler can simply provide a constant positive function as gamble, which does not provide any information about the quality of the forecasts on the respective outcomes. For this reason, let us introduce a first, elementary, minimal criterion to which the gamblers have to adhere.

We define a \emph{fair} gambler as one which only plays gambles \emph{available} to it. Intuitively, a gamble is available if and only if the forecaster (literally) expects the gamble to incur a loss to the gambler for all probabilities in the forecasting set. In other words, Forecaster allows Gambler to play all gambles which Forecaster assumes to be not desirable.\footnote{\emph{Availability} is negative ``desirability''\citep[\p 131]{shafer2019game}, a concept used in the literature on imprecise probability \citep{walley2000towards}. Desirable gambles are all gambles for which the forecaster assumes that playing those will not lead to a loss, \ie, choosing a desirable gamble and evaluating this gamble at an outcome of Nature will lead to a ``benign'' outcome.} We borrow the concept of availability from \citep[Protocol 6.11 / 6.12]{shafer2019game}.
\begin{definition}[Fair Gambler, Availability, Set of Available Gambles]
    \label{def:fair gambler and availability}
    In an evaluation protocol $(\Y, \Gamma, T, \forecaster, \sceptic^\indexsetgamblers, \nature)$, the $i$th ($i  \in \indexsetgamblers$) gambler $\sceptic^i$ is \emph{fair}, if and only if, for all $t \in T$, $\sceptic^i_t$ is \emph{available}, \ie,
    \begin{align*}
        \sup_{\phi \in \Gamma^{-1}(\forecaster_t)} \mathbb{E}_{\phi}[\sceptic^i_t] \le 0.
    \end{align*}
    For the \emph{set of available gambles} at time $t \in T$ we write,
    \begin{align*}
        \offer_{\Gamma^{-1}(\forecaster_t)} \coloneqq \left\{ \gamble \in \CY \colon \sup_{\phi \in \Gamma^{-1}(\forecaster_t)} \mathbb{E}_\phi[\gamble] \le 0 \right\}
    \end{align*}
\end{definition}
A gambler which violates the availability criterion has an unfair advantage in increasing its capital against Forecaster. Hence, if we interpret the gambler's capital as an evaluation metric, this metric would not serve the purpose to guarantee ``good'' forecasts, since the resulting capital would most likely not relate to a quality of the forecasts.

The availability criterion is a necessary, but not sufficient meta-criterion for reasonable forecast evaluators in our protocol. Assuming that the forecaster is ``clairvoyant'', \ie, it does know the next outcome of Nature, we would expect any reasonable evaluation metric to state that the forecasts are perfect. The following proposition shows that this is equivalent to fairness of the gambler.
\begin{proposition}[Sanity Check]
\label{proposition:sanity check}
    Let $(\Y, \Gamma, T, \forecaster, \sceptic^\indexsetgamblers, \nature)$ be an evaluation protocol. Let $\forecaster$ be ``clairvoyant'', \ie, $\forecaster_t \coloneqq \Gamma(\dirac(\nature_t))$\footnote{We assume that $\dirac(\nature_t) \in \Q$ for all $t \in T$ to guarantee that the forecaster is well-defined.}. The capital of any gambler $\sceptic^i$ with $i \in \indexsetgamblers$ is lower or equal to zero, \ie, $\capital(\sceptic^i) \le 0$, if and only if, the gambler is fair.
\end{proposition}
\begin{proof}
    Fix any gambler $\sceptic^i$ for some $i \in \indexsetgamblers$.
    We first show that the capital of the gambler is upper bounded, if the gambler only plays available gambles.
    The set of gambles available to the gambler is strongly restricted: for any $\sceptic^i_t \in \offer_{\Gamma^{-1}(\forecaster_t)}$ it holds,
    $\mathbb{E}_{\dirac(\nature_t)}[\sceptic^i_t] \le 0$, which is equivalent to $\sceptic^i_t(\nature_t) \le 0$.
    Hence,
    \begin{align*}
        \capital(\sceptic^i) = \frac{1}{|T|} \sum_{t \in T} \sceptic^i_t(\nature_t) \le 0.
    \end{align*}

    We show the reverse direction by contraposition. Assume that the capital of the $i$th gambler is larger than zero. Then there exists a gamble $\sceptic^i_t$, such that $\sceptic^i_t(\nature_t) > 0$. However, this implies that $\mathbb{E}_{\dirac(\nature_t)}[\sceptic^i_t] > 0$, from which follows that $\sceptic^i_t \notin \offer_{\Gamma^{-1}(\forecaster_t)}$.
\end{proof}
Fairness of the gambler can thus be interpreted as Type-I error freeness. The null hypothesis, \ie, the forecast, cannot be rejected via available gambles, if the null hypothesis is true. In fact, this interpretation is well-founded, as availability is tightly related to game-theoretic hypothesis testing (Section~\ref{Related Work - Testing Forecasts}).

However, a fair gambler does not necessarily guarantee the quality of forecasts either. A forecast can still be uninformative, even though several fair gamblers didn't obtain a large capital. For instance, the forecaster who predicts a property whose level set comprises of all distributions on $\Y$, \ie, a vacuous forecast, will enforce any fair gambler to play non-positive gambles (Lemma~\ref{lemma:vacuous forecasts only make non-positive gambles available}). Hence, the gambler's capital stays non-positive $\capital \le 0$. However, the forecasts might not necessarily be ``good'', in terms of describing the outcomes by Nature appropriately.

The difficulty of evaluating arbitrary set-valued probabilistic forecasts is a known issue and part of a longer debate on scoring rules for imprecise probabilities (\cf Appendix~\ref{appendix: Availability Criterion for Imprecise Forecasts}). In conclusion, availability is necessary, but not sufficient to guarantee ``good'' forecasts. It is a meta-criterion for reasonable evaluation metrics representable by our evaluation protocol.

\section{Characterization of Available Gambles}
\label{sec:Characterization of Available Gambles}
The evaluation protocol is a general tool to describe evaluation metrics (Section~\ref{representability of evaluation metrics}) based on protocols from game-theoretic probability (Section~\ref{relation to existing forecasting protocols}). The \emph{fairness} criterion on gambler is a minimal sanity check, that every reasonable evaluation metric should fulfill (Section~\ref{the availability criterion}). However, it remains unclear whether existing evaluation metrics are representable and fulfill the fairness criterion? Towards answering this question, we will first respond to:
\textbf{What is the potential of fair gamblers to disprove $\Gamma$-aligned forecasters? Which gambles \emph{can} be played?}

In other words, we ask for a characterization of the set of available gambles.
For general $\Gamma$ this question can only be answered to a limited extent. Essentially, it is possible to argue that the set of available gambles is equal for some level set of $\Gamma$ and the closed, convex hull of the very same level set (Section~\ref{From Forecasting Sets to Available Gambles and Back}). For a more exhaustive description we have to restrict ourselves to certain properties $\Gamma$ which fulfill additional constraints.

As we will argue, \emph{elicitable} (respectively \emph{identifiable}) $\Gamma$ do not only allow for clean characterizations of the corresponding sets of available gambles, but as well form the two pillars on which regret-type (respectively calibration-type) evaluation metrics are based. The statements and relationships are summarized in Figure~\ref{fig: this paper as a drawing}.

For instance, we will show that \emph{(unscaled) regret gambles} such as,
\begin{align}
\label{def:UnscaledRegretGambles - Brier score}
    \{ \nature \mapsto (\nature - \gamma)^2 - (\nature - c)^2 \colon c \in \propspace \},
\end{align}
are available, if the property $\Gamma$, which is forecasted, is \emph{elicited} by the loss function $ \ell \colon \nature \mapsto (\nature - c)^2$ (\cf Section~\ref{introduction}). We are interested in \emph{regret gambles} as they mimic the structure of regret-type evaluation metrics such as external or swap regret (Section~\ref{From Gambles to Regret}).
And as it turns out, what elicitability is for regret-type evaluation metrics, identifiability is for calibration-type evaluation metrics \citep{noarov2023scope}. Let us formally introduce those concepts.

\subsection{Elicitable and Identifiable Properties}
\label{elicitable and identifiable properties}
We have argued that machine learning systems generally forecast properties (denoted $\Gamma$), \ie, mappings from a set of distributions to subsets of property values. We have given examples, such as the mean, the median, or a full class probability estimate. Those specific properties are furthermore \emph{elicitable} (respectively \emph{identifiable}).
Elicitability guarantees that the property is the solution of an expected loss minimization problem \citep{osband1985providing}. A property which is identifiable can be characterized by a so-called identification function \citep{osband1985providing, gneiting2011making}.
\begin{definition}[Elicitability and Consistent Scoring Function]
\label{def:scoring function}
    Let $\Gamma \colon \Q \to 2^{\propspace}$ be a property. A function $\scoring \colon \Y \times \propspace \rightarrow \reals$ is called a \emph{consistent scoring function} for $\Gamma$ if, for all $\gamma \in \propspace$, $\scoring_\gamma \colon y \mapsto \scoring(y, \gamma) \in \CY$, and 
    \begin{align*}
        \mathbb{E}_{\phi}[\scoring_\gamma] \le \mathbb{E}_{\phi}[\scoring_c]
    \end{align*}
    for all $\phi \in \Q, \gamma \in \Gamma(\phi), c\in \propspace$. We can equivalently write
    \begin{align*}
        \Gamma(\phi) \subseteq \argmin_{c \in \propspace} \mathbb{E}_\phi[\scoring_c].
    \end{align*}
    A property which has a consistent scoring function is called \emph{elicitable}.
\end{definition}
\begin{definition}[Identifiability and Identification Function]
    \label{def:identification function}
    Let $\Gamma \colon \Q \to 2^{\propspace}$ be a property. A function $\ident \colon \Y \times \propspace  \rightarrow \reals$ is called an \emph{identification function} of $\Gamma$ if, for all $\gamma \in \propspace$, $\ident_\gamma \colon y \mapsto \ident(y ,\gamma) \in \CY$, and
    \begin{align*}
        \mathbb{E}_{\phi}[\ident_\gamma] = 0 \Leftrightarrow \gamma \in \Gamma(\phi),
    \end{align*}
    for all $\phi \in \Q$. 
    A property which has an identification function is called \emph{identifiable}.
\end{definition}
The prototypical example of an elicitable and identifiable property is the mean. A corresponding scoring function is $\scoring(y, \gamma) = (y - \gamma)^2$. An identification function is $\ident(y, \gamma) = y - \gamma$. Another standard example is the $\tau$-pinball loss which elicits the $\tau$-quantile \citep{gneiting2011making}. Furthermore, proper scoring rules \citep{williamson2022geometry} are consistent scoring functions. They elicit the entire distribution \citep{savage1971elicitation}. The corresponding property is the identity function, \ie, $\Gamma_{id} \colon \Q \to 2^\Q, \phi \mapsto \{ \phi\}$.

When $\propspace = \reals$, elicitable properties are, neglecting technicalities, identifiable and vice-versa \citep{lambert2008eliciting, steinwart2014elicitation}. This is not necessarily true for more general $\propspace \subseteq \reals^d$ \citep{fissler2016higher}. It is a matter of simple computations to show that elicitable properties and identifiable properties have convex level sets. This result goes back at least to \citep{osband1985providing}.\footnote{For finite $\Y$ and finite $\propspace$ level sets are not only convex but are the result of a Voronoi-partition intersected with the simplex \citep{lambert_elicitation_2019}.} Furthermore, the corresponding level sets are closed. Hence, if the forecaster provides a property value $\forecaster \in \propspace$ for an elicitable or identifiable property $\Gamma$, then implicitly this corresponds to a closed, convex forecasting set $\Gamma^{-1}(\forecaster) \subseteq \Delta(\Y)$.\footnote{We call such sets \emph{credal sets} (Definition~\ref{def:credal set}) for reasons explained in Section~\ref{From Forecasting Sets to Available Gambles and Back}.}
\begin{proposition}[Convex and Closed Level Sets]
\label{prop:Identifiable or Elicitable Property Give Credal Level Sets}
    For all $\gamma \in \propspace$ the level set $\Gamma^{-1}(\gamma) \subseteq \Delta(\Y)$ of an elicitable or identifiable property $\Gamma$ is credal, \ie, convex and $\qptopology$-closed.
\end{proposition}
The proof of the statement is deferred to Appendix~\ref{appendix:elictiable or identifable properties credal sets}. The statement has been shown under a variety of differing technical setups. We provide a proof as we have not found any such statement for the dual pairing between $\ca(\Y)$ and $C(\Y)$. The proof idea is not novel. The details are our contribution.

\subsection{Warm-Up: Regret Gambles and Calibration Gambles are Available}
\label{Regret Gambles and Calibration Gambles are Available}
Elicitable properties are minimizers of loss functions. Identifiable properties are identified by a zero-condition. What do those characteristics of properties have to do with regret and calibration?

In the following warm-up, we shortly argue that \emph{(unscaled) regret gambles}, for some $\gamma \in \propspace$,
\begin{align}
\label{def:UnscaledRegretGambles}
    \UnscaledRegretGambles_\gamma \coloneqq \{ \loss_\gamma - \loss_c \colon c \in \propspace \}
\end{align}
and \emph{calibration gambles}, for some $\gamma \in \propspace$,
\begin{align}
\label{def:CalibrationGambles}
    \CalibrationGambles_\gamma \coloneqq \{ \alpha \ident_\gamma \colon \alpha \in \reals \},
\end{align}
are \emph{available} if the forecasted property $\Gamma$ is \emph{elicitable} (respectively \emph{identifiable}). Those structured gambles, \ie, unscaled regret gambles and calibration gambles, are the atoms of regret-type and calibration-type notions, as we will argue exhaustively in Section~\ref{The Fair, Restricted and Rational Gambler}. However, we will need a more detailed understanding of \emph{all} available gambles before we can fully recover used evaluation metrics.
\begin{proposition}[(Unscaled) Regret Gambles are Available]
\label{prop:regret gambles are available}
    Let $\Gamma \colon \Q \rightarrow 2^\propspace$ be an elicitable property with consistent scoring function $\scoring \colon \Y \times \propspace \rightarrow \reals$. For every $\gamma \in \propspace$ such that $\Gamma^{-1}(\gamma) \neq \emptyset$, $$\UnscaledRegretGambles_\gamma \subseteq \offer_{\Gamma^{-1}(\gamma)}.$$
\end{proposition}
\begin{proof}
    The statement is a direct consequence of the consistency of the scoring function $\scoring$. For every $\gamble \in \UnscaledRegretGambles_\gamma$, $\gamble = \loss_\gamma - \loss_c$ for some $c \in \propspace$, and so,
    \begin{align*}
        \sup_{\phi \in \Gamma^{-1}(\gamma)} \mathbb{E}_{\phi}[\gamble] = \sup_{\phi \in \Gamma^{-1}(\gamma)} \mathbb{E}_{\phi}[\loss_\gamma - \loss_c] = \sup_{\phi \in \Gamma^{-1}(\gamma)} \mathbb{E}_{\phi}[\loss_\gamma] - \mathbb{E}_{\phi}[\loss_c] \le 0,
    \end{align*}
    as $\mathbb{E}_{\phi}[\loss_\gamma] \le  \mathbb{E}_{\phi}[\loss_c]$ for all $\phi  \in \Gamma^{-1}(\gamma)$, see Definition~\ref{def:scoring function}.
\end{proof}
\begin{proposition}[Calibration Gambles are Available]
\label{prop:calibration gambles are available}
    Let $\Gamma \colon \Q \rightarrow 2^\propspace$ be an identifiable property with identification function $\ident \colon \Y \times \propspace \rightarrow \reals$. For every $\gamma \in \propspace$ such that $\Gamma^{-1}(\gamma) \neq \emptyset$, $$\CalibrationGambles_\gamma \subseteq \offer_{\Gamma^{-1}(\gamma)}.$$
\end{proposition}
\begin{proof}
    The statement is a consequence of the properties of the identification function $\ident$. For every $g \in \CalibrationGambles_\gamma$, $g = \alpha \ident_\gamma$ for some $\alpha \in \reals$, and so,
    \begin{align*}
        \sup_{\phi \in \Gamma^{-1}(\gamma)} \mathbb{E}_{\phi}[g] = \sup_{\phi \in \Gamma^{-1}(\gamma)} \mathbb{E}_{\phi}[\alpha \ident_\gamma] = \sup_{\phi \in \Gamma^{-1}(\gamma)} \alpha\mathbb{E}_{\phi}[\ident_\gamma] = 0,
    \end{align*}
    as $\mathbb{E}_{\phi}[\ident_\gamma] = 0$ for all $\phi  \in \Gamma^{-1}(\gamma)$, see Definition~\ref{def:identification function}.
\end{proof}
Hence, unscaled regret gambles and calibration gambles are in the arsenal of a fair gambler to discredit a forecaster aligned to an elicitable (respectively identifiable) property. Nevertheless, the boundaries of availability offer more freedom than the suggested gambles above. In fact, for forecasts corresponding to elicitable (respectively identifiable) properties we can characterize the full set of available gambles. This way we argue that in principle the gamblers can leverage regret-gambles or calibration-gambles equivalently, \ie, with the same ability to test forecasts, when the regret-gambles are allowed to be scaled.


\subsection{Characterizing Available Gambles of Forecasts of Elicitable Properties}
\label{characerizing available gambles of elicitable properties}
If the property $\Gamma$ is elicitable, then, so we argue, \emph{scaled regret gambles},
\begin{align}
    \label{def:ScaledRegretGambles}
    \ScaledRegretGambles_\gamma \coloneqq \{ \beta(\loss_\gamma - \loss_c ) \colon \beta \in \reals_{\ge 0}, c \in \propspace \},
\end{align}
dominate, up to approximation, all available gambles. 

In order to provide a full characterization of available gambles for forecasts aligned to elicitable properties we have to introduce the concept of a superprediction set. This concept originates from the literature on proper scoring rules \citep{kalnishkan2002mixability, kalnishkan2004criterion, dawid2007geometry}.
\begin{definition}[Superprediction Set]
\label{def:superprediction set}
Let $\scoring \colon \Y \times \propspace \rightarrow \reals$ be a scoring function. We call
\begin{align*}
    \spr(\scoring) \coloneqq \{ g \in \CY \colon \exists c \in \propspace, \scoring_c \le g \},
\end{align*}
the \emph{superprediction set} of $\scoring$.
\end{definition}
The superprediction set consists of all gambles which incur no less loss than for some $c \in \propspace$. Positive values have to be interpreted as ``bad'' here. Superprediction sets largely describe the behavior of loss functions \citep{williamson2022geometry}.

For our characterization we require convexity of the superprediction set. Convexity of the superprediction set is not a strong assumption for consistent scoring functions. For instance, every consistent scoring function induces a proper scoring rule, \eg, \citep[Theorem 3]{gneiting2011making}, which, in finite dimensions, has a surrogate proper scoring rule with a convex superprediction set with same conditional Bayes risk \citep[Section 3.4]{williamson2022geometry}. The proper scoring rule and its surrogate are equivalent almost everywhere \citep[Proposition 8]{williamson2016composite}. On the other hand, it is possible to directly convexify the superprediction set by taking the convex closure of $\propspace$ and extending the scope of $\loss$ (\cf ``randomized acts'' in Section 3.2 in \citep{gruenwald2004game}).

\begin{figure}[ht]
    \centering
    \includegraphics[width = 0.5\textwidth]{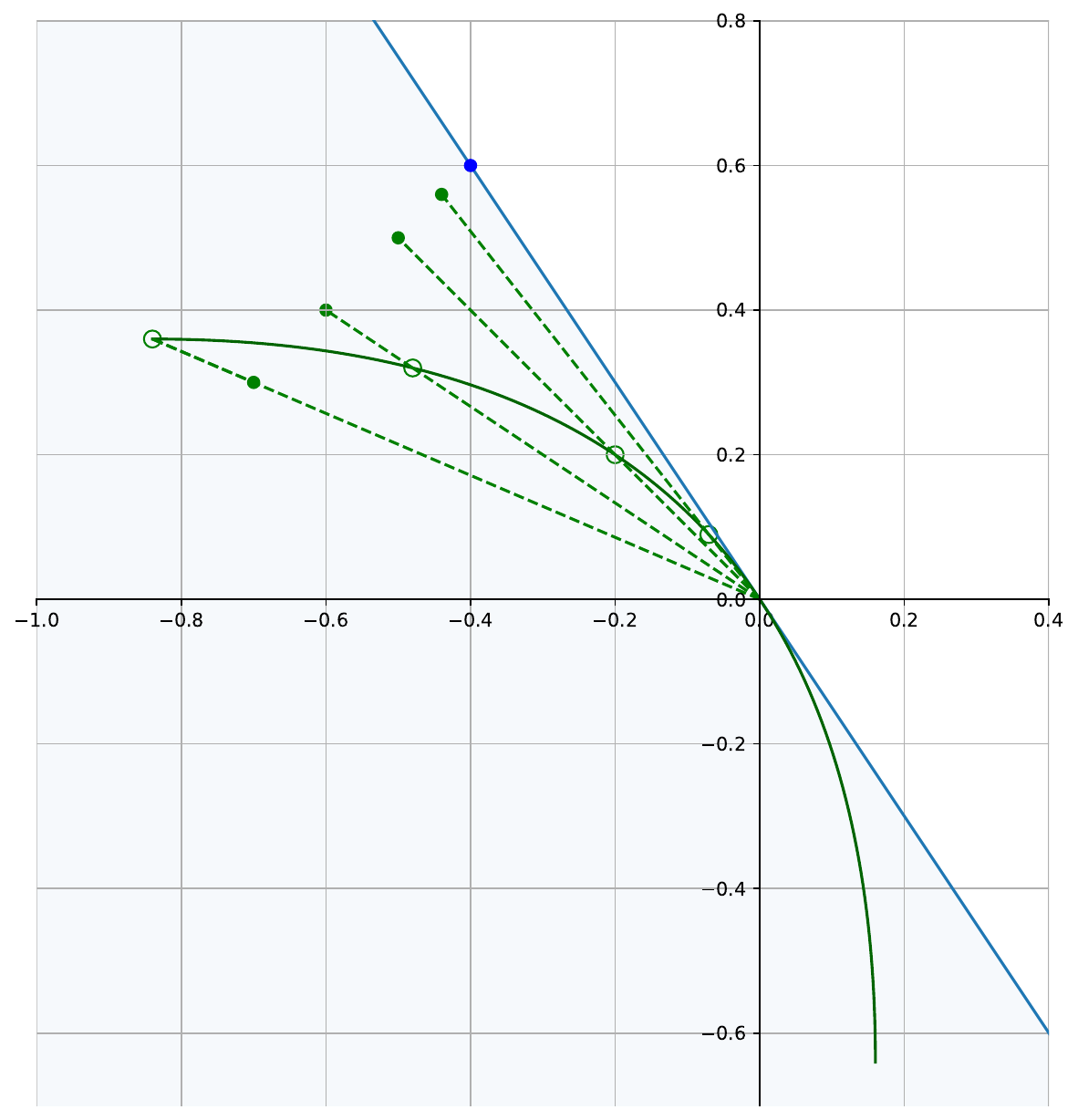}
    \caption{Available Gambles in the setup as described in Example~\ref{example:openness of set of regret gambles} -- We fix Forecaster to $\forecaster = 0.4$. The set of available gambles is shaded in light blue. The set of calibration gambles is given by the blue line. An exemplary available gamble $\sceptic = (-0.4, 0.6)$ is marked as a blue dot. The set of regret gambles for fixed $\beta = 1$ is drawn in green. An approximation of $\sceptic$ via rescaled regret gambles $\scoring_\epsilon$ as in Equation~\eqref{eq:approximating via scaled regret gambles} is shown in green. }
    \label{fig: available gambles in the binary prediction game}
\end{figure}

With this tool at hand we can provide a full characterization of the set of available gambles in the case that the property $\Gamma$ is elicitable. The proof is provided in Section~\ref{proof: characerizing available gambles of elicitable properties}. It requires the introduction of more concepts from the literature on imprecise probabilities (Section~\ref{From Forecasting Sets to Available Gambles and Back}).
\begin{theorem}[Available Gambles of Elicitable Property-Forecasts]
\label{thm:Characterization of Available Gambles in Protocol 2 with Elicitable Property}
    Let $\Gamma \colon \Q \rightarrow 2^\propspace$ be an elicitable property with scoring function $\scoring \colon \Y \times \propspace \rightarrow \reals$ which has a convex superprediction set. For a fixed $\gamma \in \propspace$ such that $\Gamma^{-1}(\gamma) \neq \emptyset$, we define
    \begin{equation}
        \label{eq:dominated by regret gambles}
        \mathcal{H}_{\scoring_\gamma} \coloneqq \cl \{ g \in \CY \colon g \le \beta (\scoring_\gamma - \scoring_c) \text{ for some } \beta \in \reals_{\ge 0}, c \in \propspace \}.
    \end{equation}
    Then, $$\mathcal{H}_{\scoring_\gamma} = \offer_{\Gamma^{-1}(\gamma)}.$$
\end{theorem}
Hence, the set of available gambles can be, up to approximation in $\pqtopology$-topology, expressed as all gambles which are upper bounded by some \emph{scaled regret gambles} (Equation~\eqref{def:ScaledRegretGambles}). It turns out that the closure is crucial here. Even in the finite dimensional case in a simple example, it is easy to show that $\{ g \in \CY \colon g \le \alpha (\scoring_\gamma - \scoring_c) \text{ for some } \alpha \in \reals_{\ge 0}, c \in \propspace \}$ is not closed.
\begin{example}
    \label{example:openness of set of regret gambles}
    Let us $\Y \coloneqq \{ 0,1\}$ and $\Gamma$ be the mean with consistent scoring function $\scoring(\nature, \gamma) \coloneqq (\nature - \gamma)^2$. In this simple setup, the mean identifies the distribution. For simplification we assume that Forecaster states $\forecaster = 0.4 $, \ie, the forecasting set is the singleton set with the distribution $(0.6, 0.4) \in \Delta(\Y)$. 

    First, we observe that in our simplified setup $\CY \equiv \reals^2, \ca(\Y) \equiv \reals^2$, $\le$ is the coordinate-wise order and the considered topology is the Euclidean topology. For a detailed discussion of those simplifications see the proof of Corollary~\ref{corollary:Characterization of Available Gambles of Identifiable Property-Forecasts in Finite Dimensions}. We further leverage an analogous representation of the set
    \begin{align*}
        \{ g \in \CY \colon g \le \alpha (\scoring_\gamma - \scoring_c) \text{ for some } \alpha \in \reals_{\ge 0}, c \in \propspace \} = \ScaledRegretGambles_\gamma + \CY_{\le 0},
    \end{align*}
    where $\CY_{\le 0} \coloneqq \{ g \in \reals^2 \colon g \le 0\}$ is the negative orthant. We show that $S_\gamma$ is open and argue that this implies $S_\gamma + \CY_{\le 0}$ is open. Some calculations give
    \begin{align*}
        \ScaledRegretGambles_{0.4} = \left \{ \alpha (0.16 - c^2, 0.16 - (1-c)^2) \colon c \in [0,1], \alpha \in \reals_{\ge 0}\right\}.
    \end{align*}
    Furthermore, solving for $c \in [0,1]$ and $\alpha \in \reals$ one notices $(-0.4,0.6) \notin \ScaledRegretGambles_{0.4}$. It would require $c = 0.4$ with $\alpha = \infty$. However, for every $\epsilon > 0$, we can give a point $\scoring_\epsilon \in S_{0.4}$ such that $\| (-0.4,0.6) - \scoring_\epsilon\|_2 \le \frac{1}{\sqrt{2}}\epsilon$. For instance, such a point is given by,
    \begin{align}
    \label{eq:approximating via scaled regret gambles}
        \scoring_\epsilon \coloneqq \beta_\epsilon (0.16 - c_\epsilon^2, 0.36 - (1-c_\epsilon)^2)
    \end{align}
    for $c_\epsilon \coloneqq 0.4 + \epsilon$ and $\beta_\epsilon \coloneqq \frac{1}{2 \epsilon}$.
    Concluding, the set $\ScaledRegretGambles_{0.4}$ is arbitrarily close to $(-0.4,0.6)$ but $(-0.4,0.6)$ is not included. The Minkowski-sum with $\CY_{\le 0}$ does not change this fact, since it only adds points which are smaller. However, the points we gave are approximating $(-0.4,0.6)$ from below. See Figure~\ref{fig: available gambles in the binary prediction game} for an illustration.
\end{example}

Identifiable properties are closely related to elicitable properties \citep{steinwart2014elicitation}. Hence, we pursue the natural follow-up question of the previous section: what is the set of available gambles in the case the forecasts are aligned to an identifiable property?

\subsection{Characterizing Available Gambles of Forecasts of Identifiable Properties}
\label{characerizing available gambles of identifiable properties}
We show that if $\Gamma$ is an identifiable property with identification function $\ident \colon \Y \times \propspace \rightarrow \reals$, the set of available gambles for the property-induced forecasting set is defined by all gambles $g \le \alpha \ident_\gamma$ for some $\alpha \in \reals$, where $\gamma \in \propspace$ is the forecasted property, or it can be approximated by such $g$. In other words, every available gamble is (approximately) dominated by a \emph{calibration gamble}.

Again, the proof is deferred to later (Section~\ref{proof: characerizing available gambles of identifiable properties}). The characterization follows analogous steps in comparison to the proof of Theorem~\ref{thm:Characterization of Available Gambles in Protocol 2 with Elicitable Property}.
\begin{theorem}[Available Gambles of Identifiable Property-Forecasts]
\label{thm:Characterization of Available Gambles in Protocol 2 with Identifiable Property}
    Let $\Gamma \colon \Q \rightarrow 2^\propspace$ be an identifiable property with identification function $\ident\colon \Y \times \propspace \rightarrow \reals$. For a fixed $\gamma \in \propspace$, we define\footnote{Closure is taken with respect to $\pqtopology$-topology.}
    \begin{equation}
    \label{eq:dominated by calibration gambles}
        \mathcal{H}_{\ident_\gamma} \coloneqq \cl \{ g \in \CY \colon g \le \alpha \ident_\gamma \text{ for some } \alpha \in \reals \}.
    \end{equation} Then, $$\mathcal{H}_{\ident_\gamma} = \offer_{\Gamma^{-1}(\gamma)}.$$
\end{theorem}
Unlike the characterization in Theorem~\ref{thm:Characterization of Available Gambles in Protocol 2 with Elicitable Property}, the closure of the set of gambles upper bounded by calibration gambles can be neglected in finite dimensions. We can show that $\mathcal{H}_{\ident_\gamma}$ is closed, if $|\Y| < \infty$.
\begin{corollary}[Theorem~\ref{thm:Characterization of Available Gambles in Protocol 2 with Identifiable Property} in Finite Dimensions]
\label{corollary:Characterization of Available Gambles of Identifiable Property-Forecasts in Finite Dimensions}
    Let $\Y$ be finite. Let $\Gamma \colon \Q \rightarrow 2^\propspace$ be an identifiable property with identification function $\ident\colon \Y \times \propspace \rightarrow \reals$. For a fixed $\gamma \in \propspace$,
    \begin{align*}
        \mathcal{H}_{\ident_\gamma} = \left\{ g \in \CY \colon g \le \alpha \ident_\gamma \text{ for some } \alpha \in \reals \right\} = \offer_{\Gamma^{-1}(\gamma)} .
    \end{align*}
\end{corollary}

\subsection{Proving the Characterizations}
\label{sec:Proving the Characterizations}
Let us now turn to proving the stated theorems. In the next three subsections, we first show that for a general forecasting set $\forecastingset \subseteq \Delta(\Y)$ (Section~\ref{the forecasts}) the set of available gambles forms a structured convex cone called an \emph{offer}. We then demonstrate that if the forecasting set is convex and closed, then the offer is in one-to-one correspondence with the forecasting set, \ie, the forecasting set characterizes the offer, and the offer characterizes the forecasting set. Finally, we leverage the obtained results to characterize the set of available gambles for elicitable and identifiable properties. 

The proof ideas for the statements in Section~\ref{From Forecasting Sets to Available Gambles and Back} are not new (\eg, \citep{follmer2011stochastic, benavoli2017polarity}). However, we are not aware of any work which has proved Theorem~\ref{thm:Available Gambles Form Offer} and Theorem~\ref{thm:representation: credal sets - offers - MAIN} for the $\CY$-$\ca(\Y)$-duality.
\subsection{The Set of Available Gambles is a Structured Convex Cone Called Offer}
\label{From Forecasting Sets to Available Gambles and Back}
Every set of available gambles possess a specific structure lent from convex analysis. They form what we call an \emph{offer}. We borrow this term from \citep[\S~6.4]{shafer2019game} who use it for an analogous structure to which we compare below.
\begin{definition}[Offer]
    \label{def:offer}
        We call a set $\offer \subseteq \CY$ an \emph{offer} if all of the following conditions are fulfilled:
        \begin{enumerate}[label=\textbf{O\arabic*.}, ref=O\arabic*]
        \item \label{def:offer - additivity}
        If $g, f \in \offer$, then $g + f \in \offer$. \quad (Additivity)
        \item \label{def:offer - positive homogeneity}
        If $ f \in \offer$ and $\alpha \ge 0$, then $\alpha f \in \offer$.\quad (Positive Homogeneity)
        \item \label{def:offer - coherence}
        If $g \in \offer$, then $\inf g \le 0$.\quad (Prohibiting Sure Gains)
        \item \label{def:offer - default availability}
        Let $g \in \CY$, if $\sup g \le 0$, then $g \in \offer$.\quad (Default Availability)
        \item \label{def:offer - closure}
        The set $\offer$ is closed with respect to the $\pqtopology$ topology. \quad (Closure)
        \end{enumerate}
\end{definition}
Closure, additivity and positive homogeneity demand that an offer $\offer$ is a closed, convex cone containing the zero element in $\CY$. Default availability then implies that all non-positive gambles are available. ``Prohibiting Sure Gains'' ensures that Gambler does not safely increase its capital via a gamble.\footnote{There are subtle differences between our notion of ``Prohibiting Sure Gains'' and the related notion of ``avoiding sure loss'' in imprecise probability, \eg, \citep[Section 3.7.3]{walley1991statistical}. Mainly, our definition contains a sign flip.}

Definition~\ref{def:offer} is closely related to the definition of offer \citep[\S~6.4]{shafer2019game}. First, our definition works for gambles in $\CY$-spaces, while \citet{shafer2019game} consider more generally functions from $\Y$ to the extended real line.\footnote{The reason why we restrict ourselves to a smaller function space of gambles is that otherwise the representability of upper expectations (Definition~\ref{def:coherent upper prevision}) as support functions over closed, convex sets of probability distributions requires more technical tools which are beyond the scope of this paper, \cf \citep[Section 5]{delbaen2002coherent}.} Their axiom G1 is equivalent our \ref{def:offer - additivity}. Their axiom G2 is equivalent to our \ref{def:offer - positive homogeneity}. Their axiom G4 is equivalent to our \ref{def:offer - coherence}. Their axiom G3 together with G2 implies our \ref{def:offer - default availability}. Our \ref{def:offer - default availability} together with \ref{def:offer - additivity} implies their axiom G3. Finally, their axiom G0 and G5 are closure conditions analogous to our \ref{def:offer - closure} adapted to the respective function spaces. As detailed in \citep[\S~6.7]{shafer2019game} offers are the sign-flipped counterparts of sets of desirable gambles \citep{walley1991statistical}.

\begin{theorem}[Available Gambles Form an Offer]
\label{thm:Available Gambles Form Offer}
    Let $\forecastingset \subseteq \Delta(\Y)$ be a forecasting set. The set,
    \begin{align*}
        \offer_{\forecastingset} \coloneqq \left\{ g \in \CY \colon \sup_{\phi \in \forecastingset} \mathbb{E}_\phi[g] \le 0 \right\} \subseteq \CY,
    \end{align*}
    is an offer.
\end{theorem}
In order to provide the proof, we introduce so-called \emph{upper expectations} which are closely related to forecasting sets.
\begin{definition}[Upper Expectation]
\label{def:coherent upper prevision}
    A functional $\localupperexp \colon \CY \rightarrow (-\infty, \infty]$ for which the following axioms hold is called an \emph{upper expectation}.
    \begin{enumerate}[label=\textbf{UE\arabic*.}, ref=UE\arabic*]
        \item \label{def:coherent upper prevision - subadditivity} If $f_1, f_2 \in \CY$, then $\localupperexp[f_1 + f_2] \le \localupperexp[f_1] + \localupperexp[f_2]$. \ (Subadditivity)
        \item \label{def:coherent upper prevision - positive homogeneity} If $f \in \CY$ and $c \in [0, \infty)$, then $\localupperexp[c f] \le c\localupperexp[f]$. \ (Positive Homogeneity)
        \item \label{def:coherent upper prevision - monotonicity} If $f_1, f_2 \in \CY$ such that $f_1 \le f_2$, then $\localupperexp[f_1] \le \localupperexp[f_2]$. \ (Monotonicity)
        \item \label{def:coherent upper prevision - translation equivariance} If $c \in \reals$, then $\localupperexp[c] = c$. \ (Translation Equivariance)
        \item \label{def:coherent upper prevision - lower semi-continuity} For every $\alpha \in \reals$ the set $\{f \in \CY\colon \localupperexp[f] \le \alpha \}$ is $\pqtopology$-closed. \ (Lower Semicontinuity)
    \end{enumerate}
\end{definition}
We first show that $\localupperexp_\forecastingset[g] \coloneqq \sup_{\phi \in \forecastingset} \mathbb{E}_\phi[g]$ for any $\forecastingset \subseteq \Delta(\Y)$ is an upper expectation. Then, the set of all gambles with non-positive upper expectation form an offer.
\begin{proof}
    Let  $\localupperexp_\forecastingset[g] \coloneqq \sup_{\phi \in \forecastingset} \mathbb{E}_\phi[g]$. We observe that \citep[\p 251]{aliprantis2006infinite}
    \begin{align*}
        \localupperexp_\forecastingset[g] \coloneqq \sup_{\phi \in \forecastingset} \mathbb{E}_\phi[g] = \sup_{\phi \in \cvxcl \forecastingset} \mathbb{E}_\phi[g],
    \end{align*}
    where $\cvxcl$ denotes the $\qptopology$-closure of the convex hull of $\forecastingset$.

    Hence, we can apply the fundamental representation result for support functions and closed convex sets \citep[Theorem 7.51]{aliprantis2006infinite}. It guarantees that $\localupperexp_\forecastingset$ satisfies axioms~\ref{def:coherent upper prevision - subadditivity},\ref{def:coherent upper prevision - positive homogeneity} and \ref{def:coherent upper prevision - lower semi-continuity}. Since, $\forecastingset \subseteq \Delta(\Y)$ it follows $\cvxcl \forecastingset \subseteq \Delta(\Y)$ because $\Delta(\Y)$ is $\qptopology$-closed \citep[Theorem 15.11]{aliprantis2006infinite} and convex. Translation equivariance \ref{def:coherent upper prevision - translation equivariance} follows because, for all $c \in \reals_{\ge 0}$,
    \begin{align*}
        \sup_{\phi \in \forecastingset} \mathbb{E}_{\phi}[c] \overset{\ref{def:coherent upper prevision - positive homogeneity}}{=} c \sup_{\phi \in \forecastingset} \mathbb{E}_{\phi}[1] = c,
    \end{align*}
    because $\forecastingset \subseteq \Delta(\Y)$, respectively, for all $c \in \reals_{< 0}$,
    \begin{align*}
        \sup_{\phi \in \forecastingset} \mathbb{E}_{\phi}[c] \overset{\ref{def:coherent upper prevision - positive homogeneity}}{=} c \inf_{\phi \in \forecastingset} \mathbb{E}_{\phi}[1] = c,
    \end{align*}
    because $\forecastingset \subseteq \Delta(\Y)$. Monotonicity \ref{def:coherent upper prevision - monotonicity} follows because $\forecastingset \subseteq \ca(\Y)_{\ge 0}$.

    It remains to show that the set of gambles for which the upper expectation is non-positive form an offer. Let $\offer_{\forecastingset} \coloneqq \left\{ g \in \CY \colon \localupperexp_\forecastingset[g] \le 0 \right\}$.
    We step-by-step prove all axioms that $\offer_\forecastingset$ has to fulfill to be an offer.
    Axiom~\ref{def:offer - additivity} follows from axiom~\ref{def:coherent upper prevision - subadditivity}. Axiom~\ref{def:offer - positive homogeneity} follows from axiom~\ref{def:coherent upper prevision - positive homogeneity}. Axiom~\ref{def:offer - coherence} is given, because if $\inf g > 0$, then $\localupperexp[g] \ge \localupperexp[\inf g] > 0$ by axiom~\ref{def:coherent upper prevision - monotonicity} and axiom~\ref{def:coherent upper prevision - translation equivariance}, from which follows that $g \notin \{ g \in \CY \colon \localupperexp[g] \le 0\}$. Axiom~\ref{def:offer - default availability} follows from axiom~\ref{def:coherent upper prevision - monotonicity} and axiom~\ref{def:coherent upper prevision - translation equivariance}. Let $g \in \CY$ with $\sup g \le 0$. Then $\localupperexp[g] \le \localupperexp[\sup g ] \le 0$ implies $g \in \offer_{\localupperexp}$. Axiom~\ref{def:offer - closure} directly follows from axiom~\ref{def:coherent upper prevision - lower semi-continuity}.
\end{proof}
The set of all available gambles is an offer. This reformulation, in fact, is not a one-way street. Given an arbitrary offer we can provide a forecasting set whose set of available gambles is exactly the offer (Theorem~\ref{thm:representation: credal sets - offers - MAIN}). This forecasting set is convex and closed. We call such forecasting sets \emph{credal}.
\begin{definition}[Credal Set]
\label{def:credal set}
    We call a non-empty set $\credal \subseteq \Delta(\Y)$ a \emph{credal set} if all of the following conditions are fulfilled
    \begin{enumerate}[label=\textbf{CS\arabic*.}, ref=CS\arabic*]
        \item \label{def:credal set - convexity} Let $p_1, \ldots, p_n \in \credal$ and $\alpha_1, \ldots, \alpha_n \in \reals_{\ge 0}$ such $\sum_{i = 1}^n \alpha_i = 1$, then $\sum_{i = 1}^n \alpha_i p_i \in \credal$. \quad (Convexity)
        \item \label{def:credal set - closure} The set $\credal$ is closed with respect to the $\qptopology$-topology. \quad (Closure)
    \end{enumerate}
\end{definition}
The term ``credal set'' is borrowed from the literature on imprecise probability \citep{augustin2014introduction}. We emphasize that a credal set is not necessarily linked to any kind of a belief of Forecaster. We use the term here to describe a forecasting set with a particular structure. Note that $\cvxcl \forecastingset$ is a credal set for every forecasting set $\forecastingset \subseteq \Delta(\Y)$.

To formally state Theorem~\ref{thm:representation: credal sets - offers - MAIN} we introduce polar sets.
\begin{definition}[Polar Set]
\label{def:polar set}
    We define the polar set for non-empty $W \subseteq \CY$ and non-empty $V \subseteq \ca(\Y)$ as
    \begin{align*}
        W^\circ \coloneqq \{ \phi \in \ca(\Y) \colon \langle \phi,g \rangle \le 1, \forall g \in W \},
    \end{align*}
    respectively
    \begin{align*}
        V^\circ \coloneqq \{ g \in \CY \colon \langle \phi,g \rangle \le 1, \forall\phi \in V \}.
    \end{align*}
\end{definition}
Finally, we are able to spell out a one-to-one correspondence between credal sets and offers.
\begin{theorem}[Representation: Credal Sets -- Offers]
\label{thm:representation: credal sets - offers - MAIN}
    Let $\offer \subseteq \CY$ be an offer. The set
    \begin{align*}
        \credal_\offer \coloneqq \offer^\circ \cap \Delta(\Y) \subseteq \ca(\Y)
    \end{align*}
    is a credal set, such that
    \begin{align*}
        \offer = \left\{ g \in \CY \colon \sup_{\phi \in \credal_\offer} \mathbb{E}_\phi[g] \le 0 \right\} = (\reals_{\ge 0}\credal_\offer)^\circ .
    \end{align*}
    Conversely, let $\credal \subseteq \Delta(\Y)$ be a credal set. The set
    \begin{align*}
        \offer_\credal \coloneqq (\reals_{\ge 0}\credal)^\circ \subseteq \CY
    \end{align*}
    is an offer, such that
    \begin{align*}
        \credal \coloneqq \offer_\credal^\circ \cap \Delta(\Y).
    \end{align*}
    Thus, every offer $\offer$ is in one-to-one correspondence to a credal set $\credal$.
\end{theorem}
In order to show Theorem~\ref{thm:representation: credal sets - offers - MAIN}, we use several helpful properties of polar sets.
\begin{proposition}[Properties of Polar Set]
\label{proposition:properties of polar set}
    For non-empty $W_1, W_2 \subseteq \CY$ we have:
    \begin{enumerate}[label=\textbf{P\arabic*.}, ref=P\arabic*]
        \item \label{prop:properties of polar set - polarity} If $W_1 \subseteq W_2$, then $W_2^\circ \subseteq W_1^\circ$.
        \item \label{prop:properties of polar set - convexclosed of polar} The set $W_1^\circ$ is non-empty, convex, $\qptopology$-closed and contains the zero element.
        \item \label{prop:properties of polar set - bipolar theorem} If $W_1$ is non-empty, convex, $\pqtopology$-closed and contains the zero element, then $W_1 = (W_1^\circ)^\circ$.
        \item \label{prop:properties of polar set - polar set of cone} If $W_1$ is a convex cone, then $W_1^\circ  = \{ \phi \in \ca(\Y) \colon \langle \phi,g \rangle \le 0, \forall g \in W_1 \}$, which is a convex cone.
        \item \label{prop:properties of polar set - polar set of negative quadrant} $\left(\CY_{\le 0} \right)^\circ = \ca(\Y)_{\ge 0}$.
    \end{enumerate}
    Analogous results hold for $V_1, V_2 \subseteq \ca(\Y)$.
\end{proposition}
\begin{proof}
    Statement~\ref{prop:properties of polar set - polarity} and \ref{prop:properties of polar set - convexclosed of polar} follow from Lemma 5.102 in \citep{aliprantis2006infinite}. Statement~\ref{prop:properties of polar set - bipolar theorem} is the famous Bipolar Theorem \citep[Theorem 5.103]{aliprantis2006infinite}. The Statement~\ref{prop:properties of polar set - polar set of cone} can be found on \p 215 in the same book. It remains to show Statement~\ref{prop:properties of polar set - polar set of negative quadrant}.

    Since $\CY_{\le 0}$ is a convex cone, we leverage Property~\ref{prop:properties of polar set - polar set of cone} to get:
    \begin{align*}
        \left(\CY_{\le 0} \right)^\circ = \{ \phi \in \ca(\Y)\colon \langle \phi, g\rangle \le 0, \forall g \in \CY_{\le 0}\}.
    \end{align*}
    We can easily see that for every $\phi \in \ca(\Y)_{\ge 0}$, $\langle \phi, g\rangle \le 0$ for all $g \in \CY_{\le 0}$. For the reverse direction,
    consider any $\phi \in \ca(\Y)$ such that $\phi$ is negative on a measurable set $A$. We show that such $\phi \notin \left(\CY_{\le 0} \right)^\circ$. Let $g(y) \coloneqq - \llbracket y \in A \rrbracket$\footnote{We denote the Iverson-bracket as $\llbracket \cdot \rrbracket$, \ie $\llbracket S \rrbracket = 1$, if $S$ is true, $0$ otherwise.}. Obviously, $g \in \CY_{\le 0}$. Furthermore,
    \begin{align*}
        \langle \phi, g\rangle = \int_\Y gd\phi
        = \int_A g d\phi + \int_{\Y \setminus A} g d\phi
        = - \int_A d\phi > 0.
    \end{align*}
    It follows that $\phi \notin \left(\CY_{\le 0} \right)^\circ$. Hence, $\left(\CY_{\le 0} \right)^\circ \subseteq \ca(\Y)_{\ge 0}$.
\end{proof}
Furthermore, we guarantee that the polar of an offer is non-trivial.
\begin{lemma}[Polar of Offer is Non-Trivial]
\label{lemma:Polar of Offer is Non-Trivial}
    Let $\offer \subseteq \CY$ be an offer. Then, $\offer^\circ \neq \{ 0 \}$.
\end{lemma}
\begin{proof}
    If $\offer^\circ = \{ 0\}$, then $\offer = \offer^{\circ\circ} = \{ 0\}^\circ = \CY$ (Proposition~\ref{proposition:properties of polar set}). But this offer $\offer$ is not legitimate, since it violates Axiom~\ref{def:offer - coherence}.
\end{proof}
The proof of Theorem~\ref{thm:representation: credal sets - offers - MAIN} is given in five steps. First, we show that $\credal_\offer$ is a credal set by going through the axioms. Then, we argue that $\offer = \left\{ g \in \CY \colon \sup_{\phi \in \credal_\offer} \mathbb{E}_\phi[g] \le 0 \right\} = (\reals_{\ge 0}\credal_\offer)^\circ$. Thirdly, $\offer_\credal$ is shown to be an offer. After that, we prove $\credal \coloneqq \offer_\credal^\circ \cap \Delta(\Y)$. Finally, we argue that the mapping between offers and credal sets is bijective.
\begin{proof}\textbf{of Theorem~\ref{thm:representation: credal sets - offers - MAIN}}
    \begin{enumerate}
        \item Let $\offer \subseteq \CY$ be an offer. We show that $\offer^\circ \cap \Delta(\Y) \subseteq \ca(\Y)$ is a credal set. First, $\CY_{\le 0} \subseteq \offer$ (Condition~\ref{def:offer - default availability}). Thus, $\offer^\circ \subseteq \ca(\Y)_{\ge 0}$ by Proposition~\ref{proposition:properties of polar set}, Statement~\ref{prop:properties of polar set - polarity} and Statement~\ref{prop:properties of polar set - polar set of negative quadrant}. Furthermore, $\offer^\circ$ is $\qptopology$-closed and convex (Proposition~\ref{proposition:properties of polar set}, Statement~\ref{prop:properties of polar set - convexclosed of polar}). In particular, $\offer^\circ$ contains at least one element $\phi \in \ca(\Y)_{\ge 0}$, such that $\| \phi \|_q = 1$. We can easily see this fact, because if $\phi \in \offer^\circ$ is not equal to zero, then $\phi' \coloneqq \frac{\phi}{\| \phi\|_q} \in \offer^\circ$ (Proposition~\ref{proposition:properties of polar set}, Statement~\ref{prop:properties of polar set - polar set of cone}) and $\| \phi'\|_q = 1$. Importantly, such a non-zero $\phi \in \offer^\circ$ exists (Lemma~\ref{lemma:Polar of Offer is Non-Trivial}).
    From all these considerations it follows that the intersection $\offer^\circ \cap \Delta(\Y)$ is non-empty. Furthermore, it is $\qptopology$-closed and convex, because both $\offer^\circ$ and $\Delta(\Y)$ fulfill those intersection-stable properties.

    \item We have,
    \begin{align*}
        \left\{ g \in \CY \colon \sup_{\phi \in \credal_\offer} \mathbb{E}_\phi[g] \le 0 \right\} &= \left\{ g \in \CY \colon \langle \phi, g \rangle \le 0, \forall \phi \in \credal_\offer \right\}\\
        &= \left\{ g \in \CY \colon \langle r \phi, g \rangle \le 0, \forall \phi \in \credal_\offer, \forall r \in \reals_{\ge 0} \right\}\\
        &= \left\{ g \in \CY \colon \langle t,g \rangle \le 0, \forall t \in \reals_{\ge 0}\credal_\offer \right\}\\
        &\overset{\ref{prop:properties of polar set - polar set of cone}}{=} (\reals_{\ge 0}\credal_\offer)^\circ\\
        &= \left(\reals_{\ge 0}(\offer^\circ \cap \Delta(\Y)) \right)^\circ\\
        &= \left( \{ r \phi \colon r \in \reals_{\ge 0}, \phi \in \offer^\circ, \phi \in \Delta(\Y) \} \right)^\circ\\
        &=  \left( \{ r \phi \colon r \in \reals_{\ge 0}, \phi \in \offer^\circ\} \cap \{r \phi \colon r \in \reals_{\ge 0}, \phi \in \Delta(\Y) \} \right)^\circ\\
        &= \left(\reals_{\ge 0}\offer^\circ \cap \reals_{\ge 0} \Delta(\Y) \right)^\circ\\
        &\overset{\ref{prop:properties of polar set - polar set of cone}}{=} \left(\offer^\circ \cap \reals_{\ge 0} \Delta(\Y) \right)^\circ\\
        &= \left(\offer^\circ \cap \ca(\Y)_{\ge 0} \right)^\circ\\
        &\overset{\ref{def:offer - default availability}}{=} (\offer^\circ)^\circ\\
        &\overset{\ref{prop:properties of polar set - bipolar theorem}}{=} \offer.
    \end{align*}

    \item Let $\credal \subseteq \ca(\Y)$ be a credal set. We show that $\offer_\credal = (\reals_{\ge 0}\credal)^\circ$ is an offer.
    First, $\offer_\credal$ is $\pqtopology$-closed (Condition~\ref{def:offer - closure}), a convex cone (Condition~\ref{def:offer - positive homogeneity} and \ref{def:offer - additivity}) that contains the zero element (Proposition~\ref{proposition:properties of polar set}, Statement~\ref{prop:properties of polar set - convexclosed of polar} and~\ref{prop:properties of polar set - polar set of cone}).
    Furthermore, $\reals_{\ge 0}\credal \subseteq \ca(\Y)_{\ge 0}$ implies $(\reals_{\ge 0}\credal)^\circ \supseteq \CY_{\le 0}$ which is Condition~\ref{def:offer - default availability} (Proposition~\ref{proposition:properties of polar set}, Statement~\ref{prop:properties of polar set - polarity} and~\ref{prop:properties of polar set - polar set of negative quadrant}). For any $g \in (\reals_{\ge 0}\credal)^\circ$ we have that $\inf g \le 0$, otherwise $c \coloneqq \inf g > 0$, hence,
    \begin{align*}
        \langle \phi, g\rangle = \int_{\Y} g d\phi
        \ge \int_{\Y} c d\phi
        = c \int_{\Y} d\phi > 0,
    \end{align*}
    for some $\phi \in \credal$ (Axiom~\ref{def:offer - coherence}).

    \item We have,
    \begin{align*}
        \credal_\offer = \offer^\circ \cap \Delta(\Y)
        = \left((\reals_{\ge 0}\credal)^\circ\right)^\circ \cap \Delta(\Y)
        = \credal.
    \end{align*}

    \item 
    For the bijectivity of the mapping, we have shown $\Tilde{\offer} = \offer_{\credal_{\Tilde{\offer}}}$ for any offer $\Tilde{\offer}$ and $\Tilde{\credal} = \credal_{\offer_{\Tilde{\credal}}}$ for any credal set $\Tilde{\credal}$. Hence, the mapping between offers and credal sets has a left and a right inverse, hence is a bijection.
    \end{enumerate}
\end{proof}
In conclusion, every forecasting set provides a set of available gambles which fulfills the axioms of an offer. An offer in turn is in one-to-one correspondence with respect to credal sets, forecasting sets which are closed and convex. We will make use of the obtained statements in the following two sections in which we characterize the set of available gambles if the forecasting sets are not obtained through arbitrary properties, but if the properties are elicitable (respectively identifiable).

\subsection{Proof of Theorem~\ref{thm:Characterization of Available Gambles in Protocol 2 with Elicitable Property}: Characterizing Available Gambles of Forecasts of Elicitable Properties}
\label{proof: characerizing available gambles of elicitable properties}
Via the introduced notion of an offer (Definition~\ref{def:offer}) and the fundamental correspondence Theorem~\ref{thm:representation: credal sets - offers - MAIN}, we can show that the set of available gambles for a forecast aligned to an elicitable property is defined through the set of scaled regret gambles.
\begin{proof}
    First, we show that $\mathcal{H}_{\scoring_\gamma}$ is an offer following Definition~\ref{def:offer}:
    \begin{itemize}[]
        \item \ref{def:offer - additivity}: Let $g, f \in \mathcal{H}_{\scoring_\gamma}$, then,
        \begin{align*}
            f + g &\le \alpha_f (\scoring_\gamma - \scoring_{c_f}) + \alpha_g (\scoring_\gamma - \scoring_{c_g})\\
            &= (\alpha_f + \alpha_g) \left( \scoring_\gamma - \left( \frac{\alpha_f}{\alpha_f + \alpha_g}\scoring_{c_f} + \frac{\alpha_g}{\alpha_f + \alpha_g}\scoring_{c_g}\right) \right)\\
            &\le (\alpha_f + \alpha_g) \left( \scoring_\gamma - \scoring_{c'} \right)
        \end{align*}
        with $(\alpha_f + \alpha_g) \in \reals_{\ge 0}$ and $c' \in \propspace$. Such a $c'$ exists because $\scoring_{c_f}, \scoring_{c_g} \in \spr(\scoring)$ and the superprediction set $\spr(\scoring)$ is convex by assumption.
        \item \ref{def:offer - positive homogeneity}:  Let $g \in \mathcal{H}_{\scoring_\gamma}$ and $r \ge 0$. Then $r g \le r \alpha (\scoring_\gamma - \scoring_{c})$ with $r \alpha \in \reals_{\ge 0}$.
        \item \ref{def:offer - coherence}: Let $g \in \mathcal{H}_{\scoring_\gamma}$, then there is $\alpha \in \reals_{\ge 0}$ and $c \in \reals$ such that $g \le \alpha (\scoring_\gamma - \scoring_{c})$. In particular, $\inf g \le \inf \alpha (\scoring_\gamma - \scoring_{c})$. Hence, we require that $\inf \alpha (\scoring_\gamma - \scoring_{c}) \le 0$ for every $\alpha \ge 0$. This is guaranteed because there exist $\phi \in \Gamma^{-1}(\gamma) \neq \emptyset$ such that $\langle \phi , \scoring_\gamma - \scoring_c \rangle \le 0$ by consistency of the scoring function. Thus, Lemma~\ref{lemma:techincal lemma 1} applies.
        \item \ref{def:offer - default availability}: Let $g \in \CY$ with $\sup g \le 0$, then $g \le \alpha (\scoring_\gamma - \scoring_{c})$ for $\alpha = 0$. Thus, $g \in \mathcal{H}_{\scoring_\gamma}$.
        \item \ref{def:offer - closure}: By definition of $\mathcal{H}_{\scoring_\gamma}$ \eqref{eq:dominated by regret gambles}.
    \end{itemize}
    Then, we compute
    \begin{align*}
            &\mathcal{H}_{\scoring_\gamma}^\circ \cap \Delta(\Y)\\
            &\overset{(a)}{=} \left\{ g \in \CY \colon g \le \alpha (\scoring_\gamma - \scoring_c) \text{ for some } \alpha \in \reals_{\ge 0}, c \in \propspace \right \}^\circ \cap \Delta(\Y)\\
            &\overset{(b)}{=} \left\{ \phi \in \ca(\Y) \colon \langle \phi, g\rangle \le 0, \forall g \in \CY \text{ \st \ } g \le \alpha (\scoring_\gamma - \scoring_c) \text{ for some } \alpha \in \reals_{\ge 0}, c \in \propspace \right\} \cap \Delta(\Y)\\
            &= \left\{ \phi \in \Delta(\Y) \colon \langle \phi, g\rangle \le 0, \forall g \in \CY \text{ \st \ } g \le \alpha (\scoring_\gamma - \scoring_c) \text{ for some } \alpha \in \reals_{\ge 0}, c \in \propspace \right\}\\
            &\overset{(c)}{=} \left\{ \phi \in \Delta(\Y) \colon \langle \phi, \alpha (\scoring_\gamma - \scoring_c) \rangle \le 0, \forall \alpha \in \reals_{\ge 0}, c \in \propspace \right\}\\
            &= \left\{ \phi \in \Delta(\Y) \colon \alpha \left( \langle \phi, \scoring_\gamma \rangle -  \langle \phi, \scoring_c \rangle\right) \le 0, \forall \alpha \in \reals_{\ge 0}, c \in \propspace \right\}\\
            &= \left\{ \phi \in \Delta(\Y) \colon \langle \phi, \scoring_\gamma \rangle \le  \langle \phi, \scoring_c \rangle, \forall c \in \propspace \right\}\\
            &= \left\{ \phi \in \Delta(\Y) \colon \argmin_{c \in \propspace}\langle \phi, \scoring_c \rangle = \gamma \right\}\\
            &= \{ \phi \in \Delta(\Y) \colon \Gamma(\phi) = \gamma\}\\
            &= \Gamma^{-1}(\gamma).
    \end{align*}
    \begin{enumerate}[(a)]
        \item For convex sets containing zero, such as, $$A \coloneqq \left\{ g \in \CY \colon g \le \beta (\scoring_\gamma - \scoring_c) \text{ for some } \beta \in \reals_{\ge 0}, c \in \propspace \right \},$$ we have $\cl A = A^{\circ \circ}$, which implies $A^{\circ \circ \circ} = A^\circ = (\cl A)^\circ$ (Proposition~\ref{proposition:properties of polar set}).
        \item We have shown that $\left\{ g \in \CY \colon g \le \alpha (\scoring_\gamma - \scoring_c) \text{ for some } \alpha \in \reals_{\ge 0}, c \in \propspace \right \}$ is a cone in the first part of the proof. Thus, Proposition~\ref{proposition:properties of polar set} Statement~\ref{prop:properties of polar set - polar set of cone} applies.
        \item 
        We know that for every $g \in \CY$ such that $g \le \alpha (\scoring_\gamma - \scoring_c)$ for some $\alpha \in \reals_{\ge 0}$ and $c \in \propspace$,
        \begin{align*}
            \langle \phi, g\rangle &= \langle \phi, \alpha (\scoring_\gamma - \scoring_c) + g - \alpha (\scoring_\gamma - \scoring_c) \rangle\\
            &= \langle \phi, \alpha (\scoring_\gamma - \scoring_c) \rangle + \langle \phi, g - \alpha (\scoring_\gamma - \scoring_c) \rangle\\
            &\le \langle \phi, \alpha (\scoring_\gamma - \scoring_c) \rangle,
        \end{align*}
        because $\phi \ge 0$ and $g - \alpha (\scoring_\gamma - \scoring_c) \le 0$.
    \end{enumerate}
    We have already proven that $\Gamma^{-1}(\gamma)$ is a credal set for all $\gamma$ (Proposition~\ref{prop:Identifiable or Elicitable Property Give Credal Level Sets}). Furthermore, credal sets and offers are in one-to-one correspondence (Theorem~\ref{thm:representation: credal sets - offers - MAIN}).
    It follows that $\offer_{\Gamma^{-1}(\gamma)}$ is equal to $\mathcal{H}_{\scoring_\gamma}$.
\end{proof}

\subsection{Proof of Theorem~\ref{thm:Characterization of Available Gambles in Protocol 2 with Identifiable Property}: Characterizing Available Gambles of Forecasts of Identifiable Properties}
\label{proof: characerizing available gambles of identifiable properties}
The characterization of available gambles for forecasts aligned to identifiable properties follows analogously.
\begin{proof}
    First, we show that $\mathcal{H}_{\ident_\gamma}$ is an offer following Definition~\ref{def:offer}:
    \begin{itemize}[]
        \item \ref{def:offer - additivity}: Let $g, f \in \mathcal{H}_{\ident_\gamma}$, then $f + g \le \alpha_f \ident_\gamma + \alpha_g \ident_\gamma = (\alpha_f + \alpha_g) \ident_\gamma$, with $(\alpha_f + \alpha_g) \in \reals$.
        \item \ref{def:offer - positive homogeneity}:  Let $g \in \mathcal{H}_{\ident_\gamma}$ and $c \ge 0$. Then $c g \le c \alpha \ident_\gamma$ with $c \alpha \in \reals$.
        \item \ref{def:offer - coherence}: Let $g \in \mathcal{H}_{\ident_\gamma}$, then there is $\alpha \in \reals$ such that $g \le \alpha \ident_\gamma$. In particular, $\inf g \le \inf \alpha \ident_\gamma$. Hence, we require that $\inf \alpha \ident_\gamma \le 0$ for every $\alpha \in \reals$.
        To see this we remind the reader that $\Gamma^{-1}(\gamma)$ is non-empty (by assumption), \ie, there exist $\phi \in \Delta(\Y)$ such that $\langle \phi , \ident_\gamma\rangle = 0$. Thus, Lemma~\ref{lemma:techincal lemma 1} applies for $\ident_\gamma$ and $-\ident_\gamma$, which gives the desired result.
        \item \ref{def:offer - default availability}: Let $g \in \CY$ with $\sup g \le 0$, then $g \le \alpha \ident_\gamma$ for $\alpha = 0$. Thus, $g \in \mathcal{H}_{\ident_\gamma}$.
        \item \ref{def:offer - closure}: By definition of $\mathcal{H}_{\ident_\gamma}$ \eqref{eq:dominated by calibration gambles}.
    \end{itemize}
    Then, we compute
    \begin{align*}
        \mathcal{H}_{\ident_\gamma}^\circ \cap \Delta(\Y)
        &\overset{(a)}{=} \left\{ g \in \CY \colon g \le \alpha \ident_\gamma \text{ for some } \alpha \in \reals \right \}^\circ \cap \Delta(\Y)\\
        &\overset{(b)}{=} \left\{ \phi \in \ca(\Y) \colon \langle \phi, g\rangle \le 0, \forall g \in \CY \text{ such that } g \le \alpha \ident_\gamma \text{ for some } \alpha \in \reals \right\} \cap \Delta(\Y)\\
        &= \left\{ \phi \in \Delta(\Y) \colon \langle \phi, g\rangle \le 0, \forall g \in \CY \text{ such that } g \le \alpha \ident_\gamma \text{ for some } \alpha \in \reals \right\}\\
        &\overset{(c)}{=} \left\{ \phi \in \Delta(\Y) \colon \langle \phi, \alpha \ident_\gamma \rangle \le 0, \forall \alpha \in \reals \right\}\\
        &= \left\{ \phi \in \Delta(\Y) \colon \alpha \langle \phi, \ident_\gamma \rangle \le 0, \forall \alpha \in \reals \right\}\\
        &= \left\{ \phi \in \Delta(\Y) \colon \langle \phi, \ident_\gamma \rangle = 0 \right\}\\
        &= \{ \phi \in \Delta(\Y) \colon \Gamma(\phi) = \gamma\}\\
        &= \Gamma^{-1}(\gamma).
    \end{align*}
    \begin{enumerate}[(a)]
        \item For convex sets containing zero, such as $A \coloneqq \left\{ g \in \CY \colon g \le \alpha \ident_\gamma \text{ for some } \alpha \in \reals \right \}$, we have $\cl A = A^{\circ \circ}$, which implies $A^{\circ \circ \circ} = A^\circ = (\cl A)^\circ$ (Proposition~\ref{proposition:properties of polar set}).
        \item We  have shown that $\left\{ g \in \CY \colon g \le \alpha \ident_\gamma \text{ for some } \alpha \in \reals \right \}$ is a cone in the first part of the proof. Thus, Proposition~\ref{proposition:properties of polar set} Statement~\ref{prop:properties of polar set - polar set of cone} applies.
        \item 
        The top to bottom inclusion is trivial. Since we consider all $g \in \CY$ such that $g \le \alpha \ident_\gamma$, we in particular consider all $\alpha \ident_\gamma$ for $\alpha \in \reals$.
        For the reverse set inclusion, we know that for every $g \in \CY$ there is $\alpha \in \reals$ such that $g \le \alpha \ident_\gamma$. Thus,
        \begin{align*}
            \langle \phi, g\rangle &= \langle \phi, \alpha \ident_\gamma + g -\alpha  \ident_\gamma \rangle\\
            &= \langle \phi, \alpha \ident_\gamma \rangle + \langle \phi, g - \alpha \ident_\gamma \rangle\\
            &\le \langle \phi, \alpha \ident_\gamma \rangle,
        \end{align*}
        because $\phi \ge 0$ and $g - \ident_\gamma \le 0$. Hence, we can simply focus on all $\alpha \ident_\gamma$ functions, instead of the previously specified $g$.
    \end{enumerate}
    Recall that we have already proven that $\Gamma^{-1}(\gamma)$ is a credal set for all $\gamma$ (Proposition~\ref{prop:Identifiable or Elicitable Property Give Credal Level Sets}). Furthermore, credal sets and offers are in one-to-one correspondence (Theorem~\ref{thm:representation: credal sets - offers - MAIN}).
    It follows that $\offer_{\Gamma^{-1}(\gamma)}$ is equal to $\mathcal{H}_{\ident_\gamma}$.
\end{proof}
The proof of Corollary~\ref{corollary:Characterization of Available Gambles of Identifiable Property-Forecasts in Finite Dimensions} then simply requires to show that the set of gambles upper bounded by calibration gambles is closed for finite $\Y$.
\begin{proof}\textbf{of Corollary~\ref{corollary:Characterization of Available Gambles of Identifiable Property-Forecasts in Finite Dimensions}}
    We have to show that,
    \begin{align*}
        \mathcal{F}_{\ident_\gamma} \coloneqq \{ g \in \CY \colon g \le \alpha \ident_\gamma \text{ for some } \alpha \in \reals \},
    \end{align*}
    is $\qptopology$-closed in a finite dimensional $\ca(\Y)$-space, \ie, $\mathcal{F}_{\ident_\gamma}  = \mathcal{H}_{\ident_\gamma}$

    Given that $\Y$ is finite, $\CY$ (respectively $\ca(\Y)$) reduces to $\reals^d$ for $d = |\mathcal{Y}|$. The topology induced by the supremum norm is equivalent to the Euclidean topology \citep[\p 580]{schechter1997handbook} and the equivalent to the $\pqtopology$-topology \citep[28.17 (e)]{schechter1997handbook}. In short, we can use the standard topology on finite dimensional Euclidean spaces to express the next results. 

    First, we observe that we can write $\mathcal{F}_{\ident_\gamma}$ in terms of a Minkowski-sum
    \begin{align*}
        \mathcal{F}_{\ident_\gamma} &= \{ g \in \CY \colon g -  \alpha \ident_\gamma \le 0 \text{ for some } \alpha \in \reals \}\\
        &= \{ \alpha \ident_\gamma \colon \alpha \in \reals \} + \{ g \in \CY \colon g \le 0\}.
    \end{align*}
    Let us introduce the shorthands $V_\gamma \coloneqq \{ \alpha \ident_\gamma \colon \alpha \in \reals \}$ for the line and $\CY_{\le 0} \coloneqq \{ g \in \CY \colon g \le 0\}$ for the negative orthant.

    We distinguish between two simple cases:
    \begin{description}
        \item [Case 1] Assume that $\ident_\gamma \ge 0$ or $-\ident_\gamma \ge 0$. Axiom~\ref{def:offer - coherence} guarantees that there exist at least one $y \in \Y$ such that $\pm\ident_\gamma(y) = 0$. We denote the subset of those $y$'s in which $\ident_\gamma$ is zero $\Y_0 \subseteq \Y$. It is relatively easy to see that
        \begin{align*}
            \mathcal{F}_{\ident_\gamma}
            = \{ g \in \CY \colon g(y) \le 0, \forall y \in \Y_0\}.
        \end{align*}
        The left to right set inclusion is clear by definition of $\mathcal{F}_{\ident_\gamma}$. For the right to left set inclusion choose $\alpha_g \coloneqq \max_{y \in \Y \setminus \Y_0}\left(\frac{g(y)}{\ident_\gamma(y)}\right)$ for arbitrary $g \in \CY$ such that $g(y) \le 0$ for all $y \in \Y_0$. Then, $g \le \alpha_g \ident_{\gamma}$.
        Concluding, the obtained set is closed in the Euclidean topology.
        
        \item [Case 2] In the other case, we have to leverage a result on the closedness of Minkowski-sums going back to \citet{debreu1959theory}. A Minkowski-sum is closed if $V_\gamma$ and $\CY_{\le 0}$ are closed and the asymptotic cones of $V_\gamma$ and $\CY_{\le 0}$ are positively semi-independent \citep[Theorem 20.2.3]{border2019lecture} or \citep[\p 23 (9)]{debreu1959theory}.
    
        Asymptotic cones, roughly stated, form the set of directions in which a set is unbounded. Positively semi-independent requires that for $v \in V_\gamma$ and $o \in \CY_{\le 0}$, $v + o = 0$ implies $v = o = 0$. In our case, $V_\gamma$ and $\CY_{\le 0}$ are closed cones, \ie, for all $\alpha \in \reals$ and elements $v \in V_\gamma$ and $o \in \CY_{\le 0}$ it holds $\alpha v \in V_\gamma$ respectively $\alpha o \in \CY_{\le 0}$. Hence,  the asymptotic cones of $V_\gamma$ and $\CY_{\le 0}$ are the sets themselves \citep[Theorem 20.2.2 f) i)]{border2019lecture}.
    
        It remains to show that the two sets are positively semi-independent, which follows from the following observation. If $a + o = 0$ for $a \in \CY$ and $o \in \CY_{\le 0}$, then $a \in \CY_{\ge 0}$ where $\CY_{\ge 0} \coloneqq \{ g \in \CY \colon g \ge 0\}$. Now, $V_\gamma \cap \CY_{\ge 0} = \{ 0\}$ because there exist $y,y' \in \mathcal{Y}$ such that $\ident_\gamma(y) < 0$ and $\ident_\gamma(y') > 0$. Hence, $a = 0$, which implies $o = 0$.
    \end{description}
\end{proof}

\subsection{On the Relationship of Unscaled Regret and Calibration Gambles}
\label{On Optimality of Unscaled Regret or Unscaled Calibration Gambles}
In Section~\ref{Regret Gambles and Calibration Gambles are Available} we have argued that (unscaled) regret gambles and calibration gambles are both available. The characterization theorems (Theorem~\ref{thm:Characterization of Available Gambles in Protocol 2 with Elicitable Property} and Theorem~\ref{thm:Characterization of Available Gambles in Protocol 2 with Identifiable Property}) complete the picture. \emph{Scaled} regret gambles and calibration gambles are, up to topological approximation, dominating all available gambles.

To remind the reader, if a gamble $g$ dominates another gamble $h$, then playing gamble $g$ is favorable to the gambler. No matter which outcome $\nature \in \Y$ realizes, the value $g(\nature) \ge h(\nature)$. In other words, scaled regret gambles (respectively calibration gambles) are among the best gambles a gambler can play to disprove Forecaster to provide adequate forecasts. In principle, a gambler has no need to explore any other gambles than the scaled regret gambles or the calibration gambles. However, practically relevant evaluation metrics due to further restrictions put on the gambler embrace a broader scope of available gambles (Section~\ref{The Fair, Restricted and Rational Gambler}).

Nevertheless, the characterization statements offer two further interesting consequences. First, in the case that a property is elicited by several distinct loss functions\footnote{ For instance, the mean is elicited by all Bregman-divergences \citep{gneiting2011making}.}, the set of available gambles does not change. Hence, the set of of scaled regret gambles still (approximately) dominate all those available gambles. But, the sets of \emph{un}scaled regret gambles for difference loss functions are not necessarily equal.

Second, in the case that a property is identifiable and elicitable\footnote{Continuous, real-valued, elicitable properties are, under mild technical conditions, identifiable and vice-versa \citep{lambert2008eliciting, steinwart2014elicitation}.}, then \emph{scaled} regret gambles are equally describing the set of all available gambles as calibration gambles do.
\begin{corollary}[Duality of Calibration and Regret]
\label{corollary:Duality of Calibration and Regret}
    Let $\Gamma \colon \Q \rightarrow 2^\propspace$ be an identifiable and elicitable property with identification function $\ident\colon \Y \times \propspace \rightarrow \reals$ and scoring function $\scoring \colon \Y \times \propspace \rightarrow \reals$ which has a convex superprediction set. Let $\gamma \in \reals$ such that $\Gamma^{-1}(\gamma) \neq \emptyset$. Then,
    \begin{align*}
        \mathcal{H}_{\scoring_\gamma} = \mathcal{H}_{\ident_\gamma}.
    \end{align*}
\end{corollary}
\begin{proof}
    The statement is a simple consequence of Theorem~\ref{thm:Characterization of Available Gambles in Protocol 2 with Elicitable Property} and Theorem~\ref{thm:Characterization of Available Gambles in Protocol 2 with Identifiable Property}.
\end{proof}
One may interpret this statement as that, in principle, scaled regret is as powerful as calibration in proving Forecaster's forecast wrong. With this statement we provide another facet of a decades-old discussion on the relation of loss (or regret) and calibration (Section~\ref{Related Work: Linking Calibration and Loss}).

Scaled regret gambles, to the best of our knowledge, have not been used to construct an evaluation metric,\footnote{However, activation functions, such as introduced in \citep[\S~4.6]{cesa2006prediction}, allow for restricted scaling by $0$ or $1$.}, but, calibration gambles have been (Section~\ref{from gambles to calibration}). The equivalence in Corollary~\ref{corollary:Duality of Calibration and Regret} states that optimizing for an evaluation metric expressed through calibration gambles is equivalent to optimizing for an evaluation metric expressed through according scaled regret gambles. Hence, scaled regret gambles have implicitly been used already. However, unscaled regret gambles seem to be of more relevance to current machine learning as we observe in the following section. The relationship of regret-type evaluation metrics and calibration-type evaluation metrics is thus more subtle than Corollary~\ref{corollary:Duality of Calibration and Regret} suggests.


\section{Fair, Restricted and Rational Gamblers for Recovery of Evaluation Metrics}
\label{The Fair, Restricted and Rational Gambler}
We already observed that a small capital of a fair gambler does not necessarily guarantee ``good'' forecasts.  (a) A gambler playing constant negative gambles (which are available by default) does not prove anything about the forecasts. (b) A gambler who has access to arbitrarily scaleable gambles can always upscale a single gamble such that the resulting capital is essentially dominated by the result of this single gamble (Example~\ref{ex:Dominance of Single Gamble by Scaleability}).

Towards more reasonable evaluation metrics, we introduce two further conditions on the gambler, which turn out to be sufficient to elicit several known evaluation metrics. We demand the gambler to have a \emph{restricted} set from which its gambles can be chosen. One might interpret this as \emph{budget constraint}. This way gambles cannot be arbitrarily scaled.  Furthermore, we equip the gambler with a belief about the actual distribution from which the next outcome is realized. The gambler behaves \emph{rationally}, \ie, as an expected utility maximizer, with respect to its belief. One might interpret the belief as \emph{alternative hypothesis} about the distribution of the outcome. This way the gambler will not play unfavourable gambles such as constant negative ones. In conclusion, we obtain a two-dimensional structure of evaluation metrics based on the type of forecast, choice of belief and the restriction summarized in Table~\ref{tab:belief + restriction gives evaulation metric}.

For instance, the capital of a gambler which is restricted to play regret gambles and has a (true) belief about the average outcome is equal to external regret (Proposition~\ref{prop:Recovery of (Generalized) External Regret}. Or, the capital of gamblers with access to all gambles with a bounded norm and a (true) belief about the average outcome on the subsets on which a certain value has been predicted is aggregated to calibration scores (Proposition~\ref{prop:recovery of calibration score}). Roughly speaking, the belief expresses a certain knowledge which gets more and more refined in Table~\ref{tab:belief + restriction gives evaulation metric} from top to bottom (Proposition~\ref{prop:Bounding Capital by Refinment on Belief}), while the restriction gets less stringent in Table~\ref{tab:belief + restriction gives evaulation metric} from left to right (Proposition~\ref{prop:Bounding Capital by Hierarchy on Restrictions}). Besides that the recovery results justify the usefulness of the evaluation protocol in understanding current evaluation metrics, the above statements shed light on the debated duality of calibration scores and loss scores Section~\ref{Related Work: Linking Calibration and Loss}.

Let us formally introduce restriction and rationality for the gamblers in an evaluation protocol.
\begin{definition}[Fair, Restricted and Rational Gambler]
\label{def:fair, restricted and rational gambler}
    Let $(\Y, \Gamma, T, \forecaster, \sceptic^\indexsetgamblers, \nature)$ be an evaluation protocol. Fix an index $i \in \indexsetgamblers$. Let $\restrictionset \subseteq \CY$ such that $0 \in \restrictionset$. Let $\belief_t \in \Delta(\Y) \cup \{ \square \}$\footnote{We use $\square$ as a placeholder for a fully non-informative belief about the next outcome.} for every $t \in T$. The gambler $\sceptic^i$ in the evaluation protocol is called \emph{fair, restricted and rational} if for all $t \in T$, all of the following conditions are fulfilled:
    \begin{enumerate}[(a)]
        \item $\sceptic^i_t \in \offer_{\Gamma^{-1}(\forecaster_t)}$. (Availability, Fairness)
        \item $\sceptic^i_t \in \restrictionset$. (Restrictiveness)
        \item If $\belief_t \in \Delta(\Y)$, then
        \begin{align*}
            \mathbb{E}_{\belief_t}[\sceptic^i_t] \ge \mathbb{E}_{\belief_t}[g],
        \end{align*}
        for all $g \in \offer_{\Gamma^{-1}(\forecaster_t)} \cap \restrictionset$. (Rationality)
        \item If $\belief_t = \square$, then $\sceptic^i_t = 0$. (Opt-Out)
    \end{enumerate}
\end{definition}
We shortly discuss three key aspects of the newly introduced conditions on Gambler: (a) the option to withdraw from gambling, \ie, there is no obligation for Gambler to gamble. (b) the relationship of rationality and optimality with respect to an alternative hypothesis and (c) the content of the belief.

\paragraph{(a) No Obligation to Gamble.} The belief $\belief_t$ for some $t \in T$ is not necessarily a distribution, but as well can be equal to $\square$. If $\belief_t = \square$, then Gambler is not willing to commit to any precise belief. We use $\square$ as a placeholder for a fully non-informative belief about the next outcome. Assuming maxmin-rationality following \citep{gilboa1989maxmin}, we can recover the opt-out rule. The following proposition simply shows that the gamble $g^* = 0$ is an optimal action facing the full ambiguity of the own belief.
\begin{proposition}[Opt-Out is Maxmin-Rationality \wrt Vacuous Belief]
    Let $\restrictionset \subseteq \CY$ such that $0 \in \restrictionset$ and $\square = \Delta(\Y)$. If $g^* = 0$, then
    \begin{align*}
        \inf_{\phi \in \square} \mathbb{E}_{\phi}[g^*] \ge \inf_{\phi \in \square} \mathbb{E}_{\phi}[g],
    \end{align*}
    for all $g \in \offer_{\forecastingset} \cap \restrictionset$.
\end{proposition}
\begin{proof}
    By definition of $g^*$, $\inf_{\phi \in \square} \mathbb{E}_{\phi}[g^*] = 0$.
    Let us assume that,
    \begin{align*}
        \inf_{\phi \in \square} \mathbb{E}_{\phi}[g] > 0.
    \end{align*}
    It follows that $\inf_{y \in \Y} g(y) > 0$, because for every $y \in \Y$, there exists $\dirac(y) \in \square$. This is a contradiction, as $g \in \offer_{\forecastingset}$ and $\offer_{\forecastingset}$ is an offer (Theorem~\ref{thm:Available Gambles Form Offer}) with condition~\ref{def:offer - coherence}.
\end{proof}



\paragraph{(b) Rationality is Type-II Error Minimization.} The belief of Gambler is nothing else than an alternative hypothesis suggested against the ``null hypothesis'' made by Forecaster. The introduction of an alternative hypothesis for testing probabilistic statements is not new and at least goes back to the seminal work by \citet{neyman1933ix}. Essentially, alternative hypotheses try to control the Type-II error of a probabilistic statement. We can provide analogous interpretations to the conditions on Gambler: Availability roughly guarantees control over Type-I error, the false rejection of true forecasts. Rationality minimizes the Type-II error, the false acceptance of a wrong forecast. This correspondence is not surprising for readers familiar to the literature on e-values. The availability criterion is captured in the definition of e-value itself, while rationality is expressed through growth rate optimality \citep{grunwald2020safe}. There are some central differences, though. For a detailed discussion see Section~\ref{Related Work - Testing Forecasts}.

\paragraph{(c) Belief can be True.} As it turns out in Section~\ref{recovery of existing evaluation notions}, many existing evaluation metrics summarize the capital of the gamblers which play ``as if'' they would know parts of the outcomes of Nature. Usually, alternative hypothesis express external knowledge about a certain phenomenon. We will leverage our a posteriori perspective to align belief to truth. If the beliefs of an agent are aligned to truth, the beliefs can be called \emph{knowledge}.
\begin{definition}[Alignment to Truth]
\label{def:alignment to truth}
    Let $(\Y, \Gamma, T, \forecaster, \{ \sceptic\}, \nature)$ be an evaluation protocol and denote by $\belief_t \in \Delta(\Y) \cup \{ \square \}$ for every $t \in T$ the belief of $\sceptic$. We define $T_\beta \coloneqq \{ t \in T \colon \belief_t = \beta\}$.
    The beliefs $(\belief_t)_{t \in T}$ are \emph{aligned to truth}, if for all $\beta \in \{ \belief_t \colon t \in T\}$, $\beta = \square$ or $\beta = \frac{1}{|T_\beta|} \sum_{t \in T_\beta} \dirac(\nature_t) \in \Delta(\Y)$.
\end{definition}

\begin{figure}
    \centering
    \subfigure[]{\includegraphics[width=0.3\textwidth]{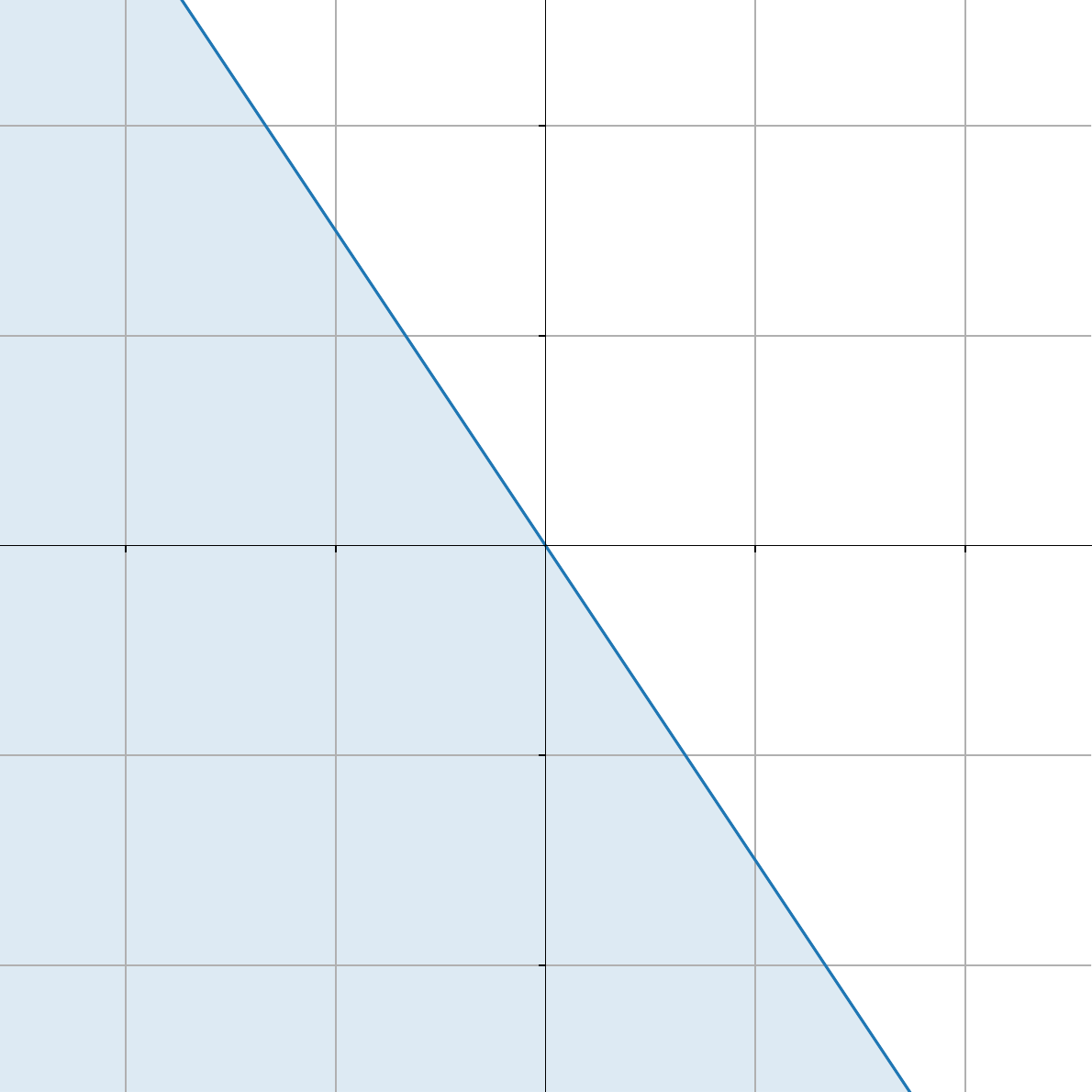}}
    \subfigure[]{\includegraphics[width=0.3\textwidth]{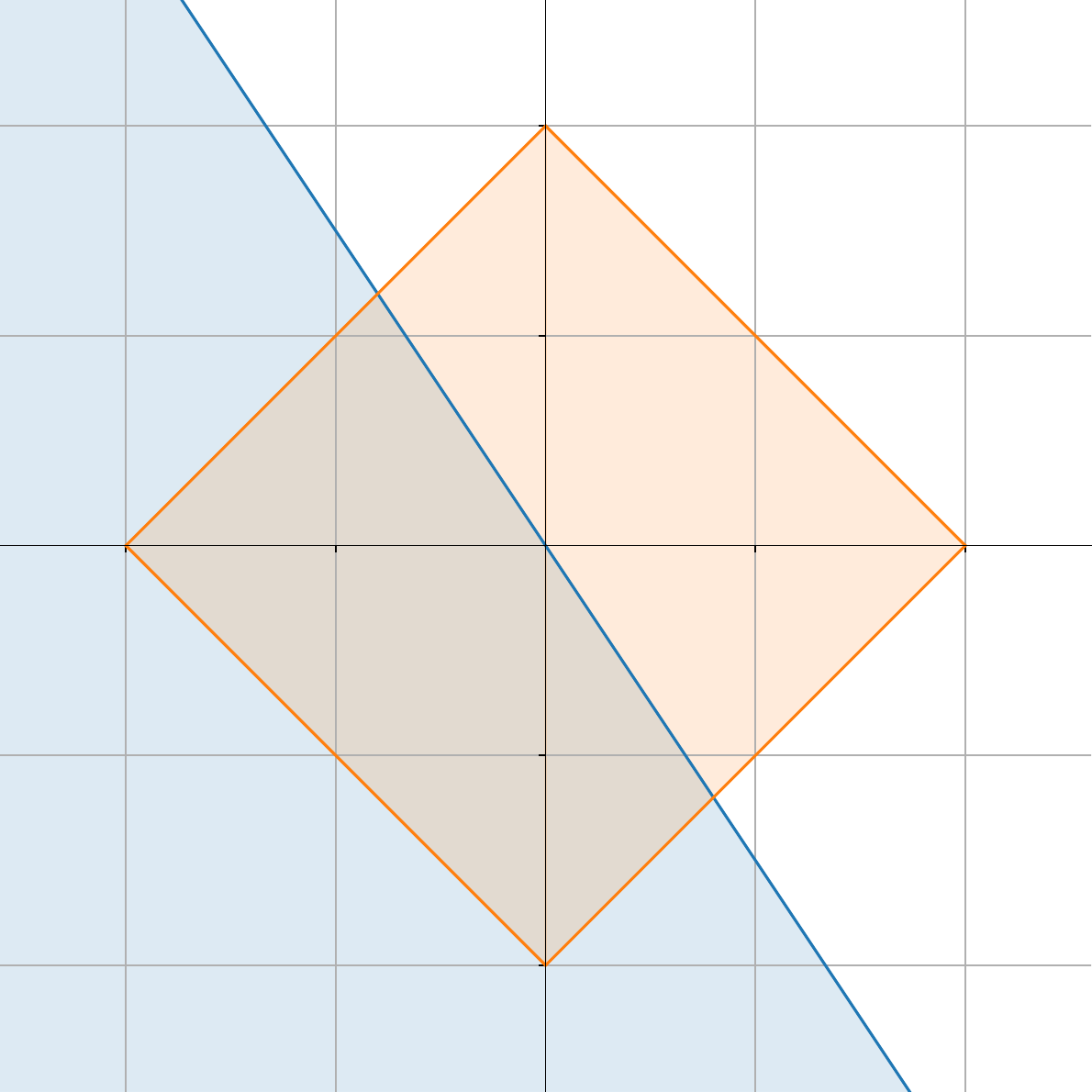}}
    \subfigure[]{\includegraphics[width=0.3\textwidth]{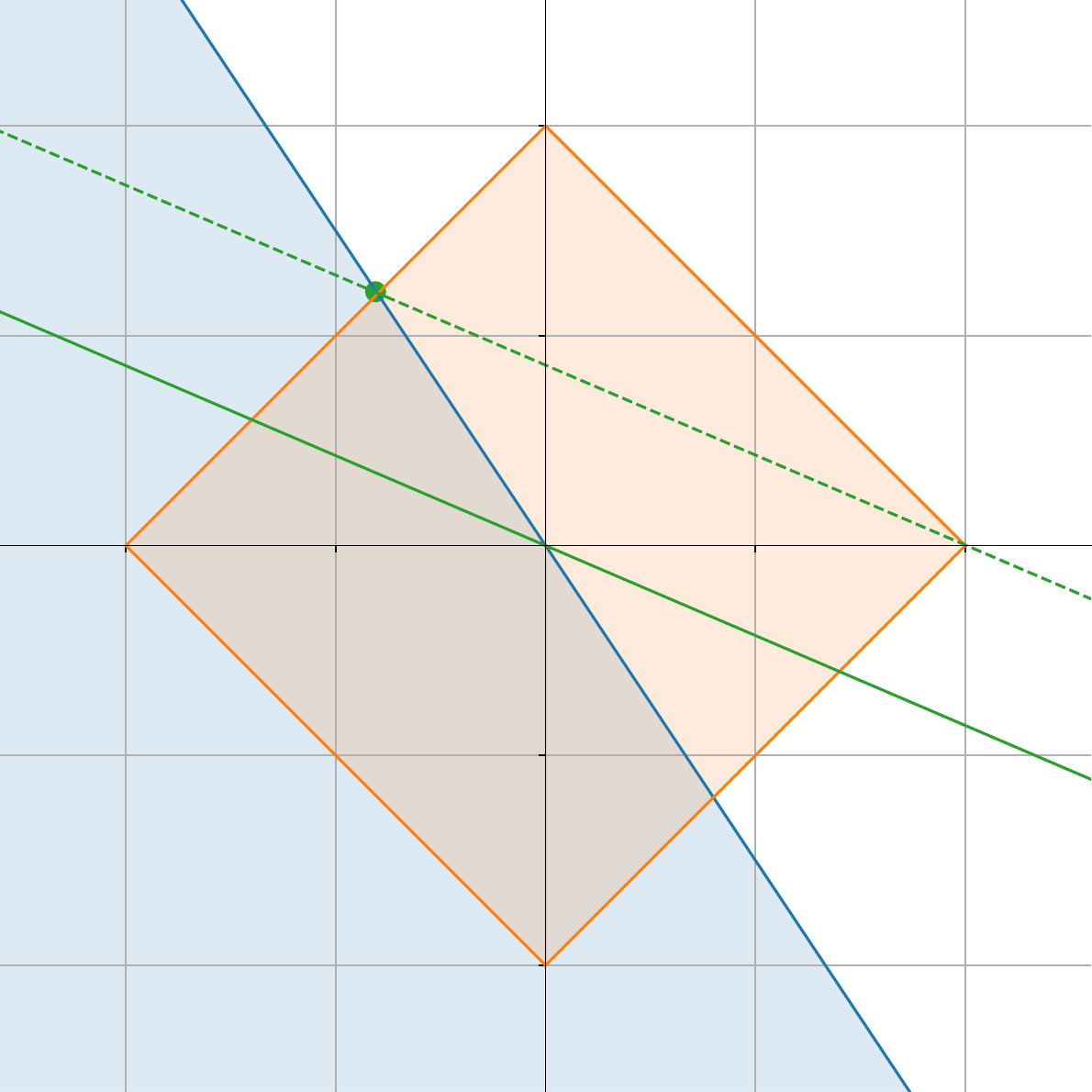}}
    \caption{Illustration of a fair, restricted and rational gambler. -- (a) Exemplary set of available gambles for a two-element outcome set (blue). (b) Restriction to norm ball $\ball_1$ (orange). (c) Optimal choice of gamble (green dot) with respect to a belief (green line).}
    \label{fig:fair+restricted+rational}
\end{figure}

We illustrate the three conditions on the gambler in Figure~\ref{fig:fair+restricted+rational}.

Finally, we provide a simple derivative of a fair, restricted and rational gambler, which is only approximately rational. In fact, this definition let's us ignore the $\pqtopology$-closure of sets in later results (\eg, Proposition~\ref{prop:Recovery of (Generalized) Multicalibration Score}).
\begin{definition}[Fair, Restricted and Approximately Rational Gambler]
\label{def:fair restricted and approximate rational gambler}
Let\\ $(\Y, \Gamma, T, \forecaster, \sceptic^\indexsetgamblers, \nature)$ be an evaluation protocol. Fix an index $i \in \indexsetgamblers$. Let $\restrictionset \subseteq \CY$ such that $0 \in \restrictionset$. Let $\belief_t \in \Delta(\Y) \cup \{ \square \}$ for every $t \in T$. The $i$th gambler $\sceptic^i$ in the evaluation protocol is called \emph{fair, restricted and approximately rational} if for all $t \in T$, all of the following conditions are fulfilled:
    \begin{enumerate}[(a)]
        \item For all $t \in T$, $\sceptic^i_t \in \offer_{\Gamma^{-1}(\forecaster_t)}$. (Availability, Fairness)
        \item For all $t \in T$, $\sceptic^i_t \in \restrictionset$. (Restrictiveness)
        \item For all $t \in T$ with $\belief_t \in \Delta(\Y)$, then
        \begin{align*}
            \mathbb{E}_{\belief_t}[\sceptic^i_t] \ge \mathbb{E}_{\belief_t}[g] - \epsilon,
        \end{align*}
        for all $\epsilon > 0$ and $g \in \offer_{\Gamma^{-1}(\forecaster_t)} \cap \restrictionset$. (Approximate Rationality)
        \item For all $t \in T$ with $\belief_t = \square$, then $\sceptic^i_t = 0$. (Opt-Out)
    \end{enumerate}
\end{definition}

\subsection{Recovery of Existing Evaluation Metrics}
\label{recovery of existing evaluation notions}
Machine learning scholars have introduced many notions which try to capture the quality of forecasts \citep{cesa2006prediction, williamson2022geometry, derr2025three}. We suggest that our approach via evaluation protocols (Definition~\ref{def:evaluation protocol}) and fair, restricted and rational gamblers (Definition~\ref{def:fair, restricted and rational gambler}) provides an insightful framework to recover and structure several of those notions.

In the following, we provide a series of results, which follow a simple scheme summarized in Proposition~\ref{prop:schematic recovery proposition}.
\begin{proposition}[Schematic Recovery Proposition]
\label{prop:schematic recovery proposition}
    Set up the evaluation protocol, in particular, the outcome space $\Y$ and the predicted property $\Gamma$. Define several gamblers by:
    \begin{enumerate}[(a)]
        \item Defining Gambler's beliefs.
        \item Restricting the allowed gambles.
    \end{enumerate}
    The aggregated capital among the gamblers is equal to some known metric of evaluation.
\end{proposition}

\subsection{From Gambles to Regret}
\label{From Gambles to Regret}
Let us start with a simple finger exercise. We show that within the evaluation protocol with a certain choice of beliefs and restrictions, we can recover external regret. The proof is relatively direct and largely builds upon a specially structured restriction, which bounds the gambler to (unscaled) regret gambles.
\begin{proposition}[Recovery of External Regret]
\label{prop:Recovery of (Generalized) External Regret}
    Let $(\Y, \Gamma, T, \forecaster, \{\sceptic\}, \nature)$ be an evaluation protocol with $\Gamma\colon \Q \to 2^\propspace$ being an elicitable property with consistent scoring function $\loss$. We define a fair, restricted and rational gambler $\sceptic$ such that:
    \begin{enumerate}[(a)]
        \item On every instance $t \in T$ the gambler $\sceptic$ has a (true) belief $\belief_t$ about the average outcome distribution,
        \begin{align*}
            \belief_t = \hat{D}_{Y} \coloneqq \frac{1}{|T|}\sum_{t' \in T} \dirac(\nature_{t'}) \in \Q.
        \end{align*}
        \item On every instance $t \in T$ the gambler is restricted to $\UnscaledRegretGambles_{\forecaster_t} \coloneqq \{ \loss_{\forecaster_t} - \loss_c\colon c \in \propspace\}$.
    \end{enumerate}
    Then, the reweighted capital of the gambler $\sceptic$ is equal to,
    \begin{align*}
        |T| \capital(\sceptic) = \sum_{t \in T} \loss(\nature_t, \forecaster_t) - \min_{c \in \propspace}\loss(\nature_t, c),
    \end{align*}
    which is external regret.
\end{proposition}
\begin{proof}
    Let us consider a fixed instance $t \in T$. Given the gambler's belief, the fair, restricted and rational gambler plays (Lemma~\ref{lemma:Optimal Gamble for Rational Gambler Restricted to Regret Gambles}),
    \begin{align*}
        \sceptic_t \colon y \mapsto \loss(y, \forecaster_t) - \loss(y, \Gamma(\hat{D}_{Y})).
    \end{align*}
    In total, we obtain the following aggregated capital for Gambler $\sceptic$,
    \begin{align*}
        \capital(\sceptic)
        =\frac{1}{|T|}\sum_{t \in T} \sceptic_t(\nature_t)
        =\frac{1}{|T|}\sum_{t \in T} \loss(\nature_t, \forecaster_t) - \loss(\nature_t, \Gamma(\hat{D}_{Y})).
    \end{align*}
    The statement follows.
\end{proof}
\begin{lemma}[Gamble for Rational Gambler Restricted to Regret Gambles]
\label{lemma:Optimal Gamble for Rational Gambler Restricted to Regret Gambles}
    Let\\$(\Y, \Gamma, T, \forecaster, \sceptic^\indexsetgamblers, \nature)$ be an evaluation protocol with $\Gamma\colon \Q \to 2^\propspace$ being an elicitable property with consistent scoring function $\loss$. Fix $t \in T$ and consider a fair, restricted and rational gambler $\sceptic^i$ with belief $\belief_t \in \Q$ and restriction $\UnscaledRegretGambles_{\forecaster_t} \coloneqq \{ \loss_{\forecaster_t} - \loss_c\colon c \in \propspace\}$. Then, in this round $t \in T$, the fair, restricted and rational gambler $\sceptic^i$ plays the gamble,
    \begin{align*}
        \sceptic^i_t \colon y \mapsto \loss(y, \forecaster_t) - \loss(y, \Gamma(\belief_t)).
    \end{align*}
\end{lemma}
\begin{proof}
    The statement follows because,
    \begin{align*}
        \mathbb{E}_{\belief_t}[\sceptic^i_t]
        &= \mathbb{E}_{ \belief_t}[ \loss_{\forecaster_t} - \loss_{ \Gamma(\belief_t)} ]\\
        &\overset{D\ref{def:scoring function}}{\ge} \max_{c \in \propspace} \mathbb{E}_{\belief_t  }[ \loss_{\forecaster_t} - \loss_{c}]\\
        &= \max_{g \in \UnscaledRegretGambles_{\forecaster_t}} \mathbb{E}_{\belief_t }[ g]\\
        &\overset{P\ref{prop:regret gambles are available}}{=} \max_{g \in \offer_{\Gamma^{-1}(\forecaster_t)} \cap \UnscaledRegretGambles_{\forecaster_t}} \mathbb{E}_{\belief_t}[ g].
    \end{align*}
\end{proof}
Arguably, the knowledge of Gambler is ``too'' coarse to provide a hard benchmark. Hence, we suggest to make the beliefs more ``fine-grained'' (\cf Definition~\ref{def:refinement of belief}). In Proposition~\ref{prop:recovery of swap regret} we equip a set of gamblers with a true belief about the average outcome on all instances on which the forecaster forecasted a certain fixed valued, which lets us recover swap regret (\cf Example 4.4 in \citep{cesa2006prediction}).
\begin{proposition}[Recovery of Swap Regret]
\label{prop:recovery of swap regret}
Let $\forecaster_T \coloneqq \{ \forecaster_t\colon t \in T \}$ and $\Gamma\colon \Q \to 2^\propspace$ be an elicitable property with consistent scoring function $\loss$. Let
$(\Y, \Gamma, T, \forecaster, \sceptic^{\forecaster_T}, \nature)$ be an evaluation protocol. For every $\gamma \in \forecaster_T$, we define a fair, restricted and rational gambler $\sceptic^\gamma$ such that:
    \begin{enumerate}[(a)]
        \item On all instances $T_\gamma \coloneqq \{ t \in T\colon \forecaster_t = \gamma \} $, the gambler $\sceptic^\gamma$ has a (true) belief $\belief^\gamma_t$ about the average outcome distribution,
        \begin{align*}
            \belief^\gamma_t = \hat{D}_{Y|\gamma} \coloneqq \frac{1}{|T_\gamma|}\sum_{t' \in T_\gamma} \dirac(\nature_{t'}) \in \Q.
        \end{align*}
        On all other instances $T \setminus T_\gamma$, the gambler has belief $\belief^\gamma_t = \square$.
        \item The gambler is restricted to play gambles in $\UnscaledRegretGambles_\gamma \coloneqq \{ \loss_{\gamma} - \loss_c\colon c \in \propspace\}$.
    \end{enumerate}
    Then, the re-weighted (by $|T|$) and summed up capitals of all gamblers $\sceptic^\gamma$ is equal to,
    \begin{align*}
        |T | \sum_{\gamma \in \forecaster_T}  \capital(\sceptic^\gamma) =  \max_{\sigma \in \{ \propspace \to \propspace \} } \sum_{t \in T} \loss(\nature_t, \forecaster_t) - \loss(\nature_t, \sigma(\forecaster_t)),
    \end{align*}
    which is the definition of swap regret following \citep[Example 4.4]{cesa2006prediction}.
\end{proposition}

\begin{proof}
    Let us fix a gambler $\sceptic^\gamma$ for some $\gamma \in  \forecaster_T$. Consider a fixed instance $t \in T_\gamma$. 
    Given the gambler's belief, the fair, restricted and rational gambler plays (Lemma~\ref{lemma:Optimal Gamble for Rational Gambler Restricted to Regret Gambles}),,
    \begin{align*}
        \sceptic^\gamma_t \colon y \mapsto \loss(y, \gamma) - \loss(y, \Gamma(\hat{D}_{Y|\gamma})).
    \end{align*}
    For any fixed instance $t \in T \setminus T_\gamma$, the gambler's belief $\belief^\gamma_t =\square$ is vacuous, hence by Definition~\ref{def:fair, restricted and rational gambler}, $\sceptic^\gamma_t = 0$. In total, we obtain the following aggregated capital for a gambler $\sceptic^\gamma$,
    \begin{align*}
        \capital(\sceptic^\gamma)
        = \frac{1}{|T|}\sum_{t \in T_\gamma} \sceptic^\gamma_t(\nature_t)
        = \frac{1}{|T|}\sum_{t \in T_\gamma} \loss(\nature_t, \forecaster_t) - \loss(\nature_t, \Gamma(\hat{D}_{Y|\gamma}).
    \end{align*}
    The statement follows by aggregating the capitals of the different gamblers,
    \begin{align*}
        |T| \sum_{\gamma \in \forecaster_T} \capital(\sceptic^\gamma)
        &= \sum_{\gamma \in \forecaster_T} \sum_{t \in T_\gamma} \loss(\nature_t, \forecaster_t) - \loss(\nature_t, \Gamma(\hat{D}_{Y|\gamma})\\
        &= \sum_{\gamma \in \forecaster_T} \max_{c \in \propspace} \sum_{t \in T_\gamma} \loss(\nature_t, \forecaster_t) - \loss(\nature_t, c)\\
        &= \max_{\sigma \in \{ \propspace \to \propspace \} } \sum_{\gamma \in \forecaster_T} \sum_{t \in T_\gamma} \loss(\nature_t, \forecaster_t) - \loss(\nature_t, \sigma(\forecaster_t))\\
        &= \max_{\sigma \in \{ \propspace \to \propspace \} } \sum_{t \in T} \loss(\nature_t, \forecaster_t) - \loss(\nature_t, \sigma(\forecaster_t)).
    \end{align*}
\end{proof}
By refining the belief to group-wise knowledge about the average outcome given a prediction we obtain a notion which we term \emph{group-wise swap regret score}. This notion is analogous to multicalibration (Section~\ref{Across Regret and Calibration}).
\begin{proposition}[Recovery of Group-Wise Swap Regret Score]
\label{prop:Recovery of Group-Wise Swap Regret}
    Let $\mathcal{S} \subseteq 2^T$ be a set of groups and $\forecaster_T \coloneqq \{ \forecaster_t \colon t \in T\}$. Let $(\Y, \Gamma, T, \forecaster, \sceptic^{\forecaster_T \times \mathcal{S}}, \nature)$ be an evaluation protocol with $\Gamma\colon \Q \to 2^\propspace$ being an elicitable property with consistent scoring function $\loss$.
    For every $\gamma \in \forecaster_T$ and $S \in \mathcal{S}$ we define a fair, restricted and approximate rational gambler $\sceptic^{\gamma,S}$ such that:
    \begin{enumerate}[(a)]
        \item On all instances $T_{\gamma, S} \coloneqq \{ t \in T\colon \forecaster_t =\gamma\} \cap S$, the gambler $\sceptic^{\gamma, S}$ has a (true) belief $\belief^{\gamma, S}_t$ about the average outcome distribution,
        \begin{align*}
            \belief^{\gamma, S}_t = \hat{D}_{Y|{\gamma, S}} \coloneqq \frac{1}{|T_{\gamma, S}|}\sum_{t' \in T_{\gamma, S}} \dirac(\nature_{t'}) \in \Q.
        \end{align*}
        On all other instances $T \setminus T_{{\gamma, S}}$, the gambler has no belief.
        \item The gambler is restricted to play gambles in $\UnscaledRegretGambles_\gamma \coloneqq \{ \loss_{\gamma} - \loss_c\colon c \in \propspace\}$.
    \end{enumerate}
    Then, the reweighted and aggregated capital of the gamblers $\sceptic^{\gamma, S}$ is equal to,
    \begin{align*}
        \sup_{S \in \mathcal{S}} \sum_{\gamma \in \forecaster_T} |\capital(\sceptic^{\gamma, S})|
        &= \frac{1}{|T|} \sup_{S \in \mathcal{S}} \max_{\sigma \in \{ \propspace \to \propspace \} } \sum_{t \in S} \loss(\nature_t, \forecaster_t) - \loss(\nature_t, \sigma(\forecaster_t)),
    \end{align*}
    what we call \emph{group-wise swap regret score}.
\end{proposition}
\begin{proof}
    Fix $\gamma \in \forecaster_T$ and $S \in \mathcal{S}$. Let us consider a fixed instance $t \in T_{\gamma, S}$.
    Given the gambler's belief, the fair, restricted and rational gambler plays (Lemma~\ref{lemma:Optimal Gamble for Rational Gambler Restricted to Regret Gambles}),
    \begin{align*}
        \sceptic^{\gamma,S}_t \colon y \mapsto \loss(y, \gamma) - \loss(y, \Gamma(\hat{D}_{Y|\gamma,S})).
    \end{align*}
    For any fixed instance $t \in T \setminus T_{\gamma, S}$, the gambler's belief $\belief^{\gamma,S}_t =\square$ is vacuous, hence by Definition~\ref{def:fair, restricted and rational gambler}, $\sceptic^{\gamma,S}_t = 0$. In total, we obtain the following aggregated capital for the gambler $\sceptic^{\gamma,S}$,
    \begin{align*}
        \capital(\sceptic^{\gamma,S})
        = \frac{1}{|T|}\sum_{t \in T_{\gamma,S}} \sceptic^\gamma_t(\nature_t)
        = \frac{1}{|T|}\sum_{t \in T_{\gamma,S}} \loss(\nature_t, \gamma) - \loss(\nature_t, \Gamma(\hat{D}_{Y|{\gamma,S}})).
    \end{align*}
    The statement follows,
    \begin{align*}
        \sup_{S \in \mathcal{S}} \sum_{\gamma \in \forecaster_T} \left|\capital(\sceptic^{\gamma, S})\right|
        &=\sup_{S \in \mathcal{S}} \sum_{\gamma \in \forecaster_T}  \left|\frac{1}{|T|}\sum_{t \in T_{\gamma,S}}\loss(\nature_t, \gamma) - \loss(\nature_t, \Gamma(\hat{D}_{Y|{\gamma,S}})) \right|\\
        &=\sup_{S \in \mathcal{S}} \sum_{\gamma \in \forecaster_T}  \frac{1}{|T|}\max_{c \in \propspace} \sum_{t \in T_{\gamma,S}} \loss(\nature_t, \forecaster_t) - \loss(\nature_t, c)\\
        &=\frac{1}{|T|} \sup_{S \in \mathcal{S}} \max_{\sigma \in \{ \propspace \to \propspace \} } \sum_{\gamma \in \forecaster_T} \sum_{t \in T_{\gamma,S}} \loss(\nature_t, \forecaster_t) - \loss(\nature_t, \sigma(\forecaster_t))\\
        &=\frac{1}{|T|} \sup_{S \in \mathcal{S}} \max_{\sigma \in \{ \propspace \to \propspace \} } \sum_{t \in S} \loss(\nature_t, \forecaster_t) - \loss(\nature_t, \sigma(\forecaster_t)).
    \end{align*}
\end{proof}
Finally, we push the knowledge of a gambler to the extreme and assume it is ``clairvoyant''. This way we recover standard, but zeroed, loss scores. Arguably, that is the most widely used evaluation metric in machine learning and often summarized in benchmark tables, \eg, accuracy, false positive rate (FPR), false negative rate (FNR), mean squared error (MSE), Brier score, cross entropy loss. The zeroed form of the loss scores might seem strange on the first sight. It is the average loss of the forecaster subtracted by the loss a forecaster which has access to the next outcome would have. For most loss functions, if the forecaster knows the outcome, then the loss is equal to $0$. Formally, $\loss(y, \Gamma(\dirac(y)) = 0$ for all $y \in \Y$. This is true for instance for the Brier score, accuracy or cross entropy loss. Hence, we call the scores ``zeroed'' as the perfect forecaster would achieve loss score $0$.
\begin{proposition}[Recovery of Zeroed Loss Scores]
\label{prop:recovery of loss scores}
    Let $(\Y, \Gamma, T, \forecaster, \{\sceptic\}, \nature)$ be an evaluation protocol with $\Gamma\colon \Q \to 2^\propspace$ being an elicitable property with consistent scoring function $\loss$ and fair, restricted and rational gambler $\sceptic$ such that:
    \begin{enumerate}[(a)]
        \item On every instance $t \in T$ the gambler $\sceptic$ is ``clairvoyant'', \ie, has a (true) belief $\belief_t$ about the outcome revealed by Nature, \ie, $\belief_t = \dirac(\nature_t) \in \Q$.
        \item The gambler is restricted to play gambles in $\UnscaledRegretGambles_{\forecaster_t} \coloneqq \{\loss_{\forecaster_t} - \loss_c\colon c \in \propspace\}$.
    \end{enumerate}
    Then, the capital of the gambler $\sceptic$ is equal to,
    \begin{align*}
        \capital(\sceptic) = \frac{1}{|T|} \sum_{t \in T} \loss(\nature_t, \forecaster_t) - \frac{1}{|T|} \sum_{t \in T}\loss(\nature_t, \Gamma(\dirac(\nature_t)).
    \end{align*}
\end{proposition}
\begin{proof}
    Let us consider a fixed instance $t \in T$. Given the gambler's belief, the fair, restricted and rational gambler plays (Lemma~\ref{lemma:Optimal Gamble for Rational Gambler Restricted to Regret Gambles}),
    \begin{align*}
        \sceptic_t \colon y \mapsto \loss(y, \forecaster_t) - \loss(y, \Gamma(\dirac(\nature_t)).
    \end{align*}
    In total, we obtain the following aggregated capital for Gambler $\sceptic$,
    \begin{align*}
        \capital(\sceptic)
        = \frac{1}{|T|}\sum_{t \in T} \sceptic_t(\nature_t)
        = \frac{1}{|T|}\sum_{t \in T}\loss(\nature_t,  \forecaster_t) - \loss(\nature_t, \Gamma(\dirac(\nature_t)).
    \end{align*}
\end{proof}
In all of the above propositions, the reader might remain disappointed in that the restriction to $\UnscaledRegretGambles_{\forecaster_t}$ (respectively $\UnscaledRegretGambles_\gamma$) artificially forces the gambler to recover certain types of regret. Clearly, regret gambles are available. However, they form only one particular choice for evaluating a forecast within the framework of a fair evaluation protocol. Hence, we are not exploiting the full freedom of available gambles.

In the next section, the restriction of the gambles will appear more natural. In particular, the results will require a full characterization of available gambles.\footnote{The use of the full characterization is marked by reference to Theorem~\ref{thm:Characterization of Available Gambles in Protocol 2 with Identifiable Property} (respectively Corollary~\ref{corollary:Characterization of Available Gambles of Identifiable Property-Forecasts in Finite Dimensions}).}

\subsection{From Gambles to Calibration}
\label{from gambles to calibration}
The following recovery results share with the above the order in which the belief of the gambler(s) is refined step-by-step. However, we consider another restricted set of gambles from which Gambler can choose which is defined through the scaled norm ball $\ball_{\alpha} \coloneqq \{ g \in \CY\colon \|g \| \le \alpha \}$ for some norm $\|\cdot \| $ on $\CY$ and we assume that the predicted property $\Gamma$ is identifiable. In particular, the first two statements consider mean predictors in the binary setting. The reason for this is that most of the existing notions focus on this limited setup.

First, we recover bias in the large (for binary outcomes) following \citep{murphy1967verification} (for more modern formulations see \citep{vovk2005defensive} or the generalization to subgroups called ``accuracy in expectation'' \citep{hebert2018multicalibration}\footnote{We can recover the exact formulation of ``accuracy in expectation'' for a subgroup by simply assuming that the gambler in Proposition~\ref{prop:recovery of bias in the large} has a true belief about the average outcome on the subgroup, and a vacuous belief, \ie, $\square$, outside the subgroup (\cf Proposition~\ref{prop:Recovery of (Generalized) Multicalibration Score})}).
\begin{proposition}[Recovery of Bias in the Large]
\label{prop:recovery of bias in the large}
    Let $(\Y, \Gamma, T, \forecaster, \{\sceptic\}, \nature)$ be an evaluation protocol with $\Y = \{0,1\} $ and $\Gamma$ being the mean. We define a fair, restricted and rational gambler $\sceptic$ such that:
    \begin{enumerate}[(a)]
        \item On every instance $t \in T$, the gambler $\sceptic$ has a (true) belief $\belief_t$ about the average outcome distribution,
        \begin{align*}
            \belief_t = \hat{D}_{Y} \coloneqq \frac{1}{|T|}\sum_{t' \in T} \dirac(\nature_{t'}) \in \Delta(\Y).
        \end{align*}
        \item The gambler is restricted to play gambles in $\ball_1 \coloneqq \{ g \in \CY\colon \|g \|_1 \le 1 \}$ where $\| \cdot\|_1$ is the $l_1$-norm.\footnote{The $l_1$-norm is well-defined as $\CY = \reals^2$ here.}
    \end{enumerate}
    Then, the capital of the gambler $\sceptic$ is equal to,
    \begin{align*}
        \capital(\sceptic) = \left|\frac{1}{|T|}\sum_{t \in T} (\nature_t - \forecaster_t) \right|,
    \end{align*}
    which is the definition of bias in the large for binary outcomes following \citep{murphy1967verification}.
\end{proposition}
\begin{proof}
    Let us consider a fixed instance $t \in T_S$. Given the gambler's belief, the fair, restricted and rational gambler plays (Lemma~\ref{lemma:Optimal Gamble for Rational Gambler Restricted to Calibration Gambles I}),
    \begin{align*}
        \sceptic_t
        &\colon \begin{cases}
            y \mapsto y - \forecaster_t \text{ if } \mathbb{E}_{\hat{D}_{Y}}[Y] > \forecaster_t\\
            y \mapsto \forecaster_t - y \text{ otherwise}.
        \end{cases}
    \end{align*}
    In total, we obtain the following aggregated capital for the gambler $\sceptic$,
    \begin{align*}
        \capital(\sceptic)
        &=\frac{1}{|T|}\sum_{t \in T} \begin{cases}
            \nature_t - \forecaster_t \text{ if } \mathbb{E}_{\hat{D}_{Y}}[Y] > \gamma\\
            \forecaster_t - \nature_t \text{ otherwise}.
        \end{cases}\\
        &= \frac{1}{|T|}\left|\sum_{t \in T} (\nature_t - \forecaster_t)\right|.
    \end{align*}
\end{proof}
\begin{lemma}[Gamble for Rational Gambler Restricted to Norm Ball - Binary $\Y$]
\label{lemma:Optimal Gamble for Rational Gambler Restricted to Calibration Gambles I}
    Let $(\Y, \Gamma, T, \forecaster, \{\sceptic\}, \nature)$ be an evaluation protocol with $\Y = \{0,1\} $ and $\Gamma$ being the mean. Fix $t \in T$ and consider a fair, restricted and rational gambler $\sceptic$ with belief $\belief_t \in \Delta(\Y)$ and restriction $\ball_1 \coloneqq \{ g \in \CY\colon \|g \|_1 \le 1 \}$. Then, in this round $t \in T$, the fair, restricted and rational gambler $\sceptic$ plays the gamble,
    \begin{align*}
        \sceptic_t
        &\colon \begin{cases}
            y \mapsto y - \forecaster_t \text{ if } \mathbb{E}_{\belief_t}[Y] > \forecaster_t\\
            y \mapsto \forecaster_t - y \text{ otherwise}.
        \end{cases}.
    \end{align*}
\end{lemma}
\begin{proof}
    The statement follows because,
    \begin{align*}
        \mathbb{E}_{\belief_t}[\sceptic_t]
        &= \left|\mathbb{E}_{\belief_t}[Y] - \forecaster_t \right|\\
        &= \max_{\alpha \in [-1,1]} \alpha \left(\mathbb{E}_{\belief_t}[Y] - \forecaster_t \right)\\
        &= \max_{\alpha \in [-1,1]} \mathbb{E}_{\belief_t}[\alpha \ident_{\forecaster_t}]\\
        &\overset{(*)}{=} \max_{\alpha \in \reals \text{ and } \|\alpha \ident_{\forecaster_t} \|_1 \le 1 } \mathbb{E}_{\belief_t}[\alpha \ident_{\forecaster_t}]\\
        &\ge \max_{g \in \{ g \in \CY \colon g \le \alpha \ident_{\forecaster_t}, \alpha \in \reals \} \cap \ball_1 } \mathbb{E}_{\belief_t}[g]\\
        &\overset{(C\ref{corollary:Characterization of Available Gambles of Identifiable Property-Forecasts in Finite Dimensions})}{=} \max_{g \in \offer_{\Gamma^{-1}(\gamma)} \cap \ball_1 } \mathbb{E}_{\belief_t}[g].
    \end{align*}
    Equation $(*)$ is true because $\|\alpha \ident_\gamma \|_1 = |\alpha| \| \ident_\gamma \|_1 = |\alpha|$ for all $\gamma \in \propspace$. 
\end{proof}
It is not a far way to the recovery of bias in the small \citep{murphy1967verification}, better known as standard calibration. The beliefs' of the gambler in the following proposition are equal to the beliefs' of the gamblers in Proposition~\ref{prop:recovery of swap regret}.
\begin{proposition}[Recovery of Calibration Score]
\label{prop:recovery of calibration score}
    Let $\forecaster_T \coloneqq \{ \forecaster_t \colon t \in T\}$, $\Y = \{0,1\} $ and $\Gamma$ being the mean. Let $(\Y, \Gamma, T, \forecaster, \sceptic^{\forecaster_T}, \nature)$ be an evaluation protocol. For every $\gamma \in  \forecaster_T$, we define a fair, restricted and rational gambler $\sceptic^\gamma$ such that:
    \begin{enumerate}[(a)]
        \item On all instances $T_\gamma \coloneqq \{ t \in T\colon \forecaster_t = \gamma \} $, the gambler $\sceptic^\gamma$ has a (true) belief $\belief^\gamma_t$ about the average outcome distribution,
        \begin{align*}
            \belief^\gamma_t = \hat{D}_{Y|\gamma} \coloneqq \frac{1}{|T_\gamma|}\sum_{t' \in T_\gamma} \dirac(\nature_{t'}) \in \Delta(\Y).
        \end{align*}
        On all other instances $T \setminus T_\gamma$, the gambler has belief $\belief^\gamma_t = \square$.
        \item The gambler is restricted to play gambles in $\ball_1 \coloneqq \{ g \in \CY\colon \|g \|_1 \le 1 \}$ where $\| \cdot\|_1$ is the $l_1$-norm.\footnote{The $l_1$-norm is well-defined as $\CY = \reals^2$ here.}
    \end{enumerate}
    Then, the re-weighted (by $\frac{|T|}{|T_\gamma|}$) and squared capitals of all gamblers $\sceptic^\gamma$ is equal to,
    \begin{align}
    \label{eq:foster vohra calibration (l2)}
        \sum_{\gamma \in \forecaster_T} \frac{|T|}{|T_\gamma|} \capital(\sceptic^\gamma)^2 = \sum_{\gamma \in \forecaster_T} \frac{|T_\gamma|}{|T|}\left(\frac{1}{|T_\gamma|}\sum_{t \in T_\gamma} \nature_t - \gamma\right)^2,
    \end{align}
    which is the definition of calibration score following \citep{foster1998asymptotic}. Furthermore, the accumulated, absolute capitals of all gamblers $\sceptic^\gamma$ indexed by $\gamma$ is equal to,
    \begin{align}
    \label{eq: ECE (l1)}
        \sum_{\gamma \in \forecaster_T} \left| \capital(\sceptic^\gamma) \right| = \sum_{\gamma \in \forecaster_T} \frac{|T_\gamma|}{|T|} \left| \frac{1}{|T_\gamma|}\sum_{t \in T_\gamma} \nature_t - \gamma\right|,
    \end{align}
    which is the definition of expected calibration score (ECE), \eg, following \citep[Definition 3.1]{gopalan2022loss}.
\end{proposition}
\begin{proof}
    Let us fix a gambler $\sceptic^\gamma$ for some $\gamma \in  \forecaster_T$. Let us consider a fixed instance $t \in T_\gamma$. Given the gambler's belief, the fair, restricted and rational gambler plays (Lemma~\ref{lemma:Optimal Gamble for Rational Gambler Restricted to Calibration Gambles I}),
    \begin{align*}
        \sceptic^\gamma_t
        &\colon \begin{cases}
            y \mapsto y - \gamma \text{ if } \mathbb{E}_{\hat{D}_{Y|\gamma}}[Y] > \gamma\\
            y \mapsto \gamma - y \text{ otherwise},
        \end{cases}
    \end{align*}
    For any fixed instance $t \in T \setminus T_\gamma$, the gambler's belief $\belief^\gamma_t =\square$ is vacuous, hence by Definition~\ref{def:fair, restricted and rational gambler}, $\sceptic^\gamma_t = 0$. In total, we obtain the following aggregated capital for the gambler $\sceptic^\gamma$,
    \begin{align*}
        \capital(\sceptic^\gamma)
        &= \frac{1}{|T|}\sum_{t \in T_\gamma} \sceptic^\gamma_t(\nature_t)\\
        &= \frac{1}{|T|}\sum_{t \in T_\gamma} \begin{cases}
            \nature_t - \gamma \text{ if } \mathbb{E}_{\hat{D}_{Y|\gamma}}[Y] > \gamma\\
            \gamma - \nature_t \text{ otherwise}.
        \end{cases}\\
        &= \frac{1}{|T|} \left|\sum_{t \in T_\gamma} \nature_t - |T_\gamma|\gamma\right|.
    \end{align*}
    Thus,
    \begin{align*}
        \sum_{\gamma \in \forecaster_T} \frac{|T|}{|T_\gamma|} \capital(\sceptic^\gamma)^2
        &= \sum_{\gamma \in \forecaster_T} \frac{|T|}{|T_\gamma|} \left( \frac{|T_\gamma|}{|T|} \left|\frac{1}{|T_\gamma|}\sum_{t \in T_\gamma} \nature_t - \gamma\right|\right)^2\\
        &= \sum_{\gamma \in \forecaster_T} \frac{|T_\gamma|}{|T|} \left(  \frac{1}{|T_\gamma|}\sum_{t \in T_\gamma} \nature_t - \gamma\right)^2,
    \end{align*}
    and
    \begin{align*}
        \sum_{\gamma \in \forecaster_T} \left| \capital(\sceptic^\gamma) \right|
        &= \sum_{\gamma \in \forecaster_T} \frac{|T_\gamma|}{|T|} \left|\frac{1}{|T|}\sum_{t \in T_\gamma} \nature_t - \gamma\right|.
    \end{align*}
\end{proof}
The previous two recovery results might seem slightly restrictive as the predictive scenarios were simply estimating the mean of a binary outcome. However, this restriction is not necessary as the following recovery result shows. In particular, we refine the beliefs of the gamblers by adding group membership information. The recovered notion of multicalibration generalizes parts of Proposition~\ref{prop:recovery of calibration score}.
\begin{proposition}[Recovery of (Generalized) Multicalibration Score]
\label{prop:Recovery of (Generalized) Multicalibration Score}
    Let $\mathcal{S} \subseteq 2^T$ be a set of groups and $\forecaster_T \coloneqq \{ \forecaster_t \colon t \in T\}$.
    Let $(\Y, \Gamma, T, \forecaster, \sceptic^{\forecaster_T \times  \mathcal{S}}, \nature)$ be an evaluation protocol with $\Y \subseteq \reals$ and $\Gamma \colon \Q \to 2^\propspace$ being an identifiable property with identification function $\ident$. Furthermore, we assume there is no $\gamma \in \propspace$ such that $\Gamma^{-1}(\gamma) = \Q$.
     For every $\gamma \in \{ \forecaster_t \colon t \in T \}$ and $S \in \mathcal{S}$ we define a fair, restricted and approximate rational gambler $\sceptic^{\gamma,S}$ such that:
    \begin{enumerate}[(a)]
        \item On all instances $T_{\gamma, S} \coloneqq \{ t \in T\colon \forecaster_t =\gamma\} \cap S$, the gambler $\sceptic^{\gamma, S}$ has a (true) belief $\belief^{\gamma, S}_t$ about the average outcome distribution,
        \begin{align*}
            \belief^{\gamma, S}_t = \hat{D}_{Y|{\gamma, S}} \coloneqq \frac{1}{|T_{\gamma, S}|}\sum_{t' \in T_{\gamma, S}} \dirac(\nature_{t'}) \in \Q.
        \end{align*}
        On all other instances $T \setminus T_{{\gamma, S}}$, the gambler has no belief.
        \item The gambler is restricted to play gambles in $\ball_{\alpha_\gamma} \coloneqq \{ g \in \CY\colon \|g \| \le \alpha_\gamma \}$ where $\alpha_\gamma = \| \ident_\gamma\|$ for some norm $\| \cdot \|$.
    \end{enumerate}
    Then, the reweighted and aggregated capital of the gamblers $\sceptic^{\gamma, S}$ indexed by $\gamma$ and $S$ is equal to,
    \begin{align*}
        \sup_{S \in \mathcal{S}} \sum_{\gamma \in \forecaster_T} \frac{|T|}{|T_{\gamma, S}|} \capital(\sceptic^{\gamma, S})^2
        &=\sup_{S \in \mathcal{S}} \sum_{\gamma \in \forecaster_T} \frac{1}{|T||T_{\gamma, S}|} \left(\sum_{t \in T_{\gamma, S}} \ident(\nature_t, \gamma )\right)^2,
    \end{align*}
    which is approximate $(\mathcal{S}, \ident)$-multicalibration \citep[Definition 4.2]{noarov2023scope}\footnote{This is a generalization of $L_2$-multicalibration following Definition 2.3 in \citep{garg2024oracle}.}. Furthermore, the aggregated capital of the gamblers $\sceptic^{\gamma, S}$ indexed by $\gamma$ and $S$ is equal to,
    \begin{align*}
        \sup_{S \in \mathcal{S}} \sum_{\gamma \in \forecaster_T} \left|\capital(\sceptic^{\gamma, S})\right|
        &=\sup_{S \in \mathcal{S}} \sum_{\gamma \in \forecaster_T} \frac{|T_{\gamma, S}|}{|T|} \left| \frac{1}{|T_{\gamma, S}|}\sum_{t \in T_{\gamma, S}} \ident(\nature_t, \gamma )\right|,
    \end{align*}
    which is a generalization of $L_1$-multicalibration \citep[Definition 2.3]{garg2024oracle}.
\end{proposition}
\begin{proof}
    Fix $\gamma \in \forecaster_T$ and $S \in \mathcal{S}$. Define $\alpha_\gamma = \frac{1}{\| \ident_\gamma \|_1}$. Note that $\|\ident_\gamma \|_1 > 0$, because otherwise $\ident_\gamma = 0$ which implies that $\Gamma(\phi) = \gamma$ for all $\phi \in \Delta(\Y)$, which we ruled out by assumption.
    
    Let us consider a fixed instance $t \in T_{\gamma, S}$. Given the gambler's belief, the fair, restricted and approximate rational gambler plays (Lemma~\ref{lemma:Optimal Gamble for Rational Gambler Restricted to Calibration Gambles II}),
    \begin{align*}
        \sceptic^{\gamma, S}_t
        &\colon \begin{cases}
            y \mapsto \ident(y, \gamma) \text{ if } \mathbb{E}_{\hat{D}_{Y|{\gamma, S}}}[\ident_\gamma] > 0\\
            y \mapsto -\ident(y, \gamma) \text{ otherwise},
        \end{cases}
    \end{align*}
    For any fixed instance $t \in T \setminus T_{\gamma, S}$, the gambler's belief $\belief^{\gamma, S}_t =\square$ is vacuous, hence by Definition~\ref{def:fair restricted and approximate rational gambler}, $\sceptic^{\gamma, S}_t = 0$. In total, we obtain the aggregated capital for the gambler $\sceptic^{\gamma, S}$,
    \begin{align*}
        \capital(\sceptic^{\gamma, S})
        &=\frac{1}{|T|}\sum_{t \in T_{\gamma, S}} \sceptic^{\gamma, S}_t(\nature_t)\\
        &=\frac{1}{|T|}\sum_{t \in T_{\gamma, S}} \begin{cases}
            \ident(\nature_t, \gamma) \text{ if } \mathbb{E}_{\hat{D}_{Y|{\gamma, S}}}[\ident_\gamma] > 0\\
            - \ident(\nature_t, \gamma) \text{ otherwise}.
        \end{cases}\\
        &= \frac{1}{|T|} \left|\sum_{t \in T_{\gamma, S}} \ident(\nature_t, \gamma )\right|.
    \end{align*}
    Thus,
    \begin{align*}
        \sup_{S \in \mathcal{S}} \sum_{\gamma \in \forecaster_T} \frac{|T|}{|T_{\gamma, S}|}\capital(\sceptic^{\gamma, S})^2
        &=\sup_{S \in \mathcal{S}} \sum_{\gamma \in \forecaster_T} \frac{|T|}{|T_{\gamma, S}|} \frac{1}{|T|^2} \left(\sum_{t \in T_{\gamma, S}} \ident(Y, \gamma )\right)^2\\
        &=\sup_{S \in \mathcal{S}} \sum_{\gamma \in \forecaster_T} \frac{1}{|T||T_{\gamma, S}|} \left(\sum_{t \in T_{\gamma, S}} \ident(Y, \gamma )\right)^2,
    \end{align*}
    and
    \begin{align*}
        \sup_{S \in \mathcal{S}} \sum_{\gamma \in \forecaster_T} \left|\capital(\sceptic^{\gamma, S})\right|
        &=\sup_{S \in \mathcal{S}} \sum_{\gamma \in \forecaster_T} \frac{1}{|T|} \left|\sum_{t \in T_{\gamma, S}} \ident(\nature_t, \gamma )\right|\\
        &=\sup_{S \in \mathcal{S}} \sum_{\gamma \in \forecaster_T} \frac{|T_{\gamma, S}|}{|T|} \left|\frac{1}{|T_{\gamma, S}|}\sum_{t \in T_{\gamma, S}} \ident(\nature_t, \gamma )\right|.
    \end{align*}
\end{proof}
\begin{lemma}[Gamble for Rational Gambler Restricted to Norm Ball - $\Y \subseteq \reals$]
\label{lemma:Optimal Gamble for Rational Gambler Restricted to Calibration Gambles II}
    Let $(\Y, \Gamma, T, \forecaster, \{ \sceptic\}, \nature)$ be an evaluation protocol with $\Y \subseteq \reals$ and $\Gamma\colon \Q \to 2^\propspace$ being an identifiable property with identification function $\ident$. Furthermore, assume there is no $\gamma \in \propspace$ such that $\Gamma^{-1}(\gamma) = \Delta(\Y)$. Fix $t \in T$ and consider a fair, restricted and approximately rational gambler $\sceptic$ with belief $\belief_t \in \Q$ and restriction $\ball_{\alpha_{\forecaster_t}} \coloneqq \{ g \in \CY\colon \|g \| \le \alpha_{t} \}$ where $\alpha_{t} = \| \ident_{\forecaster_t}\|$ for some norm $\| \cdot \|$.\footnote{Note that $\| \ident_{\forecaster_t} \| > 0$, because otherwise $\ident_{\forecaster_t} = 0$ which implies that $\Gamma(\phi) = \forecaster_t$ for all $\phi \in \Delta(\Y)$, which we ruled out by assumption.} Then, in this round $t \in T$, the fair, restricted and approximately rational gambler $\sceptic$ plays the gamble,
    \begin{align*}
        \sceptic_t
        &\colon \begin{cases}
            y \mapsto \ident(y, \forecaster_t) \text{ if } \mathbb{E}_{\belief_t}[\ident_{\forecaster_t}] > 0\\
            y \mapsto -\ident(y, \forecaster_t) \text{ otherwise},
        \end{cases}
    \end{align*}
\end{lemma}
\begin{proof}
    The statement follows since for every $\epsilon > 0$,
    \begin{align*}
        \mathbb{E}_{\belief_t}[\sceptic_t]
        &= |\mathbb{E}_{\belief_t}[\ident_{\forecaster_t}]|\\
        &= \max_{\alpha \in [-1,1]} \mathbb{E}_{\belief_t}[\alpha  \ident_{\forecaster_t}]\\
        &\overset{(i)}{=} \max_{\alpha \in \reals \text{ and } \|\alpha \ident_{\forecaster_t} \|_1 \le \alpha_t } \mathbb{E}_{\belief_t}[\alpha \ident_{\forecaster_t}]\\
        &\overset{(ii)}{\ge} \max_{g \in \mathcal{F}_{\ident_{\forecaster_t}} \cap \ball_{\alpha_t} } \mathbb{E}_{\belief_t}[g]\\
        &\overset{(iii)}{\ge} \max_{g \in \mathcal{H}_{\ident_{\forecaster_t}} \cap \ball_{\alpha_t}} \mathbb{E}_{\belief_t}[g] -\epsilon\\
        &\overset{(iv)}{=} \max_{g \in \offer_{\Gamma^{-1}(\forecaster_t)} \cap \ball_{\alpha_t} } \mathbb{E}_{\belief_t}[g] - \epsilon,
    \end{align*}
    because,
    \begin{enumerate}[(i)]
        \item It holds $\|\alpha \ident_{\forecaster_t} \| = |\alpha| \| \ident_{\forecaster_t} \| = |\alpha|\alpha_t$.
        \item By defining $\mathcal{F}_{\ident_{\forecaster_t}} \coloneqq \{ g \in \CY \colon g \le \alpha \ident_{\forecaster_t} \text{ for some }\alpha \in \reals \}$.
        \item Equation~\ref{eq:dominated by calibration gambles} and for every $\epsilon > 0$ and every $h \in \mathcal{H}_{\ident_{{\forecaster_t}}}$ there exists an $f \in \mathcal{F}_{\ident_{\forecaster_t}}$ such that $|\mathbb{E}_{\hat{D}_{Y|S}}[h] - \mathbb{E}_{\hat{D}_{Y|S}}[f] | < \epsilon$ by definition of $\pqtopology$-closure of $\mathcal{F}_{\ident_{\forecaster_t}}$.
        \item Theorem~\ref{thm:Characterization of Available Gambles in Protocol 2 with Identifiable Property}.
    \end{enumerate}
\end{proof}
Finally, we equip the gambler with ``clairvoyance'', analogous to Proposition~\ref{prop:recovery of loss scores}. We recover a notion of individual calibration inspired by Equation (4) in \citep{luo2022local} (\cf \citep{holtgen2023richness} for the notion of calibration on individuals).
\begin{proposition}[Recovery of Individual Calibration Score]
\label{prop:recovery of individual calibration}
    Let $(\Y, \Gamma, T, \forecaster, \{ \sceptic\}, \nature)$ be an evaluation protocol with $\Y \subseteq \reals$ and $\Gamma \colon \Q \to 2^\propspace$ being an identifiable property with identification function $\ident$. Furthermore, we assume there is no $\gamma \in \propspace$ such that $\Gamma^{-1}(\gamma) = \Delta(\Y)$. 
    We define a fair, restricted and approximate rational gambler $\sceptic$ such that:
    \begin{enumerate}[(a)]
        \item On every instance $t \in T$ the gambler $\sceptic$ is ``clairvoyant'', \ie, has a (true) belief $\belief_t$ about the outcome sampled from a distribution chosen by Nature, $\belief_t = \dirac(\nature_t) \in \Q$.
        \item The gambler is restricted to play gambles in $\ball_{\alpha_\gamma} \coloneqq \{ g \in \CY\colon \|g \| \le \alpha_\gamma \}$ where $\alpha_\gamma = \| \ident_\gamma\|$ for some norm $\| \cdot \|$.
    \end{enumerate}
    Then, the capital of the gambler $\sceptic$ is equal to,
    \begin{align*}
        \capital(\sceptic) = \frac{1}{|T|}\sum_{t \in T} |\ident(\nature_t, \forecaster_t)|.
    \end{align*}
\end{proposition}
\begin{proof}
    Define $\alpha_t = \frac{1}{\| \ident_{\forecaster_t} \|_1}$. Note that $\| \ident_{\forecaster_t} \|_1 > 0$, because otherwise $\ident_{\forecaster_t} = 0$ which implies that $\Gamma(\phi) = \forecaster_t$ for all $\phi \in \Delta(\Y)$, which we ruled out by assumption.
    
    Given the gambler's belief, the fair, restricted and approximate rational gambler plays (Lemma~\ref{lemma:Optimal Gamble for Rational Gambler Restricted to Calibration Gambles II}),
    \begin{align*}
        \sceptic_t
        &\colon \begin{cases}
            y \mapsto \ident(y, \forecaster_t) \text{ if } \ident(\nature_t, \forecaster_t) > 0\\
            y \mapsto - \ident(y, \forecaster_t) \text{ otherwise}.
        \end{cases}
    \end{align*}
    In total, we obtain the following aggregated capital for the gambler $\sceptic$,
    \begin{align*}
        \capital(\sceptic)
        =\frac{1}{|T|}\sum_{t \in T} \sceptic_t(\nature_t)
        = \frac{1}{|T|}\sum_{t \in T} |\ident(\nature_t, \forecaster_t)|.
    \end{align*}
\end{proof}
The given recoveries only cover a small fraction of all existing notions. However, as the breadth of provided recoveries suggests, more notions can be recovered by adopting the evaluation protocol, the restriction and the belief on a fair, restricted and (approximately) rational gambler. As the proofs share a lot of their structure, we do not provide more examples in this work.

\subsection{On the Relationship between Regret and Calibration: Second Act}
\label{Across Regret and Calibration}
Besides that the recovery results justify the usefulness of the evaluation protocol in understanding current evaluation metrics, the above statements shed light on the debated duality of calibration scores and loss scores (\cf Section~\ref{Related Work: Linking Calibration and Loss}). See Table~\ref{tab:belief + restriction gives evaulation metric} for a summary.

(Zeroed) Loss scores are the result of a clairvoyant gambler with a restriction to regret gambles, while calibration scores are the result of forecast-wise informed gamblers with a restriction to the unit ball. Hence, calibration scores are in two dimensions, belief and restriction, different to (zeroed) loss scores. That makes any comparison between the scores misled.\footnote{One might be inclined to state that this comparison would be a comparison between ``apples and oranges''. This can be a fitting allegory, as on an abstract level the comparison between spherical objects seems implementable. In our case, the comparability of sets of available gambles for regret and calibration as in Section~\ref{On Optimality of Unscaled Regret or Unscaled Calibration Gambles} forms exactly this abstraction.} In contrast to the observation that in general \emph{scaled} regret gambles are to a certain extent equivalent to calibration gambles (Section~\ref{On Optimality of Unscaled Regret or Unscaled Calibration Gambles}).

\subsection{Bounds on Evaluation Scores by Hierarchy on Beliefs and Restrictions}
What is possible is the comparison along a single dimension. For instance, what is the relationship between two notions when the belief is fixed but the restriction is loosened? Or what is the relationship between two notions when restriction is fixed, but the beliefs of one gambler are more informative about the true outcomes than the beliefs of another gambler? We provide capital bounds along both dimensions. On the one hand, a set-theoretic order on the restrictions provides a hierarchy on the capital of gamblers with fixed belief. On the other hand, a refinement order (Definition~\ref{def:refinement of belief}) of the beliefs provides a hierarchy on the capitals of gamblers with fixed restriction.

Let us first take a look on the dimension of restrictions. Intuitively speaking, a gambler which has access to more gambles can choose ``better'' gambles which make the gambler richer than one which has access to less gambles. Interestingly, the following more general proposition can be used to recover a known result on a bound of swap regret by calibration scores (Corollary~\ref{corollary:Recovery of Theorem 12}).
\begin{proposition}[Hierarchy of Capital by Order on Restrictions]
\label{prop:Bounding Capital by Hierarchy on Restrictions}
    Consider an evaluation protocol $(\Y, \Gamma, T, \forecaster, \{ \sceptic, \sceptic'\}, \nature)$.
    Let $\sceptic$ and $\sceptic'$ be fair, restricted and rational gamblers with, for every $t \in T$, equal beliefs $\belief_t \in \Delta(\Y) \cup \{ \square \}$ which are aligned to truth, but different restricting sets $\restrictionset_t \subseteq \restrictionset'_t \subseteq \CY$ such that $0 \in \restrictionset_t$. Suppose whenever $\belief_t = \belief_{t'}$ for $t, t' \in T$, then $\forecaster_t = \forecaster_{t'}$, $\restrictionset_t = \restrictionset_{t'}$ and $\restrictionset'_t = \restrictionset'_{t'}$. Then, the accumulated capital of the gambler $\sceptic$ is smaller than the accumulated capital of the gambler $\sceptic'$,
    \begin{align*}
        \capital(\sceptic) \le \capital(\sceptic').
    \end{align*}
\end{proposition}
\begin{proof}
    First, we rewrite,
    \begin{align}
    \label{eq:bounding the average capital in proof}
        \capital(\sceptic) = \frac{1}{|T|} \sum_{t \in T} \sceptic_t(\nature_t) = \frac{1}{|T|} \sum_{\beta \in \belief_T} \sum_{t \in T_\beta} \sceptic_t(\nature_t)
        = \sum_{\beta \in \belief_T}  \frac{|T_\beta|}{|T|}  \frac{1}{|T_\beta|}\sum_{t \in T_\beta} \sceptic_t(\nature_t),
    \end{align}
    where $T_\beta \coloneqq \{ t \in T \colon \belief_t = \beta\}$ and $\belief_T \coloneqq \{ \belief_t \colon t \in T\}$.
    Let us focus on a single $\beta \in \belief_T$. If $\beta = \square$, then $\sceptic_t = \sceptic'_t = 0$ for $t \in T_\beta$ and we are done. Hence, suppose that $\beta \neq \square$.

    By assumption, for every $t \in T_\beta$, $\forecaster_t, G_t, G'_t$ are constant, hence we write without loss of generality, $\forecaster_\beta, G_\beta, G'_\beta$.
    By rationality of the gamblers, for every $t \in T_\beta$,
    \begin{align*}
        \mathbb{E}_{\beta}[\sceptic_t] \ge \mathbb{E}_{\beta}[g] 
    \end{align*}
    for all $g \in G_\beta \cap \offer_{\Gamma^{-1}(\forecaster_\beta)}$, and, for every $t \in T_\beta$,
    \begin{align*}
        \mathbb{E}_{\beta}[\sceptic'_t] \ge \mathbb{E}_{\beta}[g] 
    \end{align*}
    for all $g \in G'_\beta \cap \offer_{\Gamma^{-1}(\forecaster_\beta)}$. It follows, for all $t \in T_\beta$,
    \begin{align*}
        \mathbb{E}_{\beta}[\sceptic_t] \le \mathbb{E}_{\beta}[\sceptic'_t],
    \end{align*}
    because $G_\beta \subseteq G'_\beta$. Furthermore, for all $t, t'\in T_\beta$, $\sceptic_t = \sceptic_{t'}$ (as well as $\sceptic'_t = \sceptic'_{t'}$), \ie, we can write $\sceptic_\beta$ (respectively $\sceptic'_\beta$ instead).
    
    By alignment to truth (Definition~\ref{def:alignment to truth}),
    \begin{align*}
        \frac{1}{|T_\beta|}\sum_{t \in T_\beta} \sceptic_t(\nature _t) = \mathbb{E}_{\beta}[\sceptic_\beta].
    \end{align*}
    With the last two observations we can continue rewriting Equation~\eqref{eq:bounding the average capital in proof},
    \begin{align*}
        \sum_{\beta \in \belief_T}  \frac{|T_\beta|}{|T|} \mathbb{E}_{\beta}[\sceptic_\beta] \le \sum_{\beta \in \belief_T}  \frac{|T_\beta|}{|T|} \mathbb{E}_{\beta}[\sceptic'_\beta]
        = \frac{1}{|T|} \sum_{t \in T} \sceptic'_t(y_t) = \capital(\sceptic').
    \end{align*}
\end{proof}
For Proposition~\ref{prop:Bounding Capital by Hierarchy on Restrictions} two hidden assumptions are of crucial importance. (a) The beliefs have to be aligned to truth. This guarantees that the expected outcome of the gamble aligns to the true (average) outcome. (b) The forecasts and restrictions have to be constant for instances for which the belief is constant. This guarantees that the set of gambles from which both gamblers choose remains constant. Otherwise, it is hard to show the superiority of the choice of one gambler over the choice of another gambler.

We can directly apply Proposition~\ref{prop:Bounding Capital by Hierarchy on Restrictions} to better understand the relationship between calibration scores and swap regret. We recover Theorem 12 in \citep{kleinberg2023ucalibration}, which exists in different formulations and flavors at several places of the literature (\cf Section~\ref{Related Work: Linking Calibration and Loss})
\begin{corollary}[Recovery of Theorem 12 in \citep{kleinberg2023ucalibration}]
\label{corollary:Recovery of Theorem 12}
    Let $\Y = \{ 0,1\}$ and $\forecaster_T \coloneqq \{ \forecaster_t \colon t \in T\}$. Let Forecaster $\forecaster_t \in [0,1]$ predict the mean, denoted as $\Gamma$, for all $t \in T$.
    Let $\loss \colon \{0,1 \} \times [0,1] \to [-1, 1]$ be a consistent scoring function for $\Gamma$. For a fixed tuple of outcomes $(\nature_t)_{t \in T}$, the swap regret with respect to $\loss$ of the forecaster $\forecaster$ is upper bounded by the re-scaled expected calibration error,
    \begin{align*}
        \max_{\sigma \in \{ \propspace \to \propspace \} } \sum_{t \in T} \loss(\nature_t, \forecaster_t) - \loss(\nature_t, \sigma(\forecaster_t)) \le 4 \sum_{\gamma \in \forecaster_T} |T_\gamma| \left| \frac{1}{|T_\gamma|}\sum_{t \in T_\gamma} \nature_t - \gamma\right|.
    \end{align*}
\end{corollary}
\begin{proof}
    First, we define $\loss' \coloneqq \frac{1}{4} \loss$ and an evaluation protocol $(\Y, \Gamma, T, \forecaster, \{ \sceptic, \Tilde{\sceptic}\}, \nature)$. For every $\gamma \in \{ \forecaster_t \colon t \in T \}$ let us introduce two fair, restricted and rational gamblers: $\sceptic^\gamma$ and $\Tilde{\sceptic}^\gamma$. Both share their beliefs $\belief^\gamma_t = \hat{D}_{Y|\gamma} \subseteq \Delta(\Y)$ for all $t \in T_\gamma \coloneqq \{ t \in T \colon \forecaster_t = \gamma \}$ and $\belief^\gamma_t = \square$ for all $t \in T \setminus T_\gamma$. The beliefs are aligned to the truth (Definition~\ref{def:alignment to truth}).
    But the gamblers are differently restricted, $\sceptic^\gamma_t \in \UnscaledRegretGambles'_\gamma$ and $\Tilde{\sceptic}^\gamma_t \in \ball_1$. Where $\UnscaledRegretGambles'_\gamma \coloneqq \{ \loss'_\gamma - \loss'_c \colon c \in [0,1]\}$. Note that $ \UnscaledRegretGambles'_\gamma \subseteq \ball_1$, as for every $g \in \UnscaledRegretGambles'_\gamma$,
    \begin{align*}
        \| g\|_1 = \| \loss_\gamma - \loss_c \|_1
        =|\loss(1, \gamma) - \loss(1, c)| + |\loss(0, \gamma) - \loss(0, c)|
        \le 1,
    \end{align*}
    because $\loss' \colon \{0,1 \} \times [0,1] \to [-\frac{1}{4}, \frac{1}{4}]$.

    By Proposition~\ref{prop:recovery of calibration score} Equation~\ref{eq: ECE (l1)} the right hand side of the above inequality can be written as,
    \begin{align*}
        |T | \sum_{\gamma \in \forecaster_T}  \capital(\Tilde{\sceptic}), \quad \text{   and   } \quad |T | \sum_{\gamma \in \forecaster_T}  \capital(\sceptic).
    \end{align*}
    respectively for the left hand side (Proposition~\ref{prop:recovery of swap regret}). The inequality then follows by Proposition~\ref{prop:Bounding Capital by Hierarchy on Restrictions} and rescaling the loss from $\loss'$ to $\loss$.
\end{proof}
Along the dimension of beliefs, we define a hierarchy on the beliefs by their fine-grainedness. Essentially, the more fine-grained a belief is, the more informed the gambler with the belief.
\begin{definition}[Refinement of Belief]
\label{def:refinement of belief}
    Let $T \coloneqq \{ 1,\ldots, n\}$ and $\belief_t, \belief_t' \in \Delta(\Y) \cup \{ \square \}$ for every $t \in T$. We define $T_\beta \coloneqq \{ t \in T \colon \belief_t = \beta\}$ and $\belief_T \coloneqq \{ \belief_t \colon t \in T\}$.
    The beliefs $(\belief_t)_{t \in T}$ are \emph{refined by} $(\belief_t')_{t \in T}$, if for all $\beta \in \belief_T$, $\beta = \square$ or $\belief'_t \in \Delta(\Y)$ for all $t \in T_\beta$ and $\beta = \frac{1}{|T_\beta|} \sum_{t \in T_\beta} \belief_t'$.
\end{definition}
Refinement of belief does not require any belief to be aligned to truth (Definition~\ref{def:alignment to truth}). However, by comparing the Definition~\ref{def:refinement of belief} to Definition~\ref{def:alignment to truth}, we observe that, beliefs $(\belief_t)_{t \in T}$ are aligned to truth, if and only if, $(\dirac(\nature_t))_{t \in T}$ refines $(\belief_t)$ under an appropriate choice of evaluation protocol.

The more informed a gambler is about the true outcomes the better it can optimize its capital. That is what we formalize in the following.
\begin{proposition}[Hierarchy on Capital by Refinements on Belief]
\label{prop:Bounding Capital by Refinment on Belief}
    Consider an evaluation protocol $(\Y, \Gamma, T, \forecaster, \{ \sceptic, \sceptic'\}, \nature)$.
    Let $\sceptic$ and $\sceptic'$ be fair, restricted and rational gamblers with, for every $t \in T$, equal restriction $G_t \subseteq \CY$ such that $0 \in \restrictionset$, but different beliefs $\belief_t, \belief_t' \in \Delta(\Y) \cup \{ \square \}$ for every $t \in T$. Suppose whenever $\belief_t = \belief_{t'}$ and $\belief'_t = \belief'_{t'}$ for $t, t' \in T$, then $\forecaster_t = \forecaster_{t'}$ and $G_t = G_{t'}$. If the beliefs $(\belief_t')_{t \in T}$ are aligned to truth and $(\belief_t')_{t \in T}$ refines $(\belief_t)_{t \in T}$, then the accumulated capital of the gambler $\sceptic$ is smaller than the accumulated capital of the gambler $\sceptic'$,
    \begin{align*}
        \capital(\sceptic) \le \capital(\sceptic').
    \end{align*}
\end{proposition}
\begin{proof}
    First, we rewrite, analogous to the proof of Proposition~\ref{prop:Bounding Capital by Hierarchy on Restrictions},
    \begin{align}
    \label{eq:bounding the average capital in proof 2}
        \capital(\sceptic) = \frac{1}{|T|} \sum_{t \in T} \sceptic_t(\nature_t) &= \frac{1}{|T|} \sum_{\beta \in \belief_T} \sum_{t \in T_\beta} \sceptic_t(\nature_t)\\
        &= \sum_{\beta \in \belief_T}  \frac{|T_\beta|}{|T|}  \frac{1}{|T_\beta|}\sum_{t \in T_\beta} \sceptic_t(\nature_t)\\
        &= \sum_{\beta \in \belief_T}  \frac{|T_\beta|}{|T|}  \sum_{t \in T_\beta} \frac{|T_{\beta'}|}{|T_\beta|} \sum_{t \in T_{\beta'}} \frac{1}{|T_{\beta'}|} \sceptic_t(\nature_t),
    \end{align}
    where $T_\beta \coloneqq \{ t \in T \colon \belief_t = \beta\}$, $\belief_T \coloneqq \{ \belief_t \colon t \in T \}$ and $T_{\beta'} = T_\beta \cap \{ t \in T\colon \belief'_t = \beta'\}$ for $\beta' \in \{ \belief_t' \colon t \in T_\beta \}$.

    Let us focus on the last sum over $t \in T_{\beta'}$. Clearly, $\belief_t$ and $\belief'_{t}$ are constant for all $t \in T_{\beta'}$. Hence, by assumption, for all $t \in T_{\beta'}$ $\forecaster_t$ and $G_t$ are constant. We rewrite without loss of generality $\forecaster_{\beta'}$ and $G_{\beta'}$. Furthermore, by fairness, restrictiveness and rationality of the gamblers, for all $t, t' \in T_{\beta'}$, $\sceptic_t = \sceptic_{t'} \in G_{\beta'} \cap \offer_{\Gamma^{-1}(\forecaster_{\beta'})}$ (respectively $\sceptic'_t = \sceptic'_{t'} \in G_{\beta'} \cap \offer_{\Gamma^{-1}(\forecaster_{\beta'})}$), \ie, we can write $\sceptic_{\beta'}$ (respectively $\sceptic'_{\beta'}$ instead). We open two cases:
    \begin{description}
        \item[Suppose $\beta' = \square$.] Then, $\beta = \square$ by refinement of belief (Definition~\ref{def:refinement of belief}), hence $\sceptic_{\beta'} = \sceptic'_{\beta'} = 0$. Thus,
        \begin{align*}
            \sum_{t \in T_{\beta'}} \frac{1}{|T_{\beta'}|} \sceptic_t(\nature_t) = 0 = \sum_{t \in T_{\beta'}} \frac{1}{|T_{\beta'}|} \sceptic'_t(\nature_t).
        \end{align*}
        \item[Suppose $\beta' \in \Delta(\Y)$.] Focusing on the gambler $\sceptic'$, for every $t \in T_{\beta'}$,
        \begin{align*}
            \mathbb{E}_{\beta'}[\sceptic'_{\beta'}] \ge \mathbb{E}_{\beta'}[g] 
        \end{align*}
        for all $g \in G_{\beta'} \cap \offer_{\Gamma^{-1}(\forecaster_{\beta'})}$, in particular,
        \begin{align*}
            \mathbb{E}_{\beta'}[\sceptic'_{\beta'}] \ge \mathbb{E}_{\beta'}[[\sceptic_{\beta'}].
        \end{align*}
        Since $(\belief'_t)_{t \in T}$ is aligned to truth,
        \begin{align*}
            \sum_{t \in T_{\beta'}} \frac{1}{|T_{\beta'}|} \sceptic_t(\nature_t) = \mathbb{E}_{\beta'}[\sceptic_{\beta'}]
            \le \mathbb{E}_{\beta'}[\sceptic'_{\beta'}] =\sum_{t \in T_{\beta'}} \frac{1}{|T_{\beta'}|} \sceptic'_t(\nature_t).
        \end{align*}
    \end{description}
    Combining both of the cases and Equation~\eqref{eq:bounding the average capital in proof 2} gives the result.
\end{proof}
As in Proposition~\ref{prop:Bounding Capital by Hierarchy on Restrictions}, we need consistency assumptions on the restrictions and forecasts when the beliefs don't change. Again, this guarantees that the more informed gambler actually chooses among the same set of gambles the superior one (in terms of final capital) in comparison to the less informed gambler. Otherwise the statement does not hold in general as the following examples shows.
\begin{example}
    Proposition~\ref{prop:Bounding Capital by Refinment on Belief} does \emph{not} hold if the restriction $G$ varies arbitrarily over time. For instance, let $\Y = \{ 0,1\}$, $\Gamma$ be the mean, $T = \{t_1, t_2, t_3 \}$, $\forecaster = (\frac{5}{12}, \frac{5}{12}, \frac{5}{12})$, $\nature = (0,1,0)$ and $\sceptic, \sceptic'$ be two fair, restricted and rational gamblers with beliefs $\belief_t = \frac{1}{3}$ for all $t \in T$ respectively $(\belief_t')_{t \in T} = (\frac{1}{2}, \frac{1}{2}, 0)$ and restrictions $(G_t)_{t \in T} = (\ball_1, \CY_{\le 0}, \CY_{\le 0})$. We observe the following:
    \begin{enumerate}
        \item Both gamblers have beliefs which are aligned to truth. The beliefs of $\sceptic'$ refine the beliefs of $\sceptic$.
        \item Both gamblers, as being rational and restricted with $\CY_{\le 0}$, play $\sceptic_t = \sceptic_t' = 0$ for $t \in \{ t_2, t_3\}$.
        \item For $t = t_1$, the gambler $\sceptic$ plays $\sceptic_{t_1}\colon y \mapsto \frac{5}{12} - y$, while the gambler $\sceptic'$ plays $\sceptic'_{t_1}\colon y \mapsto y - \frac{5}{12}$.
    \end{enumerate}
    As a result, the capitals compare as follows,
    \begin{align*}
        \capital(\sceptic) = \frac{1}{3}\left(0 + 0 + \frac{5}{12}\right) \ge \frac{1}{3}\left(0 + 0 - \frac{5}{12} \right)= \capital(\sceptic').
    \end{align*}
\end{example}
The attentive reader might have noticed that in Table~\ref{tab:belief + restriction gives evaulation metric} we essentially provided a hierarchy of refining beliefs, from $\hat{D}_{Y}$ over $\hat{D}_{Y|\gamma}$ to $\dirac(\nature_t)$. This suggests the two trivial recovery corollaries below. External regret is upper bounded by swap regret, which is upper bounded by (zeroed) loss score. Bias in the large is upper bounded by calibration score (ECE), which is upper bounded by individual calibration score. We omit the detailed proof as the statements follow almost immediately. The ideas are simple. From external to swap regret/bias in the large to calibration score: For every gambler with constant belief $\hat{D}_{Y|\gamma}$ which we introduce in Proposition~\ref{prop:recovery of swap regret}, we introduce a gambler with constant belief $\hat{D}_{Y}$. Then, apply Proposition~\ref{prop:Bounding Capital by Refinment on Belief}. From swap regret to zeroed loss scores/calibration score to individual calibration score: Directly apply Proposition~\ref{prop:Bounding Capital by Refinment on Belief}.
\begin{corollary}[Recovery of Known Order of Regret Scores]
\label{corollary:Recovery of Known Order of Regret}
    Let $\Y \subseteq \mathbb{R}^d$, $|T| < \infty$, $\nature_t \in \Y$ for all $t \in T$, $\forecaster_t \in \propspace$ for all $t \in T$ and $\Gamma \colon \Q \to 2^\propspace$ an elicitable property with consistent scoring function $\loss$. It holds,
    \begin{align*}
        \sum_{t \in T} \loss(\nature_t, \forecaster_t) - \min_{c \in \propspace}\loss(\nature_t, c) \le \max_{\sigma \in \{ \propspace \to \propspace \} } \sum_{t \in T} \loss(\nature_t, \forecaster_t) - \loss(\nature_t, \sigma(\forecaster_t)) \le \sum_{t \in T} \loss(\nature_t, \forecaster_t) - \loss(\nature_t, \Gamma(\dirac(\nature_t)).
    \end{align*}
\end{corollary}
\begin{corollary}[Recovery of Known Order of Calibration Scores]
\label{corollary:Recovery of Known Order of calibration scores}
    Let $\Y = \{ 0,1\}$, $|T| < \infty$, $\nature_t \in \Y$ for all $t \in T$, $\forecaster_t \in [0,1]$ for all $t \in T$ and $\Gamma \colon \Q \to [0,1]$ be the mean with identification function $\ident(y, \gamma) = y - \gamma$. It holds,
    \begin{align*}
        \left|\frac{1}{|T|}\sum_{t \in T} (\nature_t - \forecaster_t) \right| &\le \sum_{\gamma \in \forecaster_T} \frac{|T_\gamma|}{|T|} \left| \frac{1}{|T_\gamma|}\sum_{t \in T_\gamma} \nature_t - \gamma\right| \le \frac{1}{|T|}\sum_{t \in T} |\nature_t - \forecaster_t|.
    \end{align*}
\end{corollary}

\section{Related Work}
\label{sec:related work}
The content of this paper has been motivated by and built upon several strands of work. We contextualize our findings in two sections. One section focuses on the testing of forecasts. The other section concentrates on the duality of regret and calibration.

\subsection{Testing Forecasts}
\label{Related Work - Testing Forecasts}
Our suggested evaluation protocol for the recovery of existing notions was motivated by existing literature.

\paragraph{Meteorology and historical probabilistic forecast evaluation.}
Historically, the question of how to evaluate (probabilistic) forecasts has been a major concern in meteorology (\eg, \citep{brier1950verification, murphy1967verification}). With the advent of powerful machine learning systems which produce forecasts of arbitrary kinds, we can observe the renaissance of old debates on what constitutes a ``good'' (weather) forecast. In \citep{murphy1967verification}, the authors suggest a separation of operational evaluation, what is the value of the forecast to the user, and empirical evaluation, how well do forecasts and observations align. Even though, we don't fully agree to the meaningfulness of this separation we do want to highlight in this respect that our work is focusing on ``statistical adequateness'' criteria alone. We are convinced that for a ``successful'' forecast more adequateness criteria have to be met (\cf Section~\ref{future work}).

\paragraph{Manipulability and the limit of sequential tests.}
The testing of (sequential) forecasts has received interest in the economics literature. In particular, this strand of literature has been concerned with the manipulability of tests, \ie, whether Forecaster can fake knowing the truth and still pass a test. Key contributions included proofs which show that if a test is complete, \ie, a test accepts the truth (which is type-I error freeness), then the test is manipulable \citep{foster1998asymptotic, lehrer2001any, sandroni2003reproducible, olszewski2011falsifiability}. Central to these results is the option for Forecaster to randomize over possible forecasts \citep{vovk2005defensive, vovk2005defensive2, vovk2007continuous, olszewski2011falsifiability}.
We showed in Proposition~\ref{proposition:sanity check} that in our evaluation protocol completeness is equivalent to a fair gambler, \ie, a gambler which only plays available gambles. In follow-up works, scholars argued about the existence of non-manipulable but complete tests, \eg, in \citep{fortnow2009complexity} the authors show that the faking forecaster is computationally ``complex'',in \citep{shmaya2008many} the author suggests a Borel-set construction of a non-manipulable test via the axiom of choice, and in \citep{olszewski2009nonmanipulable} the authors equip the test to ask for counterfactual queries which makes it resilient to manipulating forecasts.

\paragraph{Forecast evaluation as test - From classical tests to e-variables and non-negative supermartingales.}
Finally, the evaluation of a forecast can equivalently be understood as a classical statistical test (\eg, \citep{neyman1933ix}) whether the (composite) null hypothesis, \ie, the forecasting set defined through the forecast and the forecasted property, fits to the data observed, \ie, what Nature reveals. Recently, e-values, a replacement for p-values, have gained increasing attention \citep{grunwald2020safe, vovk2021values}. Essentially, e-variables, \ie, unrealized e-values, is a gambling approach to testing statistical hypothesis testing. The studied questions and the setup in this work has been motivated by the surrounding field of game-theoretic probability and statistics \citep{shafer2019game} (going back to \citep{ville1939etude}).

E-variables are, in our language, gambles, for which the availability constraint (Definition~\ref{def:fair gambler and availability}) is replaced by an expectation being equal or less than $1$ and the restriction (Definition~\ref{def:fair, restricted and rational gambler}) is replaced by non-negativity. The analogue of rationality (Definition~\ref{def:fair, restricted and rational gambler}) is growth rate optimality (GRO). An e-variable is growth rate optimal, if its expected logarithmic value is maximal under the alternative hypothesis. In our case, a gambler is rational, if the expected values of its gambles is maximal under its belief, which is the alternative hypothesis. Hence, our rationality demand is comparable to GRO up to the concave monotone transformation of the ``units'' of the gamble's outcome.

In the realm of game-theoretic statistics, the closest work to ours is \citep{casgrain2022anytime}. \citet{casgrain2022anytime} propose test supermartingales (non-negative supermartingales with initial expected value less or equal to $1$) introduced in \citep{shafer2011test} to test a composite null hypothesis which is equivalent to the level sets which we consider for elicitable (respectively identifiable) properties (\cf Definition~\ref{def:level set} and Proposition~\ref{prop:Identifiable or Elicitable Property Give Credal Level Sets}). Test supermartingales are, (a) a special case of an e-process \citep{ramdas2022testing}, which is the sequential generalization of an e-value \citep{ramdas2023game} and (b) a sum of available gambles plus an initial value in $[0,1]$.

A central difference between ours and their work is that in \citep{casgrain2022anytime} the authors consider a stationary forecaster. In our setting, Forecaster can provide a new forecast for every instance. That is why we concentrate on identifying properties of single-instance based gambles, while \citep{casgrain2022anytime} consider sequential tests. Nevertheless, they provide answers to related questions. The authors identify a subset of the possible supermartingales which tests the level set null hypothesis. In comparison, we provide full characterizations of the set of available gambles, \ie, ``instance-wise tests'' (Theorem~\ref{thm:Characterization of Available Gambles in Protocol 2 with Elicitable Property} and Theorem~\ref{thm:Characterization of Available Gambles in Protocol 2 with Identifiable Property}).
The authors maximize the power of their test, \ie, minimize the type-II error, based on GRO against an alternative hypothesis which is the empirical distribution of the seen instances. We consider rationality, the analogue of GRO as laid out above, against an alternative hypothesis which can include the actual outcomes of unseen realizations. This is obviously irrealistic, hence, we emphasis that our (recovery) results can only be interpreted as ``as if''-statements.
Finally, they state that the test supermartingales which they suggest for level sets of identifiable properties are more powerful, than for elicitable properties \citep[\p 9 \& 10]{casgrain2022anytime}. We observe an analogous hierarchy in Section~\ref{On Optimality of Unscaled Regret or Unscaled Calibration Gambles}.

Because of the multiplicative structure of the test supermartingales in \citep{casgrain2022anytime} and the logarithm in the GRO criterion \citep{grunwald2020safe}, we summarize on an abstract level, that our theory is providing an ``additive analogue'' to the more ``multiplicative'' theory of game-theoretic statistics.


\subsection{Duality of Regret-Type and Calibration-Type Consistency}
\label{Related Work: Linking Calibration and Loss}
In Section~\ref{On Optimality of Unscaled Regret or Unscaled Calibration Gambles} and Section~\ref{Across Regret and Calibration} we stumble upon a fundamental duality of consistency criteria of probability assignments: regret-type and calibration-type consistency criteria.

\paragraph{Internal consistency.}
In understanding ``consistency'' as \emph{internal consistency}, \ie, the probability assignments on a single happening are not contradicting, the loss-type and calibration-type duality dates back at least to \citep{de2017theory}. De Finetti suggested two different definitions of internal consistency, what he calls \emph{coherence} \citep{schervish2009proper}. One definition, the regret-type, states that a probabilistic assignment is coherent if it minimizes a scoring rule. The other definition, the calibration-type, requires that a gambler with access to available, scaled calibration gambles (\cf Theorem~\ref{characerizing available gambles of identifiable properties}) does not get rich by gambling across events. If one type of coherence is fulfilled then the probabilities follow the standard rules of probability, \eg, sum to one, disjoint additivity.\footnote{\citet{schervish2009proper} trace back accounts for coherence of probabilistic statements and show limits when it comes to their generalization to imprecise probabilities. This has been done for the approach via calibration gambles in the seminal works of \citet{walley1991statistical} and \citet{williams2007notes}. An analogous generalized theory for losses is still under development \citep{konek2023evaluating}.}

\paragraph{External consistency.}
On the other hand, we can understand ``consistency'' as \emph{external consistency}, \ie, the probability assignments match empirical observations. Analogous to internal consistency there exist equivalence statements between regret-type and calibration-type definitions. First, to our knowledge, were \citet[Equation 4.1 \& 4.2]{degroot1983comparison} who provide a direct translation of calibration scores to swap regret scores for the Brier scoring function. Similar bridges have been established in \citet{foster1998asymptotic} and \citep[Section 2.3]{foster1999regret}. Independent of these works, \citet[Theorem 8.1]{dawid1985calibration} showed that a (computably) calibrated (computable) forecast has lower proper scoring rule than any compared computable forecast for infinite time horizon.

More recently, equivalences between regret and calibration criteria of forecast regained interest. The reason for this is that multicalibration, a generalization of calibration, was introduced as a possible candidate to guarantee fair forecasts \citep{chouldechova2017fair, hebert2018multicalibration}. For instance, \citet[Theorem 3.2]{globus2023multicalibration} provide a characterization for multicalibration as kind of a ``swap regret''. Closely related \citet{gopalan2023swap} showed the equivalence of ``swap regret'' omnipredictor and multicalibrated predictor. The equivalence relationship between ``swap regret'' and calibration has been generalized beyond the mean in \citep{derr2025three}. Finally, \citet[Theorem 12]{kleinberg2023ucalibration} and \citet{noarov2023high} give results which imply low swap regret for (multi-)calibrated predictors for relatively general loss functions.\footnote{Surprisingly, already \citet[Theorem 4]{degroot1983comparison} made some unnoticed progress in that regard. They argued that swap regret is controlled by the calibration of Forecaster. The authors didn't use the term ``swap regret''. We refer to Equation (5.5) in their work as swap regret.} In particular, we recover \citep[Theorem 12]{kleinberg2023ucalibration} in Corollary~\ref{corollary:Recovery of Theorem 12} using the more general Proposition~\ref{prop:Bounding Capital by Hierarchy on Restrictions}.

The close relationship between regret-type external consistency and calibration-type external consistency becomes apparent time and again. Nevertheless, cases are made advocating calibration over regret \citep{foster1998asymptotic,zhao2021calibrating, kleinberg2023ucalibration} other cases are made arguing for the advantages of regret over calibration \citep{schervish1985self,seidenfeld1985calibration, dawid1985calibration}. Our Section~\ref{On Optimality of Unscaled Regret or Unscaled Calibration Gambles} and Section~\ref{Across Regret and Calibration} highlight why the debate ``calibration versus regret'' needs to be done at the right level of comparison: along the level of available gambles (Section~\ref{On Optimality of Unscaled Regret or Unscaled Calibration Gambles}) or along the level of scores (Section~\ref{Across Regret and Calibration}).

\paragraph{Elicitation and identification of properties.}
Finally, the strand of statistical literature on property elicitation mentioned already in Section~\ref{the forecasts} provides further relationships between the regret and the calibration world. Because of the centrality of scoring functions, regret, on an abstract level, is linked to elicitability (Definition~\ref{def:scoring function}). To remind the reader, a property is elicitable if it minimizes a consistent scoring function. The fundamental result by \citet{steinwart2014elicitation} generalizes the equivalence of elicitable and identifiable properties, if the property is real-valued and continuous given in \citep{lambert2008eliciting}. In particular, a sufficiently differentiable scoring function which elicits a property and an identification function which identifies the same property are linked via the gradients of the scoring function \citep[Theorem 3.2]{fissler2016higher}. Then, in \citep{noarov2023scope} and \citep{gneiting2023regression} calibration got understood as approximating the identification criterion of a forecast. Similar to \citep{gopalan2022low, deng2023happy} they observe that standard calibration, \eg, Proposition~\ref{prop:recovery of calibration score}, largely is based on a specific instance-wise comparison between observation and forecast, namely $\nature - \forecaster$. This comparison is an identification function of the mean. Hence, \citep{noarov2023scope} and \citep{gneiting2023regression} generalized this instance-wise comparison to arbitrary identification functions. In particular, \citet{noarov2023scope} identified the set of calibratable properties by the set of elicitable respectively identifiable properties. Concluding, calibration, on an abstract level, is linked to identifiability. If elicitablity and identifiability both are given, then equivalence of the two worlds calibration and regret is, at least to a certain extent modulo scaling, given (Corollary~\ref{corollary:Duality of Calibration and Regret}).

\section{Conclusion}
\label{conclusion}
We have argued that most used evaluation metrics used in machine learning can be viewed as the result of a fair gambling protocol between a forecaster and a gambler with certain restrictions and beliefs.
Provocatively written, \textbf{all machine learning evaluation is gambling}.
We are convinced this insight adds to the debate on the value of current evaluation practice in machine learning when facing (social) reality, \eg, \citep{liu2022lost, wang2024against}.
For instance, we can now ask whether we want to let the success of gamblers decide which methods to favor when doing breast cancer prediction, \eg, \citep{chen2025deep}?

Furthermore, our work sheds light on the relationship of regret-type and calibration-type evaluation metrics. Both those types of metrics are fair. The difference of the types are different restrictions on the set of gambles which the gambler is allowed to play beyond the available ones. Then, a meaningful comparison between the obtained scores is possible when the belief is fixed, \eg, individual calibration scores with loss scores, standard calibration scores with swap regret or multicalibration scores with group-wise swap regret. Nevertheless, scaled regret-type evaluation metrics can, in principle, make up for the difference between the set restrictions. Regret and calibration are two flavors of evaluation metrics, theoretically equivalent in their power, practically not comparable.


\subsection{Future Work}
\label{future work}
Several interesting, open questions remain to be answered in future work. For instance, can we characterize the set of available gambles for a forecasting set induced by an elicitable or identifiable property by a criterion beyond calibration or regret? Related to this, how can it be guaranteed that a forecasting set of an arbitrary property, which we only demand to be credal, adequately describes Nature's outcome (\cf Appendix~\ref{appendix: Availability Criterion for Imprecise Forecasts})? On the other hand, what are more meta-criteria for reasonable evaluation metrics beyond ``fairness''? Finally, how does this evaluation protocol interact with evaluation methodologies, \eg, benchmarking \citep{hardt2025emerging}, which provide more semantics to the obtained scores? 


\acks{
The authors are very grateful to Christian Fröhlich, Benedikt Höltgen, Aaron Roth, Giacomo Molinari and Min Jae Song for insightful discussions. Thanks as well for the feedback of the audience attending the talk ``Four Facets of Evaluating Predictions'' at Harvard University, USA, November 2023, and the talk ``Four Facets of Forecast Felicity'' during the Workshop on ``Learning under Weakly Structured Information'' at the University of Tübingen, Germany, April 2025. Thanks to the International Max Planck Research School for
Intelligent System (IMPRS-IS) for supporting Rabanus Derr. 

This work was funded in part by the Deutsche Forschungsgemeinschaft
(DFG, German Research Foundation) under Germany’s Excellence Strategy –-
EXC number 2064/1 –- Project number 390727645; it was also supported by the 
German Federal Ministry of Education and Research (BMBF): Tübingen AI Center.
}

\appendix

\section{Representability of Evaluation Metrics}
\label{appendix:representability of evaluation metrics}
\begin{proposition}[Characterization of Representability]
    \label{prop:Recoverability of Evaluation Metric_appendix}
    Let $T \coloneqq \{ 1, \ldots, n\}$, $\Y$ be an outcome set and $\propspace$ a set of property values.
    An evaluation metric $m \colon \Y^T \times \propspace^T \to \reals$ is \emph{representable in an evaluation protocol}, if and only if, for all $\nature \in \Y^T$ and $\forecaster \in \propspace^T$, there exists a set of functions $f^i_t \colon \Y \times \propspace \to \reals$ indexed by $t \in T$ and $i \in I$ for some finite $I$ and an aggregation $\aggregation \colon \reals^{|I|} \to \reals$, such that,
    \begin{align*}
        m((\nature_t)_{t \in T}, (\forecaster_t)_{t \in T}) = \aggregation\left[\left(\frac{1}{|T|} \sum_{t \in T}f^i_t(\nature_t,\forecaster_t)\right)_{i \in I}\right].
    \end{align*}
\end{proposition}
\begin{proof}
    Given $T, \forecaster, \nature$ and $\propspace$, let $\Gamma \colon \Q \to 2^\propspace$ be arbitrary and define the evaluation protocol $(\Y, \Gamma, T, \forecaster, \sceptic^\indexsetgamblers, \nature)$, where for every $i \in \indexsetgamblers$, we define a gambler $\sceptic^i$ such that $\sceptic^i_t \colon y \mapsto f^i_t(y, \forecaster_t)$. Then, the capital of each such gambler is,
    \begin{align*}
        \capital(\sceptic^i) = \frac{1}{|T|} \sum_{t \in T} f^i_t(\nature_{t}, \forecaster_{t}).
    \end{align*}
    By choosing the appropriate aggregation function $\aggregation$ we obtain,
    \begin{align*}
        \aggregation[(\capital(\sceptic^i))_{i \in I}] = \aggregation\left[\left(\frac{1}{|T|} \sum_{t \in T}f^i_t(\nature_t,\forecaster_t)\right)_{i \in I}\right].
    \end{align*}
    The reverse direction follows analogously, by demanding that for every $i \in \indexsetgamblers$ and $t \in T$ there is a function $f^{\sceptic^i}_t(\cdot, \forecaster_t) \colon y \mapsto \sceptic^i_t(y)$.
\end{proof}
In particular, every single-instance based evaluation metric is representable.
\begin{proposition}[Single-Instance Based Evaluation Metrics are Representable]
\label{prop:single-instance based evaluation metrics are representable of Evaluation Metric}
    Let $T \coloneqq \{ 1, \ldots, n\}$, $\Y$ be an outcome set and $\propspace$ a set of property values.
    The single-instance based evaluation metric $m \colon \Y^T \times \propspace^T \to \reals$,
    \begin{align*}
        m((\nature_t)_{t \in T}, (\forecaster_t)_{t \in T}) = \aggregation[(f_t(\nature_t,\forecaster_t))_{t \in T}],
    \end{align*}
    for some aggregation function $\aggregation \colon \reals^T \to \reals$, and some $f_t \colon \Y \times \propspace \to \reals$ for each $t \in T$ is representable in an evaluation protocol.
\end{proposition}
\begin{proof}
    Given $T, \forecaster, \nature$ and $\propspace$, let $\Gamma \colon \Q \to 2^\propspace$ be arbitrary and define the evaluation protocol $(\Y, \Gamma, T, \forecaster, \sceptic^\indexsetgamblers, \nature)$, where for every ${\Tilde{t}} \in T$, we define a gambler $\sceptic^{\Tilde{t}}$ such that $\sceptic^{\Tilde{t}}_t = 0$ if $t \neq {\Tilde{t}}$ and $\sceptic^{\Tilde{t}}_{\Tilde{t}} \colon y \mapsto f_{\Tilde{t}}(\nature_{\Tilde{t}}, y)$. Then, the averaged capital of each such gambler is,
    \begin{align*}
        \capital(\sceptic^{\Tilde{t}}) = \frac{1}{|T|} f_{\Tilde{t}}(\nature_{\Tilde{t}}, \forecaster_{\Tilde{t}}).
    \end{align*}
    By choosing the appropriate aggregation function $\aggregation'$, which is $\aggregation'[(r_t)_{t \in T}] = \aggregation[|T|(r_t)_{t \in T}]$ for $(r_t)_{t \in T} \in \reals^{|T|}$, we obtain,
    \begin{align*}
        \aggregation'[(\capital(\sceptic^t))_{t \in T}] = \aggregation[(f_t(\nature_t,\forecaster_t))_{t \in T}] .
    \end{align*}
\end{proof}

\section{Elicitable or Identifiable Properties Have Convex and Closed Level Sets}
\label{appendix:elictiable or identifable properties credal sets}
\begin{proposition}[Convex and Closed Level Sets]
\label{prop:Identifiable or Elicitable Property Give Credal Level Sets-Appendix}
    For all $\gamma \in \propspace$ the level set $\Gamma^{-1}(\gamma) \subseteq \Delta(\Y)$ of an elicitable or identifiable property $\Gamma$ is credal, \ie, convex and $\qptopology$-closed.
\end{proposition}
\begin{proof}
    Let $\Gamma$ be elicitable with scoring function $\scoring$. By Definition~\ref{def:scoring function}, for all $\gamma \in \propspace$, $\scoring_\gamma \in \CY$. Hence, all integrals are well-defined. Fix any $\gamma \in \propspace$. By assumption in Definition~\ref{def:level set}, $\Gamma^{-1}(\gamma)\neq \emptyset$. 
    The convexity of the level sets follows by a straightforward argument comparable to \citep[Appendix B Theorem 13]{steinwart2014elicitation}). Let $\phi_1, \phi_2 \in \Gamma^{-1}(\gamma)$ and $\alpha \in [0,1]$, for all $\gamma' \in \propspace$,
    \begin{align*}
        \langle \alpha \phi_1 + (1-\alpha) \phi_2 , \scoring_{\gamma'} \rangle &= \alpha \langle  \phi_1, \scoring_{\gamma'} \rangle + (1-\alpha) \langle \phi_2 , \scoring_{\gamma'} \rangle\\
        &\ge \alpha \langle  \phi_1, \scoring_{\gamma} \rangle + (1-\alpha) \langle \phi_2 , \scoring_{\gamma} \rangle\\
        &= \langle \alpha \phi_1 + (1-\alpha) \phi_2 , \scoring_{\gamma} \rangle,
    \end{align*}
    hence, $\alpha \phi_1 + (1-\alpha) \phi_2  \in \Gamma^{-1}(\gamma)$ by convexity of $\Q$. For the closure, observe that
    \begin{align*}
        \Gamma^{-1}(\gamma) &\coloneqq \{ \phi \in \Q \colon \gamma \in \Gamma(\phi) \}\\
        &= \left\{ \phi \in \Q \colon \mathbb{E}_{\phi}[\scoring(Y, \gamma)] \le \mathbb{E}_{\phi}[\scoring(Y, c)], \forall c \in \propspace \right\}\\
        &= \left\{ \phi \in \Q \colon \langle \phi, \scoring_\gamma - \scoring_c \rangle \le 0, \forall c \in \propspace \right\}\\
        &= \left\{ \phi \in \ca(\Y) \colon \langle \phi, \scoring_\gamma - \scoring_c \rangle \le 0, \forall c \in \propspace \right\}  \cap \Q\\
        &= \bigcap_{c \in \propspace} \left\{ \phi \in \ca(\Y) \colon \langle \phi, \scoring_\gamma - \scoring_c \rangle \le 0 \right\}  \cap \Q,
    \end{align*}
    is $\qptopology$-closed since $\left\{ \phi \in \ca(\Y) \colon \langle \phi, \scoring_\gamma - \scoring_c \rangle \le 0 \right\}$ is closed by the definition of $\qptopology$-topology and any intersection of closed sets is closed.

    Let $\Gamma$ be identifiable with identification function $\ident$. By Definition~\ref{def:identification function}, for all $\gamma \in \propspace$, $\ident_\gamma \in \CY$. Hence, all integrals are well-defined. Fix any $\gamma \in \propspace$. By assumption in Definition~\ref{def:level set}, $\Gamma^{-1}(\gamma)\neq \emptyset$. 
    The convexity of the level sets follows directly by the definition of identification function. For the closure, observe that
    \begin{align*}
        \Gamma^{-1}(\gamma) &\coloneqq \{ \phi \in \Q \colon \gamma \in \Gamma(\phi\}\\
        &= \{ \phi \in \Q \colon \langle \phi, \ident_\gamma \rangle = 0 \}\\
        &= \{ \phi \in \ca(\Y) \colon \langle \phi, \ident_\gamma \rangle = 0 \} \cap \Q\\
        &= \{ \phi \in \ca(\Y) \colon \langle \phi, \ident_\gamma \rangle \le 0 \} \cap \{ \phi \in \ca(\Y) \colon \langle \phi, \ident_\gamma \rangle \ge 0 \} \cap \Q,
    \end{align*}
    is $\qptopology$-closed.
\end{proof}

\section{Additional Lemmas and Examples}
\begin{lemma}[Convex Cone]
\label{lemma:conic hull}
    If $A \subseteq \CY$ or $A \subseteq \ca(\Y)$ is convex, then $\reals_{\ge 0} A$ is a convex cone.
\end{lemma}
\begin{proof}
    Note, $\reals_{\ge 0} A$ is closed under positive scalar multiplication. It remains to show that for $r_1a_1, r_2a_2 \in \reals_{\ge 0} A$ the sum $r_1a_1 + r_2a_2 \in  \reals_{\ge 0} A$. To this end, we choose $r' \coloneqq r_1 + r_2 \in \reals_{\ge 0}$ and $a' \coloneqq \frac{r_1}{r_1 + r_2} a_1 + \frac{r_2}{r_1 + r_2} a_2 \in A$ by convexity. We note $r' a' \in \reals_{\ge 0}A$ and $r' p' = r_1a_1 + r_2a_2$.
\end{proof}

\begin{lemma}[Vacuous Forecasts Only Make Non-Positive Gambles Available]
\label{lemma:vacuous forecasts only make non-positive gambles available}
    Let $\forecastingset = \Delta(\Y)$. Then, 
    \begin{align*}
        \{ g \in \CY \colon  \sup_{\phi \in \forecastingset}\mathbb{E}_\phi[g] \le 0\} = \CY_{\le 0}.
    \end{align*}
\end{lemma}
\begin{proof}
    Note that $\forecastingset$ is credal, \ie, $\qptopology$-closed and convex. Theorem~\ref{thm:representation: credal sets - offers - MAIN} and Proposition~\ref{prop:properties of polar set - polar set of negative quadrant} give 
    \begin{align*}
        \{ g \in \CY \colon \sup_{\phi \in \forecastingset}\mathbb{E}_\phi[g] \le 0\} &= \offer_\forecastingset = (\reals_{\ge 0} \forecastingset)^\circ = (\ca(\Y)_{\ge 0})^\circ = \CY_{\le 0}. 
    \end{align*}
\end{proof}

\begin{lemma}
\label{lemma:techincal lemma 1}
    Let $f \in \CY$. If there exists $\phi \in \Delta(\Y)$ such that $\langle \phi, f \rangle \le  0$, then $\inf f \le 0$.
\end{lemma}
\begin{proof}
    Let us give a proof by contraposition. Suppose that $\inf f > 0$. Then, for all $\phi \in \Delta(\Y)$, $\langle \phi, \alpha f \rangle \le  0$, because $\phi(A) \ge 0$ for all $A \in \Sigma(\Y)$.
\end{proof}

\begin{example}[Dominance of Single Gamble by Scaleability]
\label{ex:Dominance of Single Gamble by Scaleability}
    Consider an evaluation protocol $(\Y, \Gamma, T, \forecaster, \{\sceptic \}, \nature)$ with $\capital(\sceptic) = \frac{1}{|T|} \sum_{t \in T} \sceptic_t(\nature_t)$. Pick an arbitrary $\Tilde{t} \in T$ and change the gamble $\sceptic_{\Tilde{t}}' = \alpha \sceptic_{\Tilde{t}}$ for $\alpha \in [0,\infty)$. The updated capital is,
    \begin{align*}
        \capital(\sceptic') = \frac{1}{|T|} \sum_{t \in T} \sceptic_t(\nature_t) + \frac{1}{|T|}(\alpha-1) \sceptic_{\Tilde{t}}(\nature_{\Tilde{t}}),
    \end{align*}
    hence,
    \begin{align*}
        \frac{\capital(\sceptic')}{\alpha} = \frac{1}{|T| \alpha} \sum_{t \in T} \sceptic_t(\nature_t) + \frac{\alpha-1}{|T| \alpha} \sceptic_{\Tilde{t}}(\nature_{\Tilde{t}}) \ \overset{\alpha \to \infty}{\longrightarrow} \ \sceptic_{\Tilde{t}}(\nature_{\Tilde{t}}).
    \end{align*}
\end{example}

\section{Availability Criterion for Imprecise Forecasts}
\label{appendix: Availability Criterion for Imprecise Forecasts}
The problem of ``What is a good \emph{precise} forecast?'' has been part of debates for decades \citep{schervish1989general, schervish2009proper, gneiting2007strictly, gneiting2011making, zhao2021right, zhao2021calibrating}. However, the more general, ``what are ``good'' \emph{imprecise} forecasts?'', regains focus just recently \citep{zhao2021right, gupta2022faster, konek2023evaluating,verma2024calibration,frohlich2024scoring, singh2025truthful}\footnote{Arguably, \citet{schervish2009proper} started this project during the search for a generalization of de Finetti's coherence type II. The first generalization of de Finetti's coherence type I already let to the development of the field of imprecise probability \citep{walley1991statistical, williams2007notes}.}. The question's answer has not been settled yet. We hope our work can contribute by clarifying the problem setup.

One of the central problems in evaluating imprecise forecasts is the fine balance of (de-)incentivizing imprecision.
A vacuous forecast, \ie, the totality of all probability distributions, is, for example, a forecast in which a fair player cannot increase his capital beyond $0$.
In this paper, we circumvent the ``problem of imprecision'' by requiring the forecaster to make property-based forecasts. 
More specifically, in our approach, we focus on the evaluation of imprecise forecasts which are level sets of elicitable (respectively identifiable) properties.

\bibliography{main}

\end{document}